\documentclass{article}

\usepackage[preprint]{corl_2026} 

\usepackage{enumitem}       
\usepackage{xcolor}
\usepackage{pifont}
\usepackage{amssymb}
\usepackage{amsmath}
\usepackage{caption}
\usepackage{booktabs}
\usepackage{graphicx}
\usepackage{multirow}
\usepackage{wrapfig}
\usepackage{tikz}
\usepackage{hyperref}
\usepackage{subcaption}
\usetikzlibrary{positioning,arrows.meta,fit}
\usepackage{array}
\newcolumntype{L}[1]{>{\raggedright\arraybackslash}p{#1}}
\newcolumntype{C}[1]{>{\centering\arraybackslash}p{#1}}

\definecolor{formalgreen}{rgb}{0.1, 0.7, 0.1}
\definecolor{formalred}{rgb}{0.9, 0.2, 0.2}
\definecolor{formalorange}{rgb}{0.95, 0.55, 0.1}

\newcommand{\cmark}{\textcolor{formalgreen}{\checkmark}}
\newcommand{\xmark}{\textcolor{formalred}{\ding{55}}}
\newcommand{\pmark}{\textcolor{formalorange}{\(\triangle\)}}

\definecolor{cfblue}{HTML}{6C8EBF}

\definecolor{zhenyu_color}{RGB}{102,192,255}

\usepackage{longtable}

\definecolor{citecolor}{HTML}{0071BC}
\hypersetup{colorlinks,linkcolor={red},citecolor={citecolor}}

\title{DexVerse: A Modular Benchmark for Multi-Task, Multi-Embodiment Dexterous Manipulation}

\author{
\parbox{\dimexpr\textwidth-2\tabcolsep\relax}{\centering
  Yunchao Yao$^{1*}$ \quad
  Zhuxiu Xu$^{1,2*}$ \quad
  Tianqi Zhang$^{1}$ \quad
  Zixian Liu${^1}$ \quad
  Sikai Li$^{1}$ \quad \\
  Zhenyu Wei$^{1}$ \quad
  Feng Chen$^{2}$ \quad
  Dihong Huang$^{1}$ \quad 
  Kechang Wan$^{1}$ \quad 
  Chenyang Ma$^{1}$ \quad 
  Shuqi Zhao$^{3}$ \quad
  Shenghua Gao$^{2}$ \quad 
  Masayoshi Tomizuka$^{3}$ \quad 
  Yi Ma$^{2}$ \quad 
  Mingyu Ding$^{1\dagger}$\quad 
  \\[4pt]
  {\mdseries
  $^{1}$UNC-Chapel Hill \quad
  $^{2}$The University of Hong Kong\quad 
  $^{3}$UC Berkeley 
  }
  \\[4pt]
  {\mdseries
  {\small $^{*}$Equal contribution. \quad
  $^{\dagger}$Corresponding author}
  }
  }
}

\begin{document}
\maketitle


\begin{center}
    \vspace{-20pt}
    \includegraphics[width=\linewidth]{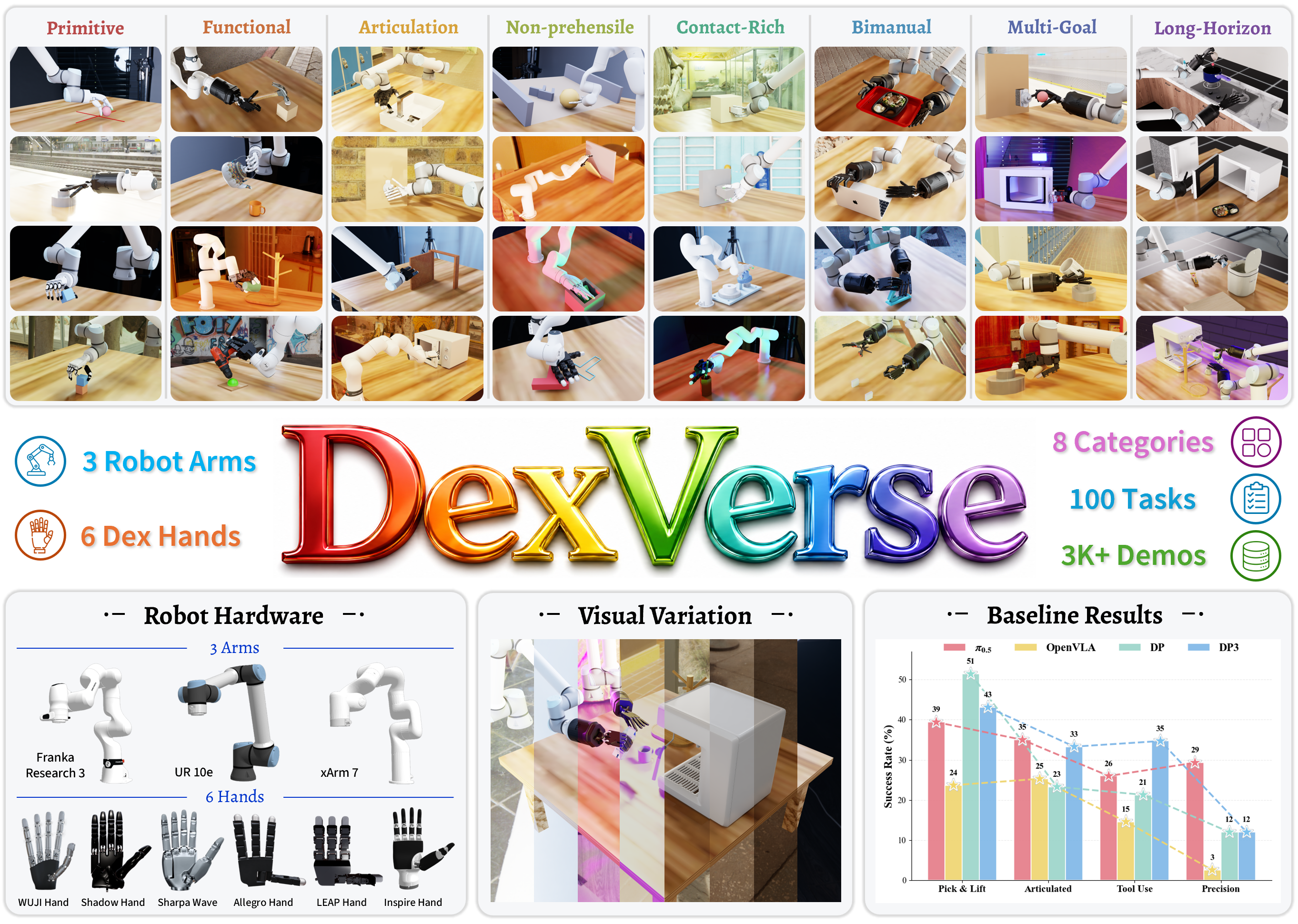}
    \vspace{-15pt}
    \captionof{figure}{Overview of DexVerse, a modular benchmark for multi-task, multi-embodiment dexterous manipulation with diverse tasks, visual variations, demonstration datasets, and baseline evaluations.}
    \label{fig:teaser}
\end{center}

\begin{abstract}
Building general-purpose dexterous manipulation policies requires benchmarks that go beyond isolated tasks to systematically evaluate policies across diverse interaction modes, sensory conditions, and robot embodiments.
However, existing benchmarks remain limited in task and data diversity, embodiment coverage, or controllable visual variation, hindering studies of cross-task and cross-embodiment generalization.
We present \textbf{DexVerse}, a large-scale and modular benchmark for dexterous manipulation.
DexVerse includes 100 tasks spanning a broad range of manipulation skills, including object grasping and relocation, articulated-object interaction, functional tool use, bimanual coordination, non-prehensile control, contact-rich behaviors, multi-goal execution, and long-horizon multi-stage task completion.
It supports 3 robot arms and 6 dexterous hands, and is extensible to new tasks, assets, and embodiments.
To evaluate visuomotor generalization, DexVerse provides configurable visual variations in textures, background, lighting, and camera viewpoints.
We further provide a VR-based teleoperation interface and 3,180 demonstrations with synchronized proprioceptive, RGB, depth, point-cloud, and state observations.
We benchmark representative methods, including Diffusion Policy, DP3, OpenVLA, and $\pi_{0.5}$, across 19 tasks.
Results reveal substantial challenges in task generalization and visuomotor robustness, establishing DexVerse as a promising testbed for general-purpose dexterous manipulation.
Project page: \url{https://ycyao216.github.io/DexVerse.site/}
\end{abstract}



\section{Introduction}

Dexterous manipulation is a central capability for building general-purpose robots~\cite{kroemer2020reviewrobotlearningmanipulation, doi:10.1126/science.aat8414}.
Moving beyond contact-poor, gripper-based manipulation that can often be approximated by reaching, grasping, transporting, and releasing, dexterous manipulation requires coordinated control of high-DoF hands and arms under intermittent, contact-rich interactions, while grounding actions in object geometry, visual affordances, force closure, and long-horizon task structure~\cite{rajeswaran2017learning, SUOMALAINEN2022104224}. 
Recent generalist robot policies have advanced this frontier from several complementary directions: image-conditioned imitation learning, such as action chunking and diffusion policies~\cite{zhao2023learning,chi2025diffusion,jing2025mixture_of_horizons}; scalable demonstration and data generation pipelines~\cite{robomimic2021,mandlekar2023mimicgen,jiang2024dexmimicen,wang2024robogenunleashinginfinitedata}; 3D-aware policies that exploit voxel, multi-view, or point-cloud representations~\cite{shridhar2022peract,goyal2023rvt,goyal2024rvt2,Ze2024DP3}; and vision-language-action policies trained on increasingly heterogeneous robot data~\cite{rt12022arxiv,embodimentcollaboration2025openxembodimentroboticlearning,octo_2023,kim24openvla,black2024pi_0,intelligence2025pi}. 
Yet it remains unclear how well these methods scale from isolated skills and controlled task distributions to functionally diverse, long-horizon, contact-rich dexterous manipulation across embodiments and environments.

A key bottleneck is the lack of benchmarks that jointly evaluate the major axes of dexterous generalization.
%
Existing long-horizon manipulation benchmarks such as CALVIN~\cite{mees2022calvin}, RoboTwin 2.0~\cite{chen2025robotwin}, ManiSkill3~\cite{taomaniskill3}, and LIBERO~\cite{liu2023libero} primarily focus on gripper-based manipulation, while dexterous benchmarks often specialize in narrower settings without expert demonstrations.
%
%
DexMimicGen~\cite{jiang2024dexmimicen} focuses on dexterous demonstration generation, Bi-DexHands~\cite{chen2022towards} emphasizes RL-based bimanual dexterous control, DexJoCo~\cite{wang2026dexjoco} studies functionally grounded dexterous tasks, and DexHoldem~\cite{chen2026dexholdemplayingtexasholdem} and DexH2R~\cite{wang2025dexh2rbenchmarkdynamicdexterous} focus on real-world poker manipulation and dynamic human-to-robot handover, respectively. 
%
%
These benchmarks have each advanced the field, but none jointly support broad dexterous tasks, multiple arm-hand embodiments, controllable visual variation, demonstration data, parallel simulation, and representative policy evaluation. This makes it difficult to compare policies across contact regimes, visual conditions, and embodiments.

We present \textbf{DexVerse}, a modular benchmark suite for multi-task, multi-embodiment dexterous manipulation, unifying broad dexterous task coverage, multi-embodiment hand-arm control, modular extensibility, expert demonstrations, parallel RL, and visuomotor generalization.
DexVerse includes \textbf{100 tasks} spanning a broad range of manipulation skills, including object grasping and relocation, articulated-object manipulation, functional tool use, bimanual coordination, and long-horizon multi-stage manipulation. The benchmark supports multiple robot arms, including Franka Research 3~\cite{franka_research3}, UR10e~\cite{ur10e_datasheet} and xArm 7~\cite{xarm_user_manual}, and multiple dexterous hands, including Sharpa Wave~\cite{sharpa_wave}, WUJI Hand~\cite{wuji_hand}, Shadow Hand~\cite{shadow_hand}, Inspire Hand~\cite{inspire_hand}, Allegro Hand~\cite{allegro_hand} and LEAP Hand~\cite{shaw2023leap}, while using reusable task templates and configuration files to add new tasks, assets, embodiments, observation modalities, and action spaces. To evaluate visuomotor generalization, DexVerse provides configurable observation-level variations, including object and scene textures, lighting conditions, and camera viewpoints.

Together with the benchmark, we release a dataset of \textbf{3,180} demonstrations collected via VR teleoperation. Each demonstration contains synchronized proprioceptive, RGB, depth, point-cloud, and simulator state observations, enabling the evaluation of state-based, image-based, and 3D policy learning methods under a unified benchmark setting. We evaluate representative imitation learning and vision-language-action policies, including Diffusion Policy~\cite{chi2025diffusion}, DP3~\cite{Ze2024DP3}, OpenVLA~\cite{kim24openvla}, and $\pi_{0.5}$~\cite{intelligence2025pi}, across 19 tasks. The results indicate that DexVerse remains highly challenging for current methods: even the best-performing baselines achieve only \textbf{34\%} mean online success rate, and no single method consistently dominates across task categories. While some policies can solve selected lifting or articulated-object tasks, they struggle substantially on tool-use and precision tasks, with several tasks receiving zero success from all baselines. These findings demonstrate that DexVerse is not saturated by existing policy learning approaches and highlight open challenges in contact-rich dexterity, fine-grained visuomotor control, and general-purpose robotic manipulation.

Our contributions are threefold.
\textbf{1)} We introduce DexVerse, a unified large-scale dexterous manipulation benchmark with 100 tasks spanning diverse interaction patterns, object dynamics, and task horizons, with support for multiple arm-hand embodiments and controllable visual variation.
\textbf{2)} We develop a VR-based teleoperation interface for collecting dexterous manipulation demonstrations and provide a multi-modal dataset of 3,180 expert demonstrations across diverse tasks and robot embodiments.
\textbf{3)} We benchmark representative policies on DexVerse and analyze their success rates across different task categories, showing that current policies remain limited on many dexterous manipulation regimes. These results highlight DexVerse as a challenging testbed for developing more general dexterous manipulation policies.

    

\vspace{-2pt}
\section{Related Works}
\vspace{-2pt}

\begin{table}[t]
    \centering
    \caption{Comparison with representative robot manipulation and dexterous manipulation benchmarks. DexVerse features broad 100 dexterous manipulation tasks, multi-embodiment support, visual variation, and demonstration datasets. \cmark: supported; \pmark: partially supported; \xmark: not supported.}
    \vspace{3pt}
    \resizebox{\textwidth}{!}{
        \begin{tabular}{@{}lccccccc@{}}
            \toprule
            Benchmark 
            & Task Coverage
            & \begin{tabular}[c]{@{}c@{}}Dexterous\\ Hand\end{tabular}
            & Bimanual
            & \begin{tabular}[c]{@{}c@{}}Multi-\\ Embodiment\end{tabular}
            & \begin{tabular}[c]{@{}c@{}}Visual\\ Variation\end{tabular}
            & \begin{tabular}[c]{@{}c@{}}Demo\\ Dataset\end{tabular}
            & \begin{tabular}[c]{@{}c@{}}Parallel\\ RL Env.\end{tabular} \\
            \midrule
            CALVIN~\cite{mees2022calvin}
            & 34 long-horizon gripper tasks
            & \xmark & \xmark & \xmark & \pmark & \cmark & \xmark \\
            
            LIBERO~\cite{liu2023libero}
            & 130 language-conditioned gripper tasks
            & \xmark & \xmark & \xmark & \pmark & \cmark & \xmark \\
            
            RoboTwin 2.0~\cite{chen2025robotwin}
            & 50 bimanual gripper tasks
            & \xmark & \cmark & \cmark & \cmark & \cmark & \pmark \\

            ManiSkill3~\cite{taomaniskill3}
            & diverse general tasks
            & \pmark & \pmark & \cmark & \pmark & \pmark & \cmark \\
            
            DexMimicGen~\cite{jiang2024dexmimicen}
            & 9 bimanual dexterous tasks
            & \cmark & \cmark & \cmark & \xmark & \cmark & \xmark \\
            
            Bi-DexHands~\cite{chen2022towards}
            & 20 RL bimanual dexterous tasks
            & \cmark & \cmark & \xmark & \xmark & \xmark & \cmark \\
            
            DexHoldem~\cite{chen2026dexholdemplayingtexasholdem}
            & 14 Texas Hold'em manipulation primitives
            & \cmark & \xmark & \xmark & \pmark & \cmark & \xmark \\
            
            DexJoCo~\cite{wang2026dexjoco}
            & 11 task-oriented dexterous tasks
            & \cmark & \cmark & \xmark & \cmark & \cmark & \xmark \\
            
            \midrule
            \textbf{DexVerse (Ours)}
            & 100 diverse dexterous tasks
            & \cmark & \cmark & \cmark & \cmark & \cmark & \cmark \\
            \bottomrule
        \end{tabular}
    }
    \label{tab:benchmark_comparison}
    \vspace{-18pt}
\end{table}

\textbf{Dexterous Manipulation Benchmarks.}
Robot manipulation benchmarks provide standardized testbeds for multi-task learning, imitation learning, and reinforcement learning. General-purpose benchmarks such as Meta-World~\cite{mclean2025metaworld}, RLBench~\cite{james2019rlbench}, ManiSkill~\cite{taomaniskill3}, CALVIN~\cite{mees2022calvin}, and LIBERO~\cite{liu2023libero} cover diverse tabletop tasks and demonstration-based learning. However, they mostly rely on parallel-jaw grippers or simple end-effectors, and thus do not fully capture the challenges of dexterous hand-arm manipulation, such as high-dimensional control, contact-rich interactions, and functional object affordances.
Recent benchmarks have explored dexterous and bimanual manipulation. DexArt~\cite{bao2023dexart} focuses on articulated-object dexterity, Adroit environments~\cite{rajeswaran2017learning} support dexterous control and imitation learning, RoboTwin 2.0~\cite{chen2025robotwin} studies multi-embodiment bimanual manipulation, and DexMimicGen~\cite{jiang2024dexmimicen} generates bimanual dexterous demonstrations from limited human data. Most closely related, DexJoCo~\cite{wang2026dexjoco} provides task-oriented dexterous manipulation tasks with human demonstrations and policy evaluation. In contrast, DexVerse offers broader dexterous task coverage, multiple arm-hand embodiments, controllable visual variation, teleoperation-collected demonstrations, and representative policy evaluations in a unified platform.

\textbf{Generalization Across Embodiments and Visual Conditions.}
Generalization remains a major challenge for dexterous manipulation policies~\cite{huang2026dexcompose,li2026coordex,liang2025dexhanddiff,bao2023dexart,wang2023dexgraspnet,xu2023unidexgrasp,wan2023unidexgrasppp,zhang2025dexgraspnet2,zhang2026unidex,zhao2025dexh2r,ma2026current}. Visual changes in textures, lighting, camera viewpoints, backgrounds, and distractors can significantly affect visuomotor policies, motivating domain randomization, visual augmentation, and multi-modal observations~\cite{lei2025rl, tobin2017domain, zhu2023learning, andrychowicz2020learning}. In dexterous manipulation, such visual shifts are especially important because policies must infer object geometry, contact affordances, and task-relevant regions from sensory inputs. Embodiment transfer is also challenging because different arms and hands vary in kinematics, degrees of freedom, workspace, joint limits, and contact geometry; policies or demonstrations for one embodiment may not directly transfer to another without retargeting or embodiment-aware representations~\cite{wei2024dro, doshi2024scaling, li2022gendexgrasp, wei2026one}. These challenges are often studied separately, but general-purpose dexterous manipulation requires robustness to both sensory changes and embodiment changes. DexVerse is designed to evaluate them in a unified setting by combining multiple robot arms and dexterous hands with controllable visual variations and synchronized proprioceptive, RGB, depth, point-cloud, and state observations.
\vspace{-2pt}
\section{DexVerse Environment}
\vspace{-2pt}

DexVerse is a large-scale modular simulation environment with 100 dexterous manipulation tasks.
The task suite covers a broad spectrum of manipulation problems, ranging from primitive and functional object manipulation to non-prehensile and contact-rich interactions, bimanual coordination, multi-goal and long-horizon tasks.
Each environment specifies the scene, assets, observation interfaces, initialization distribution, success conditions, and optional visual or physical randomization.

This design decouples task environments from robot embodiments, supporting diverse task variants and robot configurations within a unified environment framework, while keeping task objectives and evaluation conditions explicit.
This section describes the main environment components of DexVerse, including task suite (Sec.~\ref{sec:task_suite}), modular environment design (Sec.~\ref{sec:modular_design}), robot embodiments (Sec.~\ref{sec:embodiments}), visual variation (Sec.~\ref{sec:visual_variation}), and asset sources (Sec.~\ref{sec:assets}).

\begin{table*}[t]
    \centering
    \small
    \caption{
        DexVerse taxonomy with 8 categories and 100 tasks. Tasks are grouped by the dominant interaction pattern and manipulation challenge rather than only by object type.
    }
    \vspace{-3pt}
    \label{tab:task_categories}

    \begingroup
    \resizebox{0.9\linewidth}{!}{
    \begin{tabular}{
        @{}
        C{0.15\textwidth}
        C{0.08\textwidth}
        C{0.32\textwidth}
        L{0.35\textwidth}
        @{}
    }
        \toprule
        \multicolumn{1}{c}{Category}
        & \multicolumn{1}{c}{\#}
        & \multicolumn{1}{c}{Representative tasks}
        & \multicolumn{1}{c}{Key challenge} \\
        \midrule

        Primitive
        & 9
        & \texttt{PickCube}, \texttt{StackCube}, \texttt{RelocateSphere},
          \texttt{PushButton}
        & Direct interaction with simple goals and limited action complexity. \\

        \addlinespace
        Functional
        & 11
        & \texttt{HammerStrike}, \texttt{RetrieveCup}, \texttt{GraspKettle},
          \texttt{PourCan}
        & Affordance-aware interaction with task-relevant object regions. \\

        \addlinespace
        Articulation
        & 18
        & \texttt{OpenStapler}, \texttt{OpenLaptop}, \texttt{SqueezeScissors},
          \texttt{OpenPhone}
        & Controlling object parts and joints, under constrained motion. \\

        \addlinespace
        Non-prehensile
        & 5
        & \texttt{PushT}, \texttt{TakeBook}, \texttt{PivotCuboid}, \texttt{PushSphereObstacle}
        & Using pushing, sliding, pivoting, or environmental contact. \\

        \addlinespace
        Contact-rich
        & 8
        & \texttt{InsertPeg}, \texttt{PlugCharger}, \texttt{NutThread}, \texttt{InsertGear}
        & Precise alignment under sustained contact and tight constraints. \\

        \addlinespace
        Bimanual Coordination
        & 5
        & \texttt{BiLiftTray}, \texttt{BiHandover}, \texttt{BiLiftBox}, \texttt{BiLiftCart}
        & Coordinated stabilization, transfer, or cooperative manipulation. \\


        \addlinespace
        Multi-goal
        & 39
        & \texttt{GraspMug + PushButton}, \texttt{GraspCan + TurnOnSwitch}
        & Satisfying multiple goals or compositional objective conditions. \\

        \addlinespace
        Long-horizon
        & 5
        & \texttt{MakeCoffee}, \texttt{MicrowaveFood}, \texttt{CleanTable}, \texttt{OvenBake}
        & Completing temporally extended multi-stage procedures. \\

        \bottomrule
    \end{tabular}
    }
    \endgroup
    \vspace{-6pt}
\end{table*}

\begin{figure*}[t]
    \centering
    \includegraphics[width=0.88\linewidth]{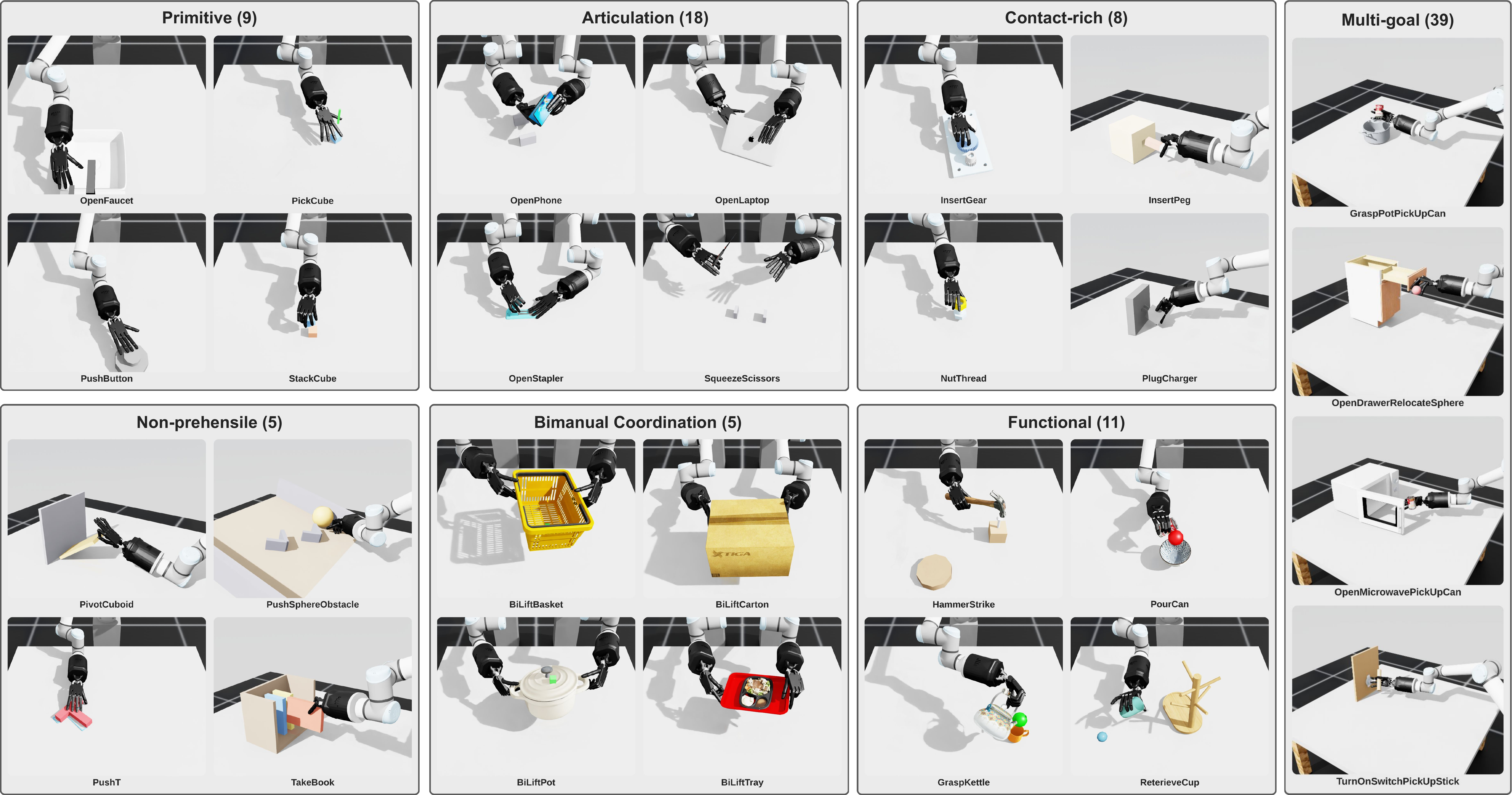}
    \caption{Visualization of selected tasks from the DexVerse environments.}
    \vspace{-8pt}
    \label{fig:env_visualizations}
\end{figure*}

\textbf{Task Suite.}
\label{sec:task_suite}
DexVerse contains 100 dexterous manipulation tasks organized into 8 categories: primitive, functional, articulation, non-prehensile, contact-rich, bimanual coordination, multi-goal, and long-horizon tasks.
The categories are summarized in Table~\ref{tab:task_categories}, and representative environments are shown in Figure~\ref{fig:env_visualizations}.
We categorize the tasks according to the dominant interaction pattern and manipulation challenge, so that the suite covers diverse forms of contact, coordination, object state change, and task temporal structure.

Each task is specified as $\mathcal{T}=(\Omega,\mathcal{S}_0,\mathcal{O},\mathcal{A},\mathcal{G})$, where $\Omega$ denotes the interactive objects in the scene, $\mathcal{S}_0$ denotes the initial-state distribution, $\mathcal{O}$ and $\mathcal{A}$ denote the task observation and action interfaces, and $\mathcal{G}$ denotes the task-level success conditions.
The success conditions specify when the intended manipulation objective is completed and are implemented as simulator predicates.
Together with the observation and action interfaces, they define the concrete environment used by a policy.

\begin{figure*}[t]
    \centering
    \includegraphics[width=\linewidth]{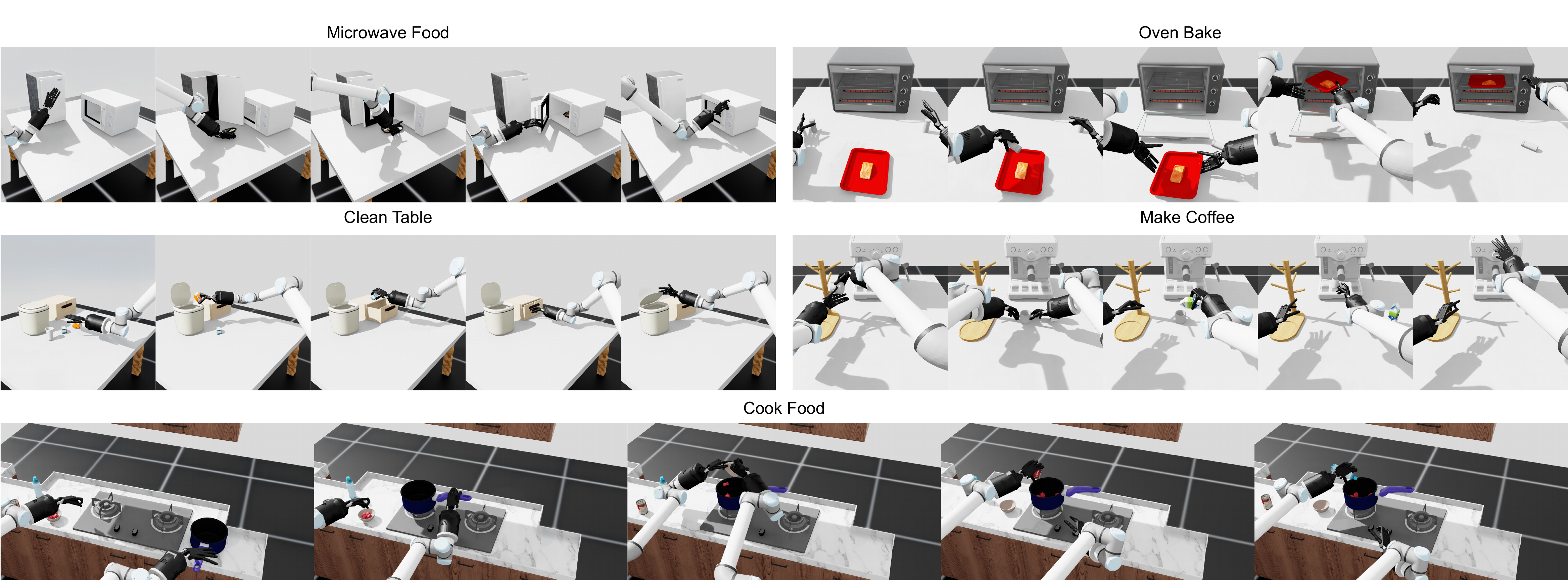}
    \caption{Visualization of task progression of the 5 long-horizon tasks in DexVerse environments.}
    \vspace{-10pt}
    \label{fig:long-horizon-rollouts}
\end{figure*}

The task suite is designed to cover complementary dexterous manipulation challenges.
Primitive and functional tasks evaluate basic object interaction and affordance-aware manipulation.
Articulation, non-prehensile, and contact-rich tasks emphasize contact regulation, constraint exploitation, and precise interaction with object geometry or articulated structure.
Bimanual coordination tasks introduce coordination requirements across two hands or arms, while multi-goal and long-horizon tasks extend the suite beyond isolated skills by requiring policies to satisfy multiple objectives or complete temporally extended procedures (Figure~\ref{fig:long-horizon-rollouts}). 
This taxonomy supports category-level analysis across manipulation regimes, while the complete task list, initialization ranges, object assets, and success thresholds are provided in the supplementary material.

\vspace{-2pt}
\subsection{Modular Environment Design}
\vspace{-2pt}
\label{sec:modular_design}

\begin{wrapfigure}{r}{0.49\linewidth}
    \centering
    \vspace{-10pt}
    \includegraphics[width=\linewidth]{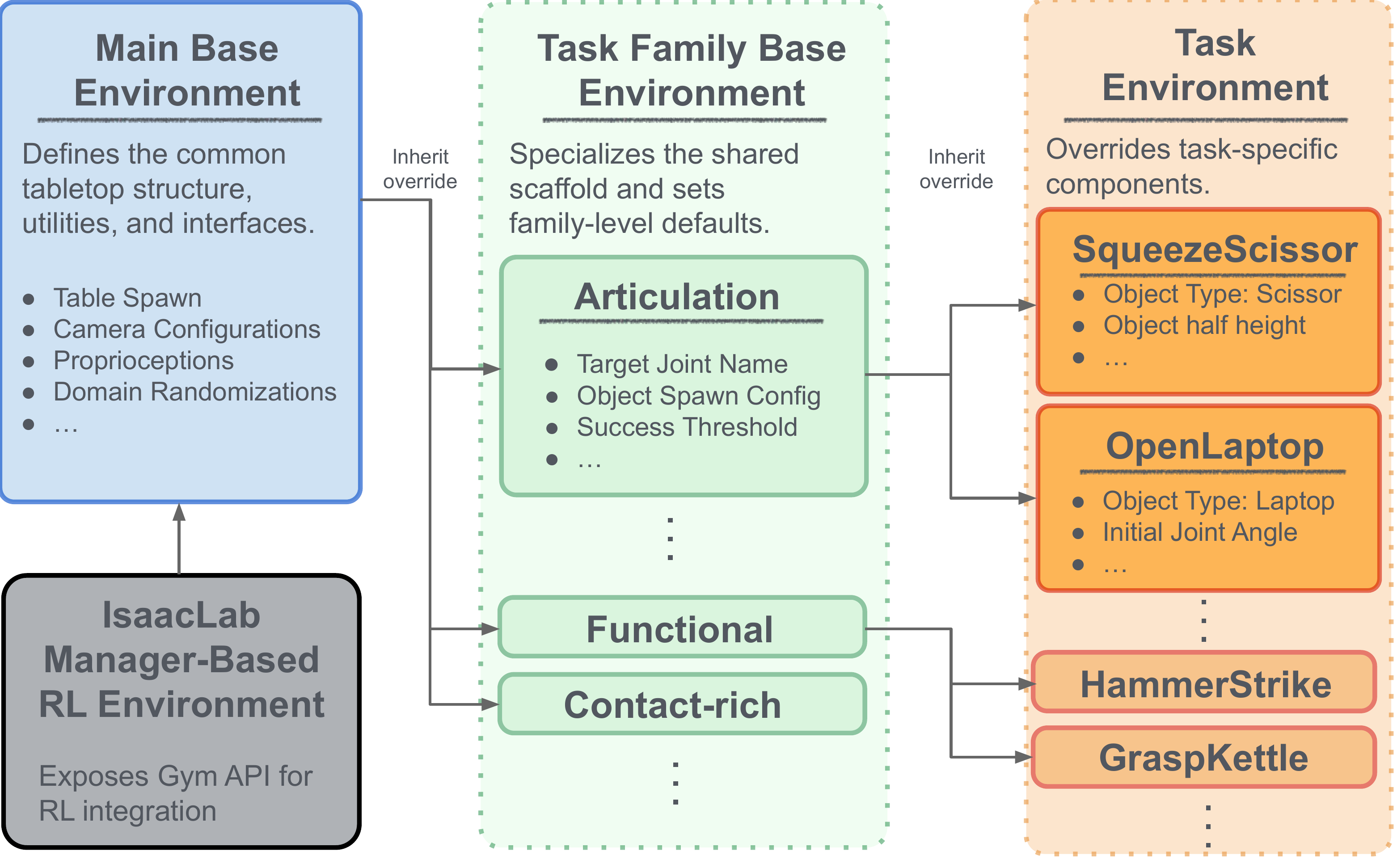}
    \vspace{-16pt}
    \caption{Modular Environment Architecture}
    \label{fig:system-design}
    \vspace{-0.5em}
\end{wrapfigure}

DexVerse uses a configuration-driven design to specify and instantiate manipulation environments. Each environment is defined by a set of structured components, including the scene layout, object assets, robot embodiment, observation and action interfaces, initialization rules, success conditions, and randomization settings. Tasks within the same family share reusable templates for common logic such as asset loading, state initialization, reset events, and success checking, while task-specific parameters define the manipulated objects, target states, sampling ranges, and completion thresholds. This design reduces duplicated implementation across related environments and makes task variants easier to construct and maintain, as illustrated in Fig.~\ref{fig:system-design}.

DexVerse builds on the manager-based environment interface of Isaac Lab, where observations, actions, events, terminations, and optional reward terms are specified through configuration classes and executed by a shared simulation loop. Most environment parameters can be adjusted through configuration overrides, enabling controlled changes to initialization ranges, camera settings, randomization options, or success thresholds without modifying the core environment code. For robot embodiments, DexVerse additionally provides a compact specification for selecting arm-hand combinations, simplifying the instantiation of feasible embodiment variants.
\vspace{-2pt}
\subsection{Robot Embodiments}
\vspace{-2pt}
\label{sec:embodiments}

\begin{figure}[t]
    \centering
    \begin{subfigure}[t]{0.58\linewidth}
        \centering
        \includegraphics[width=\linewidth]{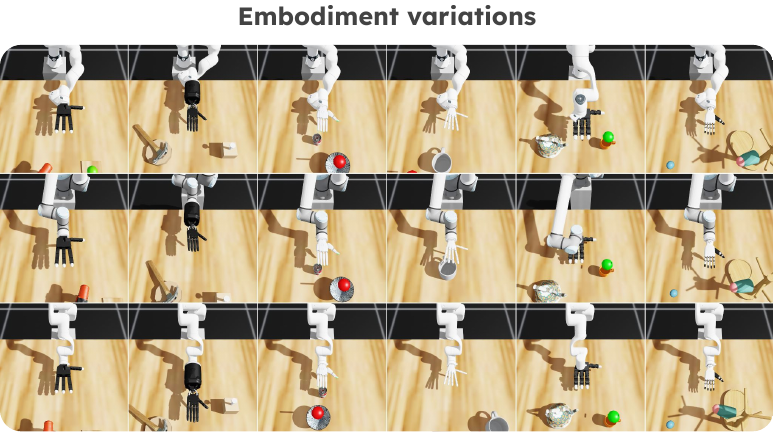}
        \caption{}
        \label{fig:embodiment_variations}
    \end{subfigure}
    \hfill
    \begin{subfigure}[t]{0.39\linewidth}
        \centering
        \includegraphics[width=\linewidth]{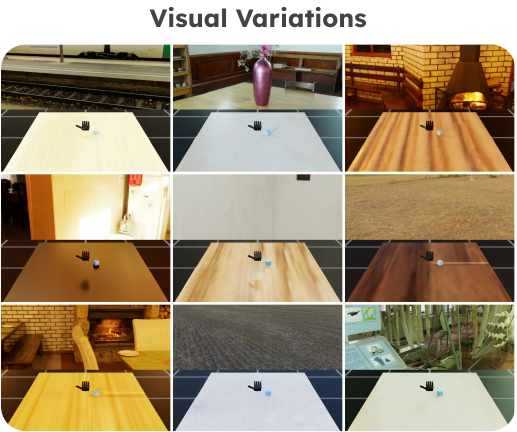}
        \caption{}
        \label{fig:visual_variations}
    \end{subfigure}
    \caption{Visual demonstration of embodiments and visual variation.}
    \label{fig:embodiment_vis}
    \vspace{-10pt}
\end{figure}


DexVerse specifies each robot embodiment through a robot configuration, which defines the arm and hand models, initial pose, action interface, controller parameters, and embodiment-specific constants.
As long as a robot implements the required interface, it can be instantiated in a task without rewriting the task-specific environment logic.
For physically feasible task-embodiment combinations, users can override the default robot choice, such as replacing a single-arm setup with a bimanual setup or switching to a different arm-hand pair.

The current benchmark supports 3 robot arms (Franka Research 3, UR10e, and xArm 7) and 6 dexterous hands (Sharpa Wave, WUJI Hand, Shadow Hand, Inspire Hand, Allegro Hand, and LEAP Hand), covering diverse kinematics, degrees of freedom, joint limits, actuation ranges, and hand morphologies.
Figure~\ref{fig:embodiment_variations} shows the supported arm-hand combinations. Each hand also has a floating variant, where the wrist is directly controlled by prismatic and revolute joints. Detailed embodiment specifications are provided in the supplementary material.
\vspace{-2pt}
\subsection{Visual Variation}
\vspace{-2pt}
\label{sec:visual_variation}



DexVerse provides configurable visual variation as part of its environment specification.
Each environment uses a fixed default appearance when visual randomization is disabled.
When enabled, visual properties are sampled at reset from predefined libraries, including object materials, table materials, lighting conditions, background skyboxes, exposure, and color-temperature settings.
DexVerse also supports camera-viewpoint changes, allowing the same task to appear under different observation conditions while preserving the task objective and success conditions.
Fig.~\ref{fig:visual_variations} illustrates representative visual randomizations.

Beyond visual variation, DexVerse also supports non-visual variations, including object initial poses, task sampling ranges, proprioceptive and object-state perturbations, and dynamics parameters. Visual and non-visual variations can be enabled independently or jointly, making appearance changes explicit while preserving standard environment variability for manipulation tasks.

\subsection{Asset Sources}
\label{sec:assets}
DexVerse combines assets from research datasets, simulation libraries, public 3D repositories, and image-to-3D generation tools. Rigid and articulated objects are drawn from PartNet-Mobility~\citep{xiang2020sapien}, ManiTwin~\citep{wang2026manitwin}, NVIDIA Isaac Lab/Isaac Sim assets~\citep{mittal2025isaaclab,nvidia2026isaacsimassets}, AutoBio~\citep{autobio} and publicly available Synthesis assets~\citep{extwin2025synthesis}; when suitable objects are not available, we generate candidate meshes from reference images using Meshy~\citep{meshy2026image3d} and manually process them for simulation. For visual variation, DexVerse uses 100 HDR skyboxes from Poly Haven~\citep{polyhaven2026hdri} and table material randomization from Isaac Lab/Isaac Sim assets~\citep{nvidia2026isaacsimassets}. These assets provide diverse object geometries and scene appearances while keeping task construction consistent across the benchmark.

\vspace{-2pt}
\section{Dataset and Evaluation}
\vspace{-2pt}

\begin{wrapfigure}{rt}{0.31\linewidth}
    \centering
    \vspace{-1em}
    \includegraphics[width=\linewidth]{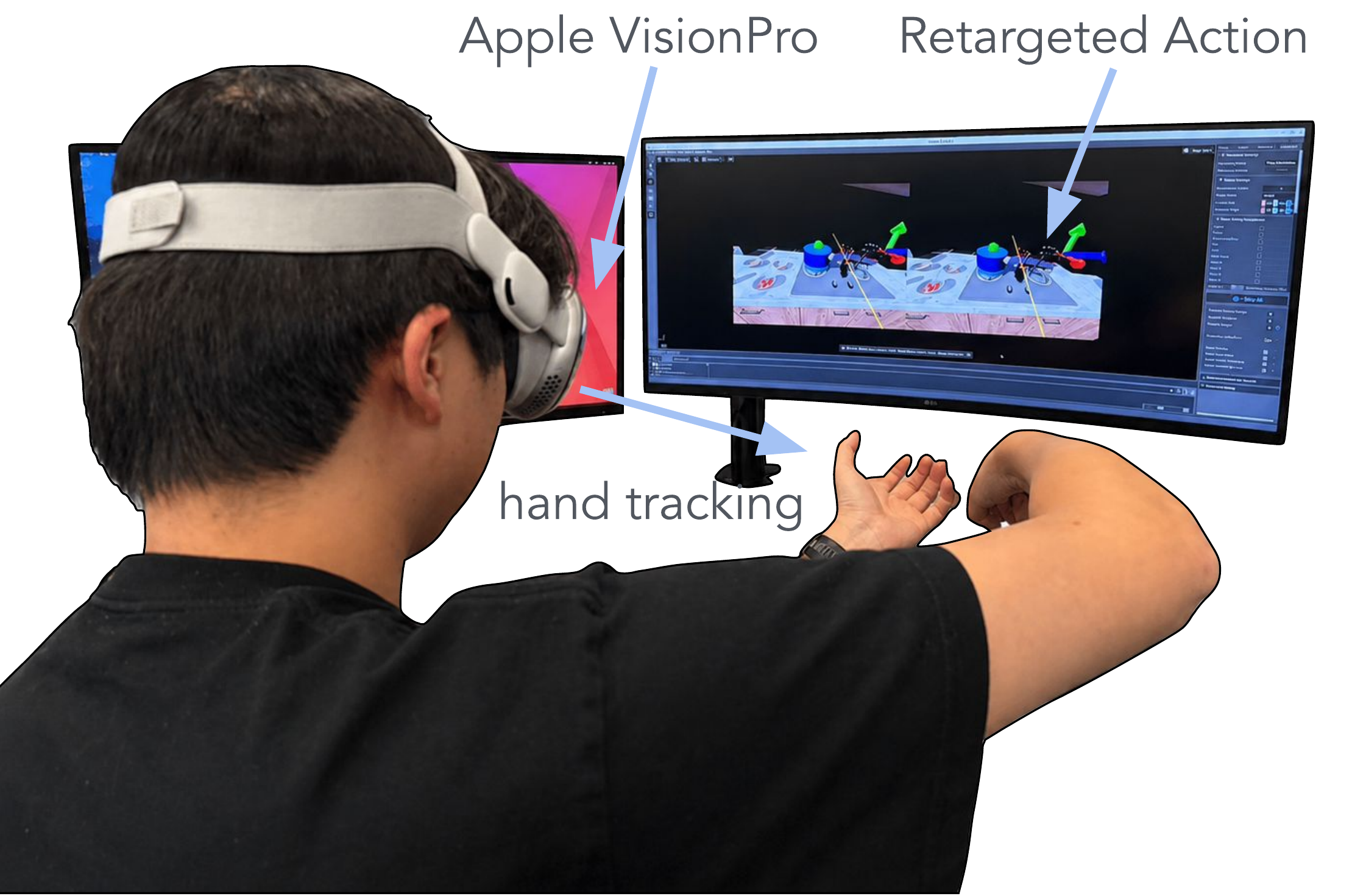}
    \caption{Teleoperation data collection system.}
    \label{fig:teleop}
    \vspace{-1em}
\end{wrapfigure}

\textbf{Teleoperation Data Collection.}
We develop an embodiment-adaptive teleoperation pipeline to scale demonstration collection in DexVerse. The system uses Apple Vision Pro through Isaac Lab’s CloudXR-based XR teleoperation interface~\cite{mittal2025isaaclab}, which streams simulation feedback to the headset and returns hand-tracking inputs for robot control. The tracked human wrist pose is used as the target pose for the robot end-effector, and the robot arm follows this target through an inverse-kinematics controller. Human hand motion is converted into target joint poses for different dexterous hands using optimization-based dex-retargeting~\cite{qin2023anyteleop} (Figure~\ref{fig:teleop}).

The pipeline is designed to reduce embodiment-specific changes when adapting to new arm-hand platforms. Adding new robot arms typically requires updating the end-effector frame, initial poses, and low-level controller parameters, while adding dexterous hands requires configuring the hand URDF with necessary keypoints, correspondence links, and retargeting scales.

\textbf{Dataset Statistics.}
DexVerse provides teleoperation demonstrations for most of the task suites. For each of the 56 single-goal tasks, we collect 55 demonstrations: 50 with the Shadow Hand and one with each of the other five hand embodiments. For each of the 5 long-horizon tasks, we collect 20 demonstrations. This results in a total of 3,180 demonstration trajectories.

Each demonstration is stored as a sequence of action-state pairs recorded during
teleoperation. We provide the replay utility that restores the recorded simulator states and queries the environment locally to regenerate the requested observation terms. This design is important because physics simulation can diverge across machines due to differences in physics computation, hardware, and floating-point rounding. Direct
state replay avoids accumulated rollout drift and makes demonstrations more portable across local setups.

The action-state format also keeps the dataset compact. The replay mechanism also makes modifying observation presets, adding observation terms, or changing camera-based inputs more flexible without requiring a separate copy of the demonstration dataset.
\vspace{-2pt}
\subsection{Imitation Learning Policy Evaluation}
\vspace{-2pt}

\begin{table}[t]
\centering
\caption{Imitation learning baselines' online success rates on DexVerse baseline tasks.
}
\vspace{3pt}
\label{tab:il-baseline-sr}
    \setlength{\tabcolsep}{10pt}
\resizebox{0.98\linewidth}{!}{
\begin{tabular}{llcccc}
\toprule
Task Characteristics & Task & Pi0.5 & OpenVLA & 3D Diffusion Policy & Diffusion Policy \\
\midrule
\multirow{7}{*}{Pick-and-Lift}
 & BimanualLiftCarton & \textbf{1.00} & 0.60 & 0.90 & 0.94 \\
 & BimanualLiftTray & \textbf{0.84} & 0.72 & 0.56 & 0.60 \\
 & GraspBleach & 0.10 & 0.06 & \textbf{0.32} & 0.10 \\
 & GraspCup & 0.16 & 0.08 & 0.22 & \textbf{0.50} \\
 & GraspKettle & 0.58 & 0.16 & 0.80 & \textbf{0.90} \\
 & GraspPan & 0.06 & 0.02 & 0.16 & \textbf{0.52} \\
 & RetrieveCup & 0.02 & 0.02 & \textbf{0.06} & 0.04 \\
\midrule
\multirow{6}{*}{Articulated}
 & OpenFaucet & \textbf{0.84} & 0.36 & 0.76 & 0.28 \\
 & OpenFlatFolder & 0.00 & 0.00 & \textbf{0.18} & 0.16 \\
 & OpenLaptop & 0.04 & 0.02 & 0.02 & \textbf{0.10} \\
 & OpenStapler & 0.86 & \textbf{0.92} & 0.84 & 0.86 \\
 & SlideUtilityKnife & 0.00 & 0.00 & 0.00 & 0.00 \\
 & SqueezeScissors & \textbf{0.36} & 0.22 & 0.20 & 0.00 \\
\midrule
\multirow{3}{*}{Tool Use}
 & FunctionalHammerStrike & 0.22 & 0.18 & \textbf{0.26} & 0.00 \\
 & FunctionalPourCan & 0.04 & 0.10 & 0.14 & \textbf{0.38} \\
 & FunctionalPourMug & 0.52 & 0.16 & \textbf{0.64} & 0.26 \\
\midrule
\multirow{3}{*}{Precision}
 & InsertPen & 0.06 & 0.00 & \textbf{0.08} & 0.00 \\
 & PushSmallSphereObstacleSlope & \textbf{0.82} & 0.08 & 0.28 & 0.36 \\
 & PushT & 0.00 & 0.00 & 0.00 & 0.00 \\
\midrule
\multicolumn{2}{l}{Mean} & \textbf{0.34} & 0.19 & \textbf{0.34} & 0.32 \\
\bottomrule
\end{tabular}}
\vspace{-18pt}
\end{table}

We evaluate four open-source imitation-learning policy families on the DexVerse baseline split: two vision-language-action (VLA) transformers fine-tuned from internet-scale pretrained backbones: $\pi_{0.5}$~\cite{intelligence2025pi} and \mbox{OpenVLA}~\cite{kim24openvla}, and two from-scratch diffusion policies: Diffusion Policy (DP)~\cite{chi2025diffusion} and 3D Diffusion Policy (DP3)~\cite{Ze2024DP3}.
All four methods are trained using the same set of 950 episodes (19 tasks × 50 episodes per task) of the DexVerse teleoperation corpus, and evaluated closed-loop in the same simulator under identical termination criteria.

\textbf{Results.}
For every task we roll out 50 episodes and report the mean success rate.
Table~\ref{tab:il-baseline-sr} summarizes the per-task results.
DP3 ties $\pi_{0.5}$ for the highest overall success rate (0.34), ahead of DP (0.32) and OpenVLA (0.19).
The aggregate ranking is close, but the per-skill profiles diverge: DP is strongest on simple Pick-and-Lift, DP3 on Tool Use where point-cloud inputs help disambiguate object geometry, and $\pi_{0.5}$ on Precision Contact, the same category where OpenVLA falls behind.
We highlight three findings.

\textbf{1) Internet-scale VLA pretraining does not yet translate into a dexterous-manipulation advantage.}
Despite being initialized from web-scale pretrained backbones, the stronger VLA ($\pi_{0.5}$, 0.34) only matches the best from-scratch policy (DP3, 0.34), while OpenVLA (0.19) trails both diffusion baselines.
We attribute this to the gap between the pretraining distribution and the target embodiment: the priors carried by these backbones come from web images and low-DoF action spaces, which transfer to perception but not to the high-DoF multifinger control manifold of DexVerse.

\textbf{2) The most informative observation modality is skill-dependent, and no single representation dominates.}
DP is strongest on Pick-and-Lift (0.51), where a successful grasp pose is largely a function of object appearance, and a 2D image plus low-dimensional state already suffices.
3D DP leads on Functional Tool Use (0.35), where explicit point-cloud geometry helps localize the tool tip and regulate the pour/strike pose.
$\pi_{0.5}$ leads on both Articulated-object Manipulation (0.35) and Precision Contact (0.29), where language conditioning and a flow-matching action expert help disambiguate multi-stage subgoals and contact timing.
This spread, a different method wins each of the four skill families, motivates DexVerse's multi-modal observation interface rather than committing to a single sensing paradigm.

\textbf{3) Fine contact reasoning and sub-centimeter alignment remain unsolved across the board.}
Tight-tolerance tasks collapse for every method: PushT is 0.00 for all four policies, and InsertPen, SlideUtilityKnife, and OpenLaptop stay at or near zero everywhere.
These tasks demand sustained force regulation and sub-centimeter alignment that behavior cloning without explicit force feedback or closed-loop contact correction cannot yet provide, and they constitute the principal headroom that DexVerse exposes for future imitation-learning research.

\vspace{-2pt}
\section{Conclusion}
\vspace{-2pt}

We presented DexVerse, a modular benchmark for multi-task, multi-embodiment dexterous manipulation that unifies diverse task categories, multiple arm-hand embodiments, configurable visual variation, VR-based teleoperation, multi-modal demonstrations, and representative policy evaluation. Our experiments show that current imitation-learning and vision-language-action policies remain far from solving general dexterous manipulation, particularly for precise contact, bimanual coordination, functional tool use, and robust interaction with complex objects. These findings establish DexVerse as a challenging and extensible testbed for studying contact-rich dexterous control, visuomotor robustness, and embodiment-aware robot learning.

\textbf{Limitation and Future Work.} The current release focuses on building a broad, reproducible, and multi-task, multi-embodiment benchmark. Future extensions will study real-robot transfer, expand demonstrations across more embodiments and task families, and provide broader standardization for cross-task and cross-embodiment evaluation. These directions will further strengthen DexVerse as a foundation for developing general, robust, and transferable dexterous manipulation policies.

\bibliography{reference}  

\clearpage
\onecolumn

\appendix

\section*{Appendix}

\section{Task List and Visualization}
\begingroup
\scriptsize
\setlength{\tabcolsep}{3pt}
\renewcommand{\arraystretch}{1.15}

\newcommand{\categorycell}[2]{%
    \multirow[c]{#1}{*}{%
        \rotatebox[origin=c]{90}{%
            \normalsize\bfseries\MakeUppercase{#2}%
        }%
    }%
}

\newcommand{\singlecategorycell}[1]{%
    \rotatebox[origin=c]{90}{%
        \normalsize\bfseries\MakeUppercase{#1}%
    }%
}

\begin{longtable}{
    @{}
    >{\centering\arraybackslash}m{0.10\textwidth}
    >{\raggedright\arraybackslash}m{0.17\textwidth}
    >{\raggedright\arraybackslash}m{0.18\textwidth}
    >{\raggedright\arraybackslash}m{0.32\textwidth}
    >{\centering\arraybackslash}m{0.17\textwidth}
    @{}
}
    \caption{
        Task list with category, task name, task description, success condition, and environment rendering.
    }
    \label{tab:task_list} \\

    \toprule
    \multicolumn{1}{c}{Category}
    & \multicolumn{1}{c}{\texttt{task\_name}}
    & \multicolumn{1}{c}{Description}
    & \multicolumn{1}{c}{Success condition}
    & \multicolumn{1}{c}{rendering} \\
    \midrule
    \endfirsthead

    \toprule
    \multicolumn{1}{c}{Category}
    & \multicolumn{1}{c}{\texttt{task\_name}}
    & \multicolumn{1}{c}{Description}
    & \multicolumn{1}{c}{Success condition}
    & \multicolumn{1}{c}{rendering} \\
    \midrule
    \endhead

    \midrule
    \multicolumn{5}{r}{Continued on next page} \\
    \endfoot

    \bottomrule
    \endlastfoot

    \addlinespace
    \categorycell{6}{Primitive}
    & \texttt{PickCube}
    & Grasp a cube on the tabletop and lift it away from the surface.
    & The cube is lifted at least 0.20 m above its resetting height.
    & \includegraphics[width=0.95\linewidth]{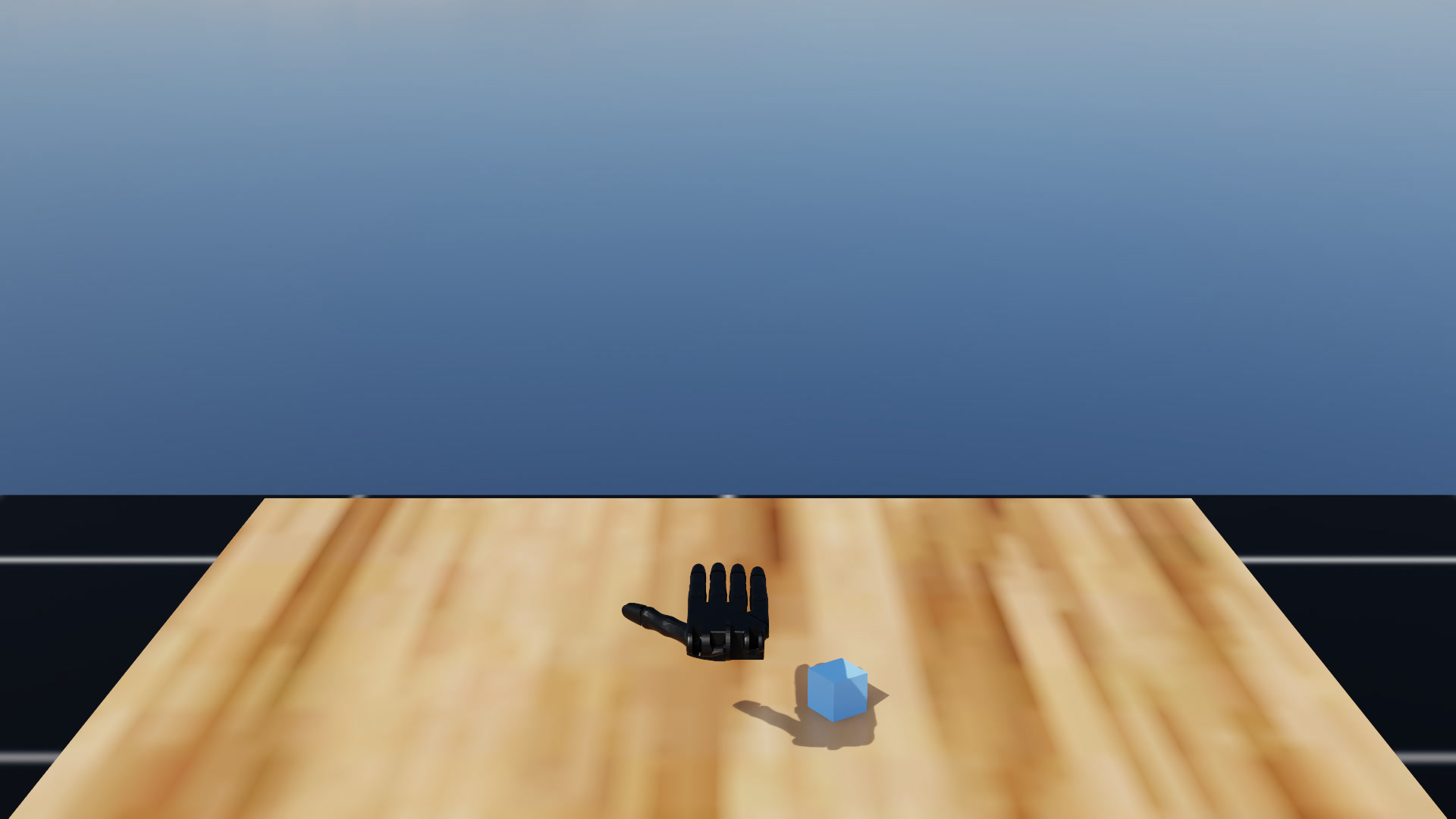} \\

    \addlinespace

    & \texttt{StackCube}
    & Pick and place one cube on top of another cube to form a stable stack.
    & The moving cube is stabilized within 0.035 m horizontally and 0.025 m vertically of resting directly on top of the base cube.
    & \includegraphics[width=0.95\linewidth]{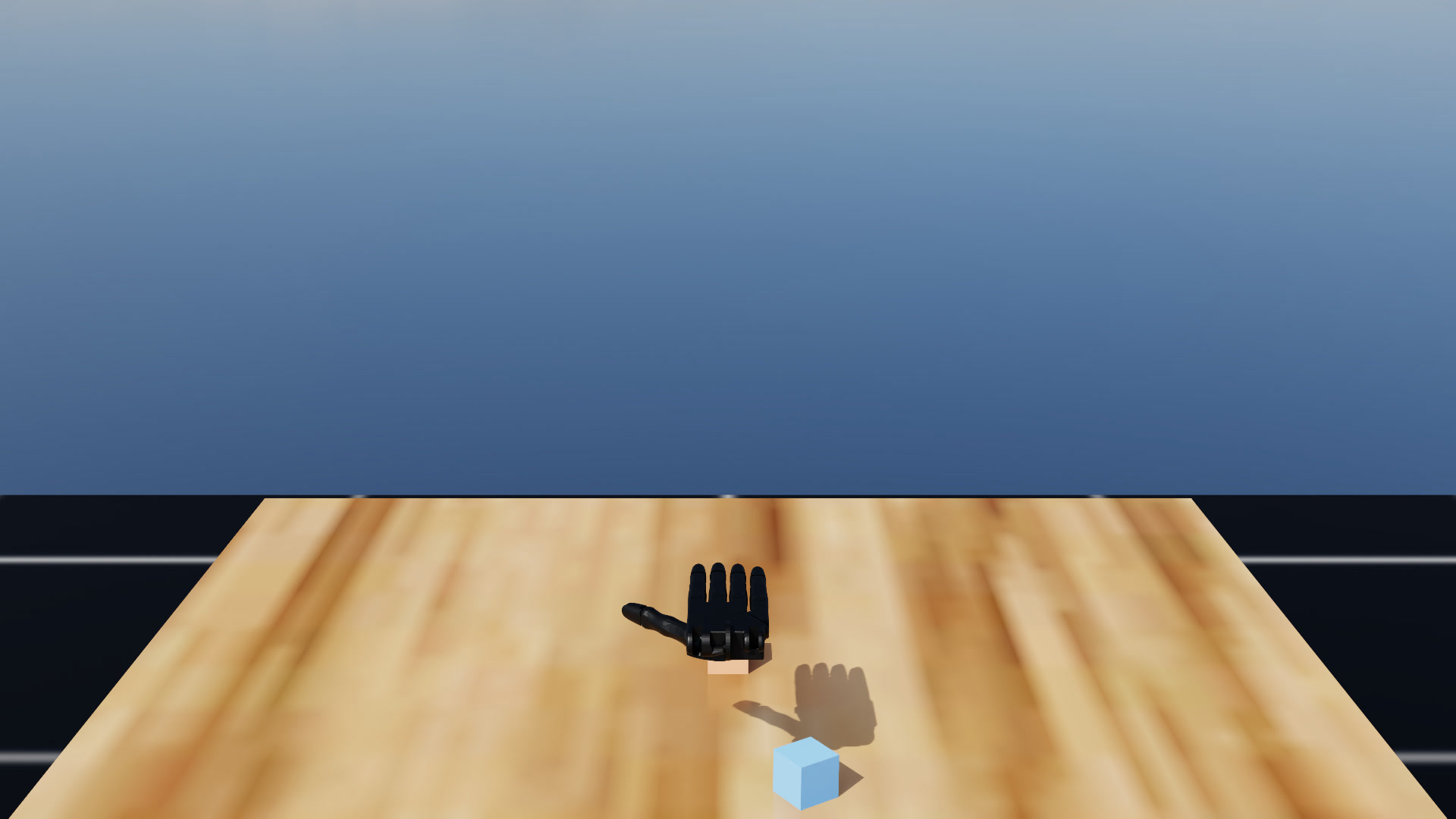} \\

    \addlinespace

    & \texttt{RelocateSphere}
    & Move a sphere from its initial tabletop position to the commanded target location.
    & The sphere comes within 0.03 m of a goal point sampled once per episode.
    & \includegraphics[width=0.95\linewidth]{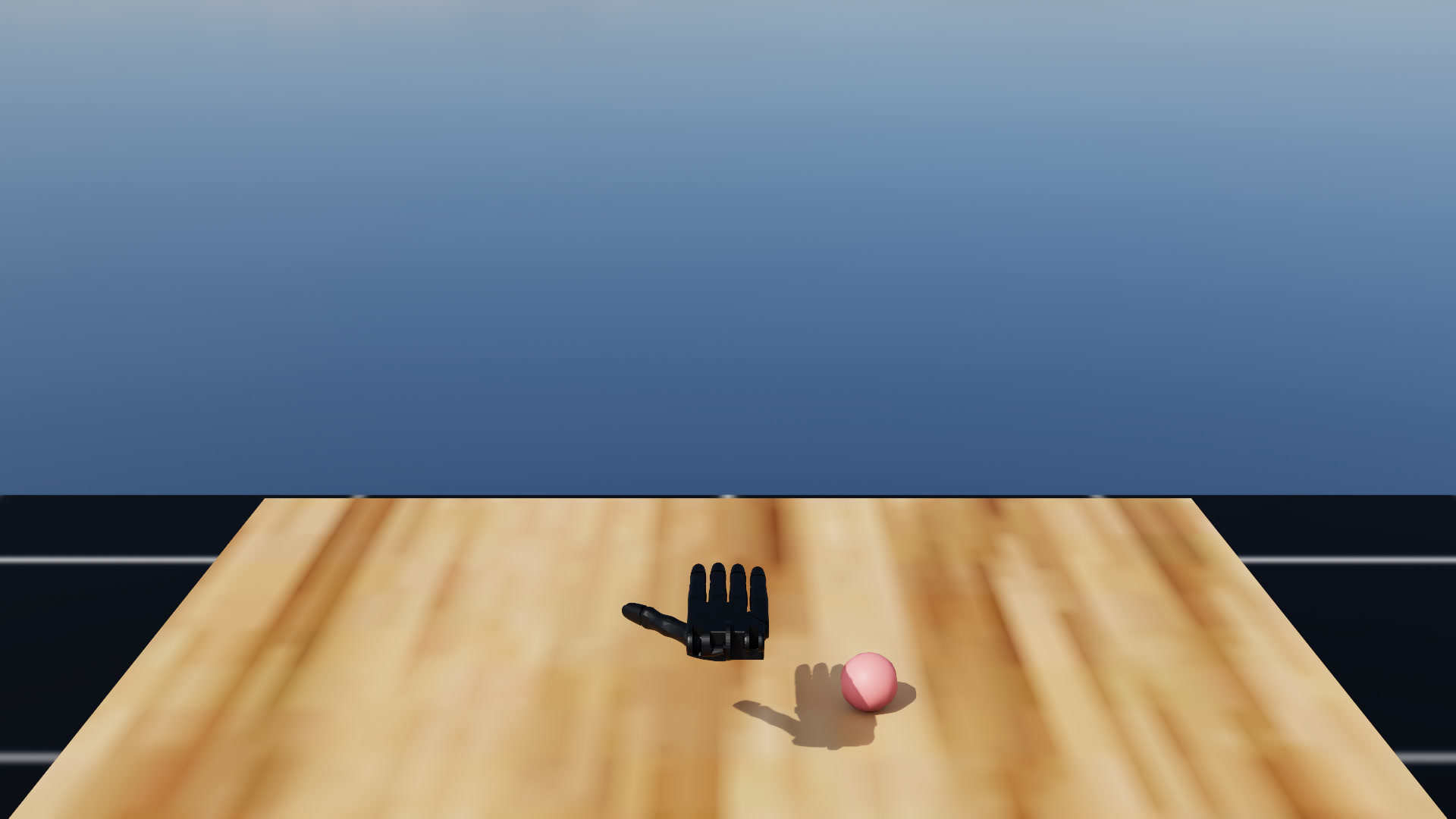} \\

    \addlinespace

    & \texttt{PickUpStick}
    & Grasp a stick on the tabletop and lift it while maintaining an upright orientation.
    & The stick is lifted at least 0.20 m above its reset height and stays within 30$^\circ$ of vertical.
    & \includegraphics[width=0.95\linewidth]{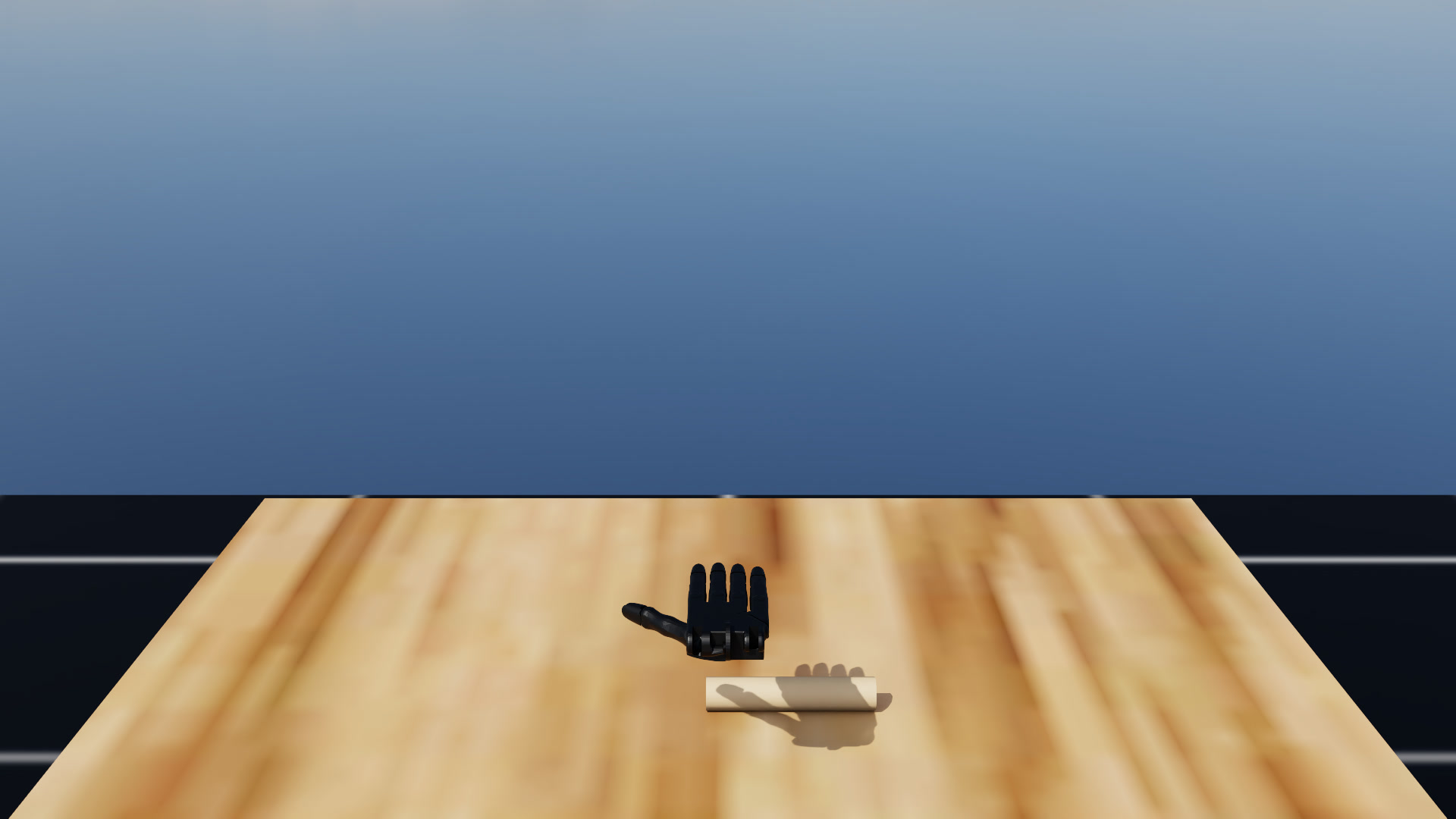} \\

    \addlinespace

    & \texttt{RelocateObject}
    & Move a rigid object from the tabletop reset region to a sampled 3D goal position above the table.
    & The object is moved to within 0.03 m of a sampled goal point, roughly within 0.2 m of table center and 0.15-0.25 m above the table.
    & \includegraphics[width=0.95\linewidth]{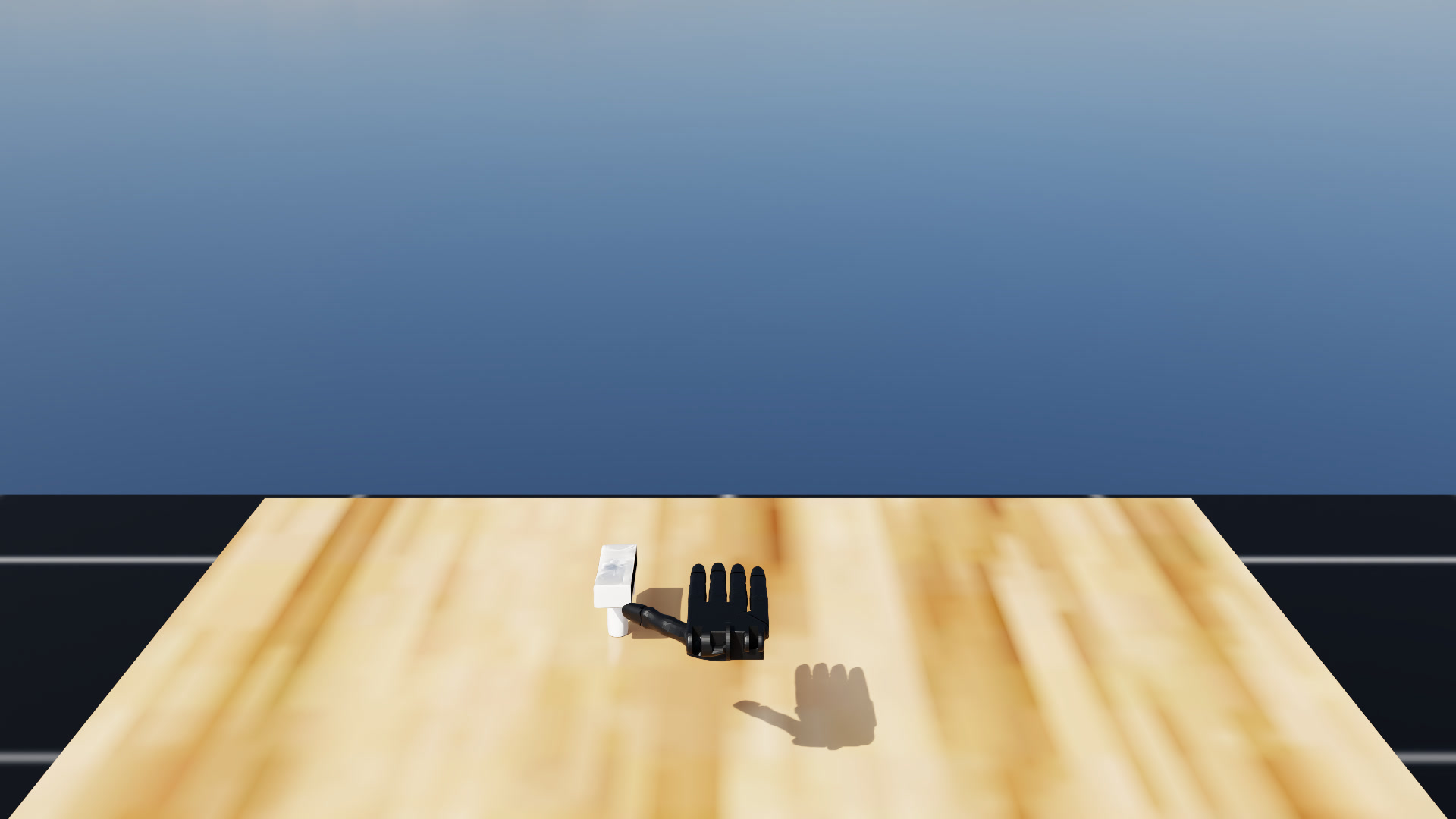} \\

    \addlinespace

    & \texttt{TurnOnSwitch}
    & Flip an articulated switch from the off state toward the on state.
    & The switch moves at least 80\% of its reachable travel from its reset position.
    & \includegraphics[width=0.95\linewidth]{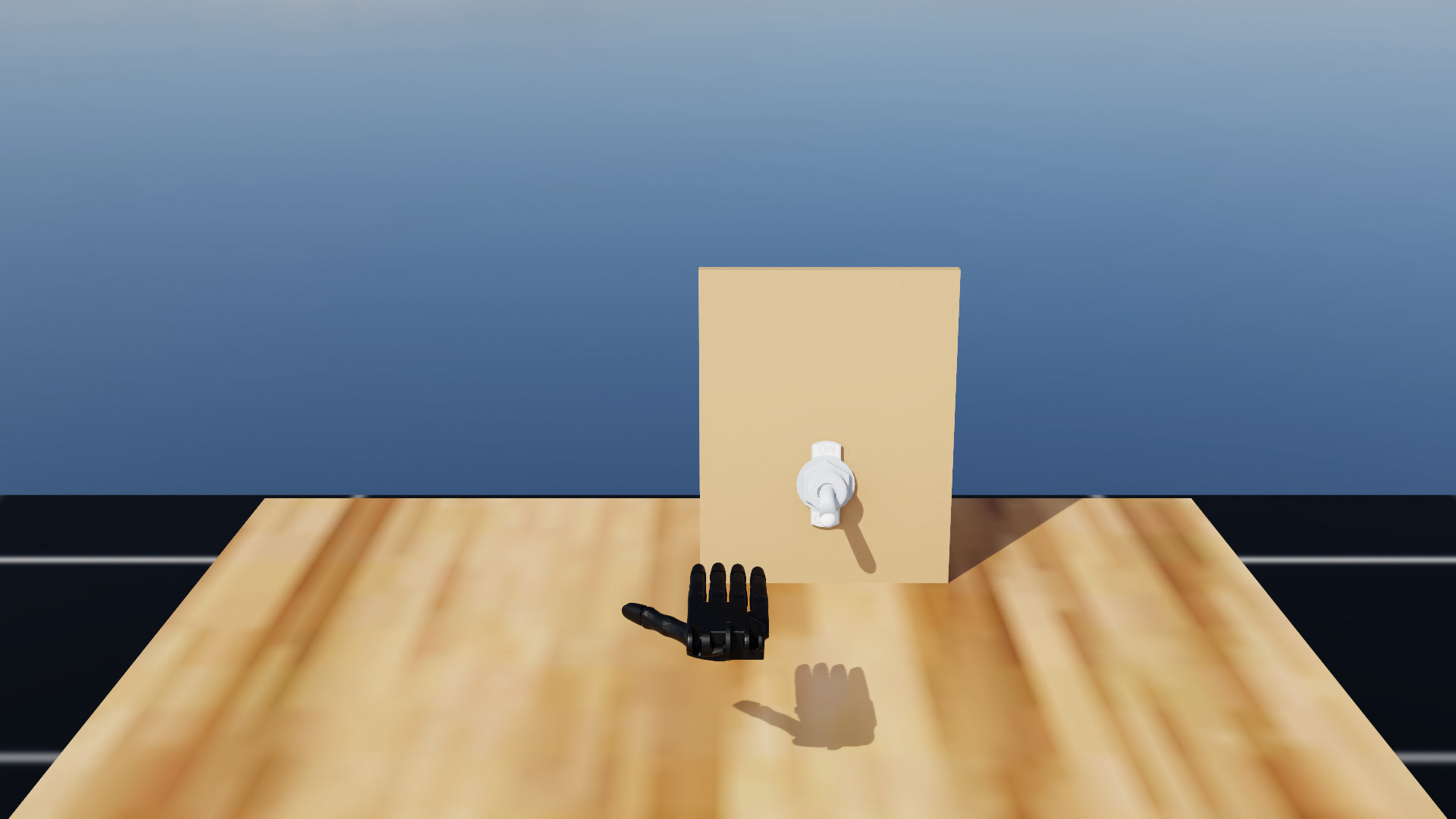} \\

    \addlinespace
    & \texttt{PushButton}
    & Move the fingertip or hand into an articulated button and press it down.
    & The button is pressed at least 80\% of its roughly 0.015 m travel.
    & \includegraphics[width=0.95\linewidth]{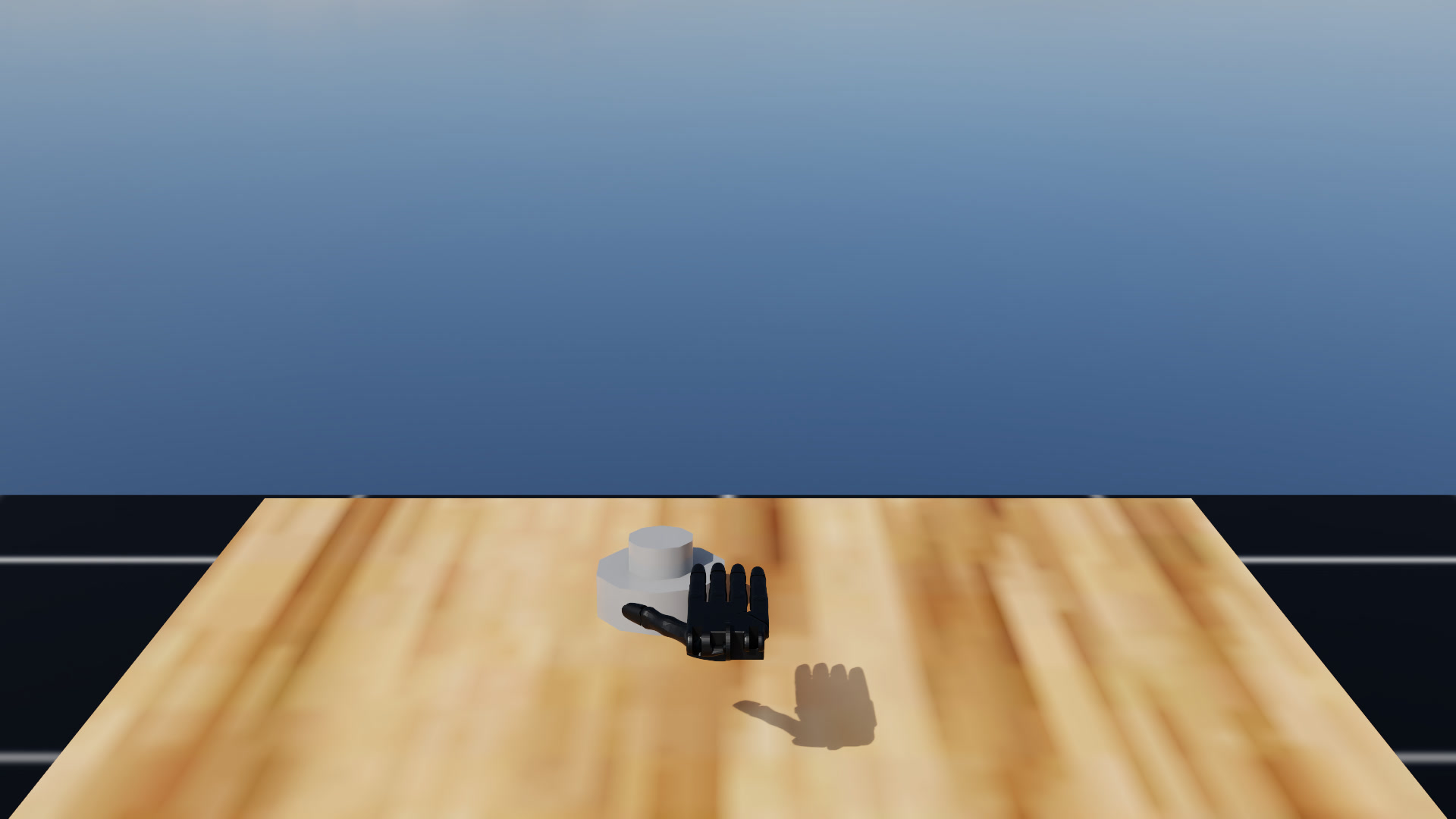} \\

    \addlinespace

    & \texttt{OpenFaucet}
    & Turn an articulated faucet handle from its initial pose toward the target turned state.
    & The handle is turned at least 80$^\circ$ from its reset angle.
    & \includegraphics[width=0.95\linewidth]{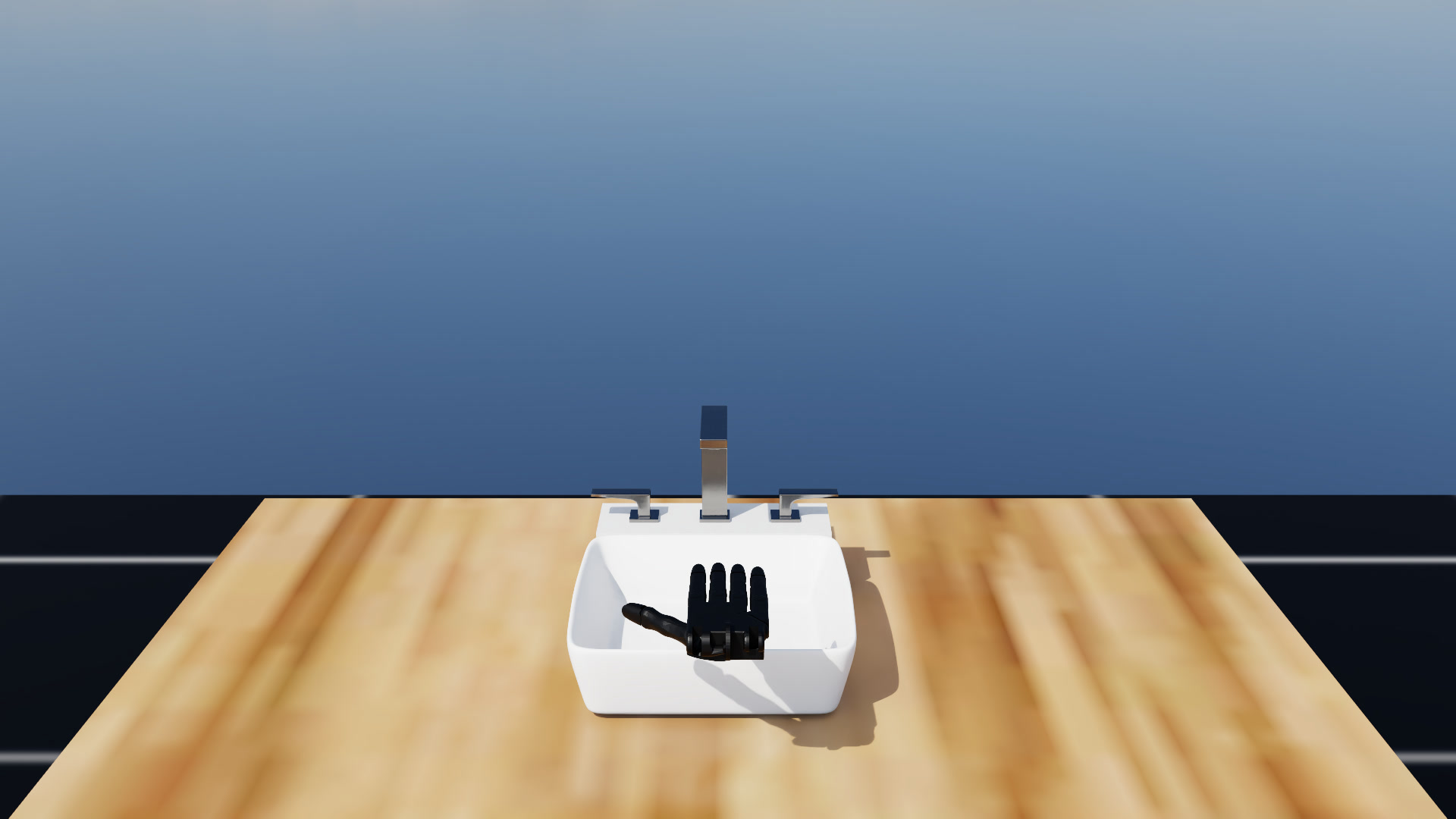} \\

    \addlinespace

    & \shortstack[l]{\texttt{GraspTwo}\\\texttt{Items}}
    & Randomly select two primitive objects from the candidate set and lift both selected objects.
    & Both of the two randomly selected primitive shapes, out of 5 candidates, are lifted at least 0.20 m above the tabletop at the same time.
    & \includegraphics[width=0.95\linewidth]{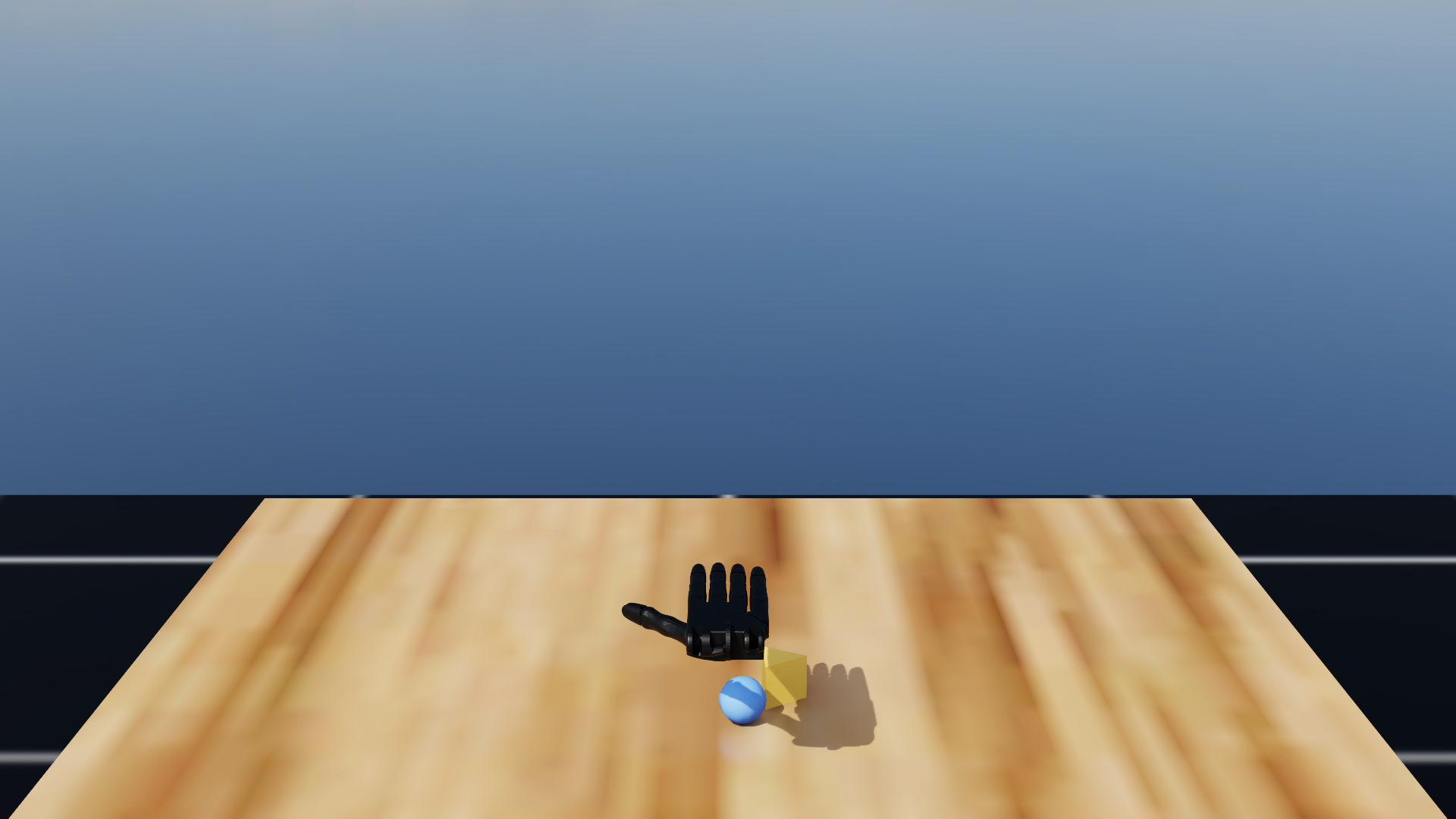} \\

    \addlinespace[4pt]
    \midrule
    \addlinespace[2pt]

    \addlinespace
    \categorycell{5}{Articulation}
    & \texttt{OpenCabinet}
    & Open the cabinet by actuating target joint \texttt{joint\_11} from its reset pose.
    & The target cabinet joint moves at least 80\% of the way toward its limit from its reset pose.
    & \includegraphics[width=0.95\linewidth]{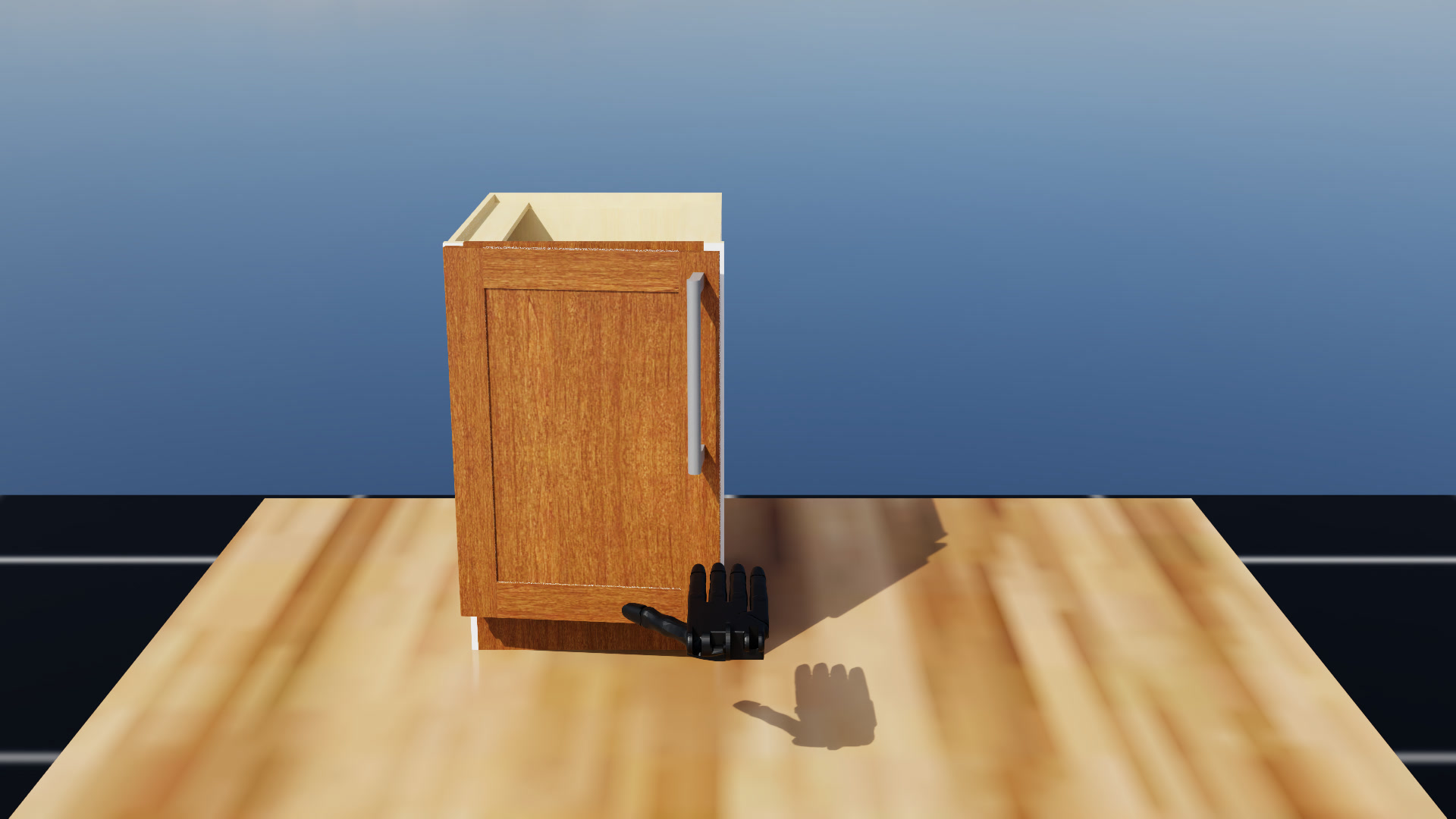} \\

    \addlinespace

    & \texttt{LiftLid}
    & Interact with an articulated object and lift its lid to the open state.
    & The lid hinge reaches at least 80\% of its full opening range.
    & \includegraphics[width=0.95\linewidth]{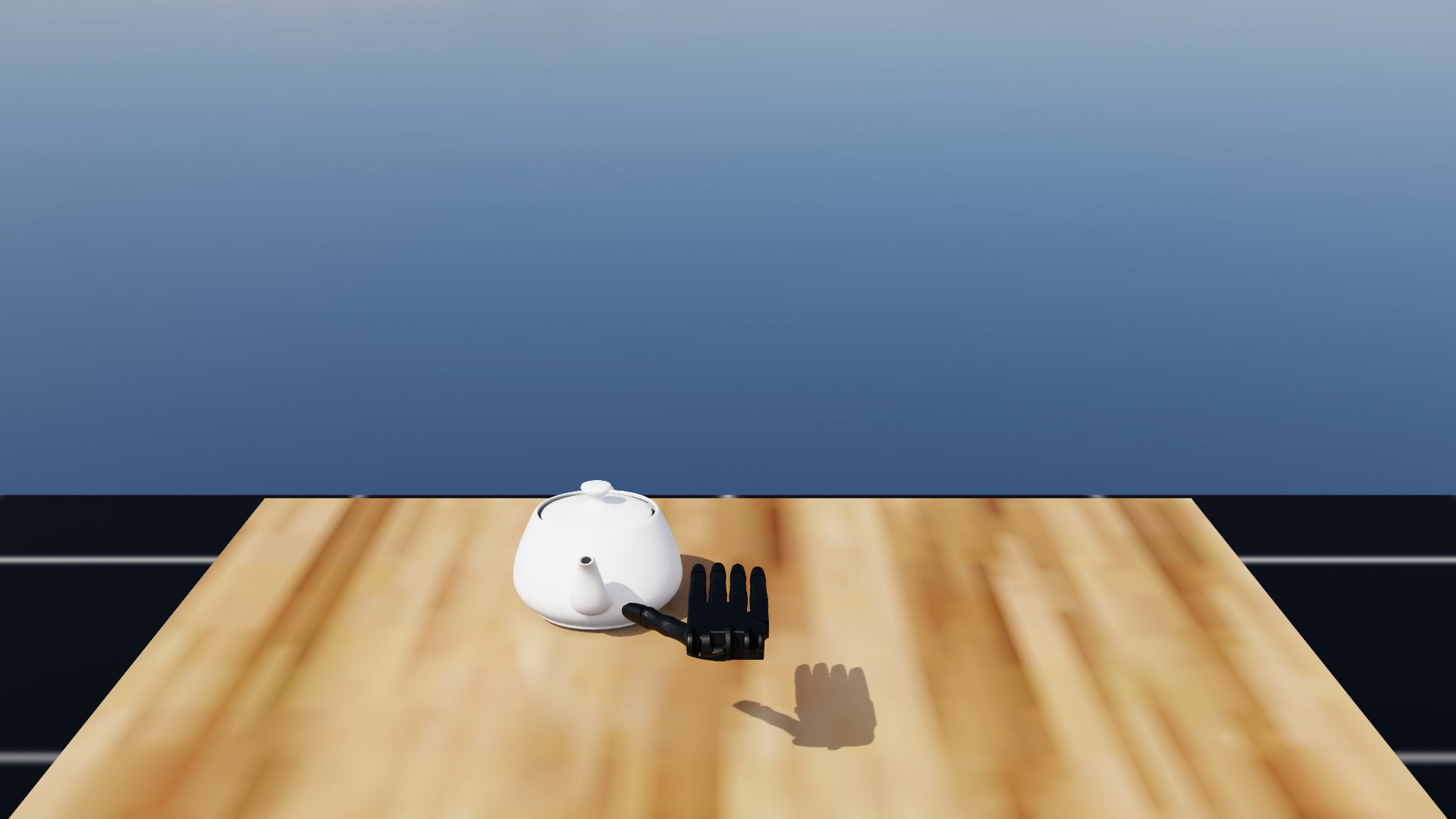} \\

    \addlinespace

    & \texttt{OpenDoor}
    & Grasp the handle, then rotate to unblock the door. Then pull the articulated door until it reaches the open state.
    & The door hinge reaches at least 80\% of its full swing range.
    & \includegraphics[width=0.95\linewidth]{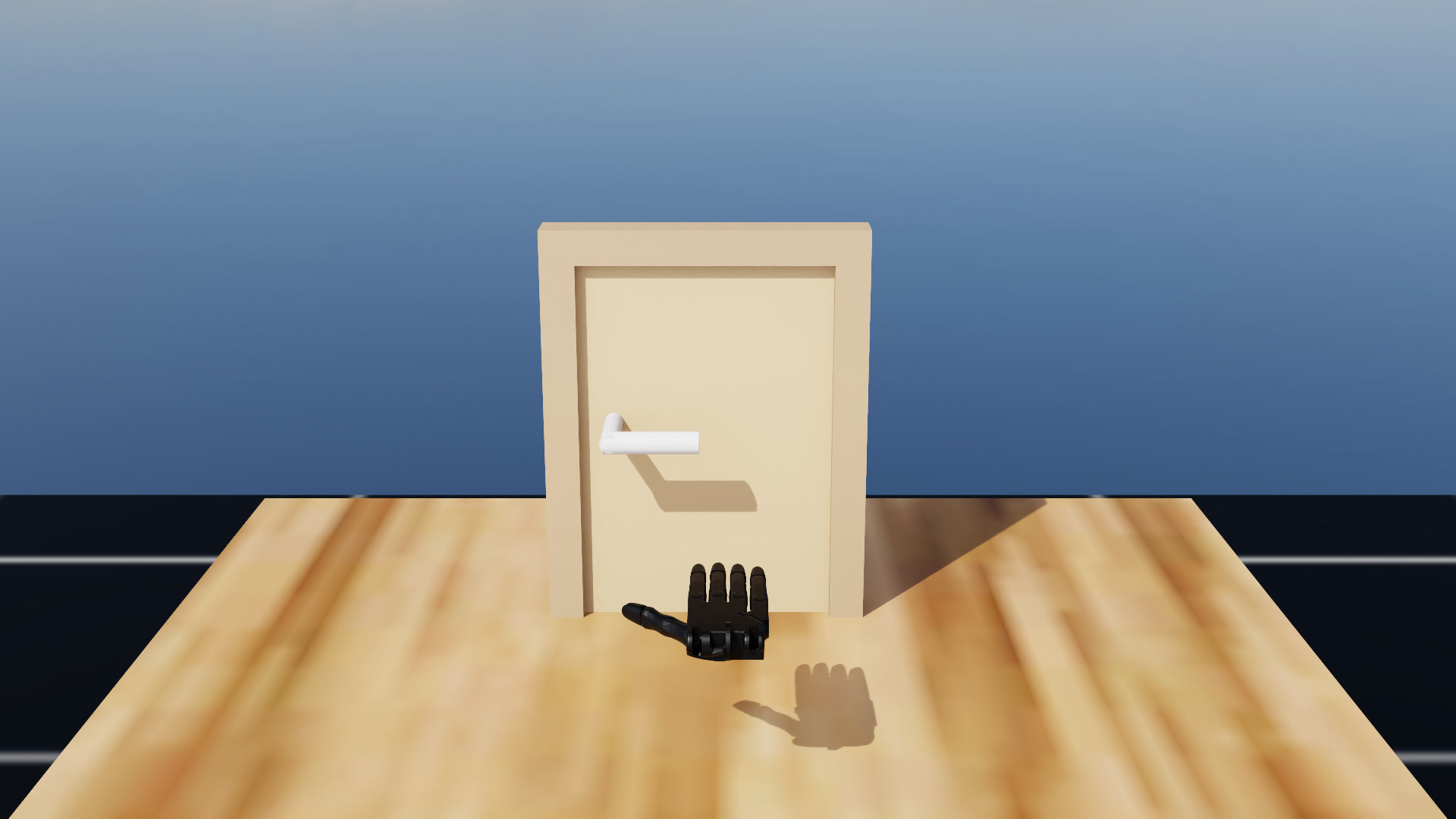} \\

    \addlinespace

    & \texttt{RotateKnob}
    & Grasp or contact an articulated knob and rotate it to the target setting.
    & The knob is turned to either end of its 180$^\circ$ joint limits.
    & \includegraphics[width=0.95\linewidth]{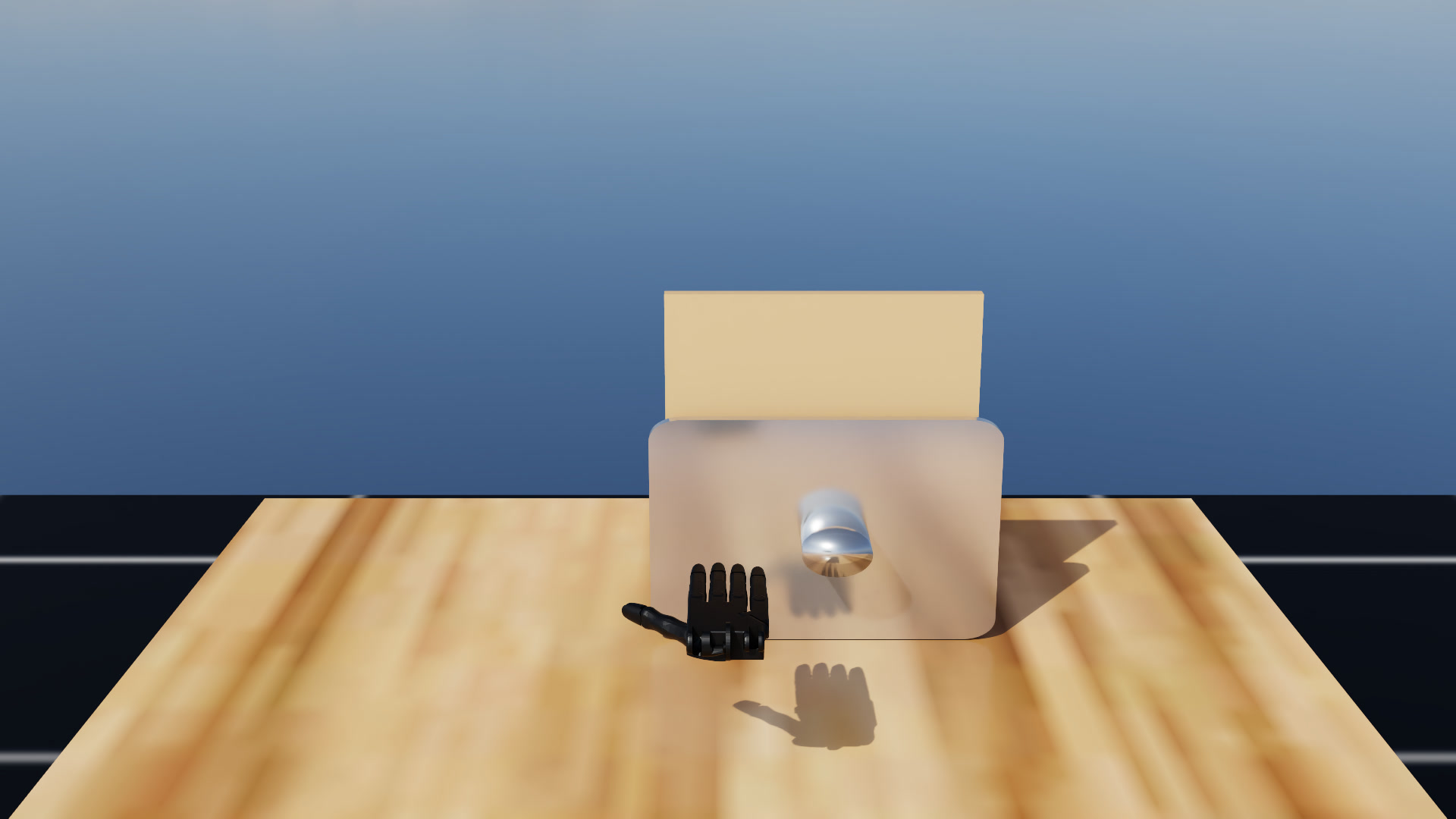} \\

    \addlinespace
    & \texttt{OpenMicrowave}
    & Grasp and pull an articulated microwave door until it reaches the open state.
    & The door hinge reaches at least 80\% of its full opening range.
    & \includegraphics[width=0.95\linewidth]{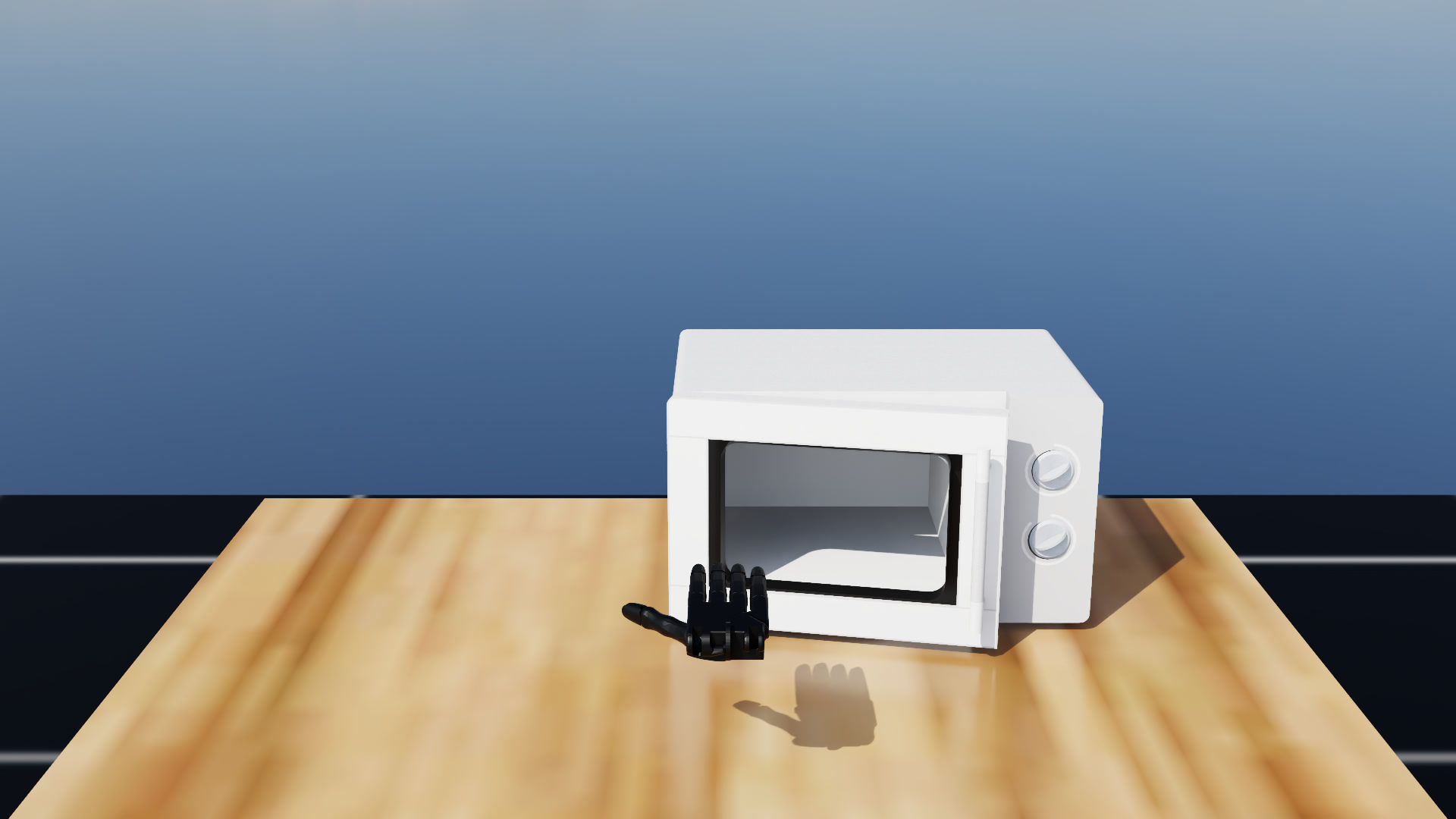} \\

    \addlinespace

    & \texttt{OpenDrawer}
    & Grasp and pull an articulated drawer outward from the cabinet.
    & The drawer slides out at least 80\% of its full travel.
    & \includegraphics[width=0.95\linewidth]{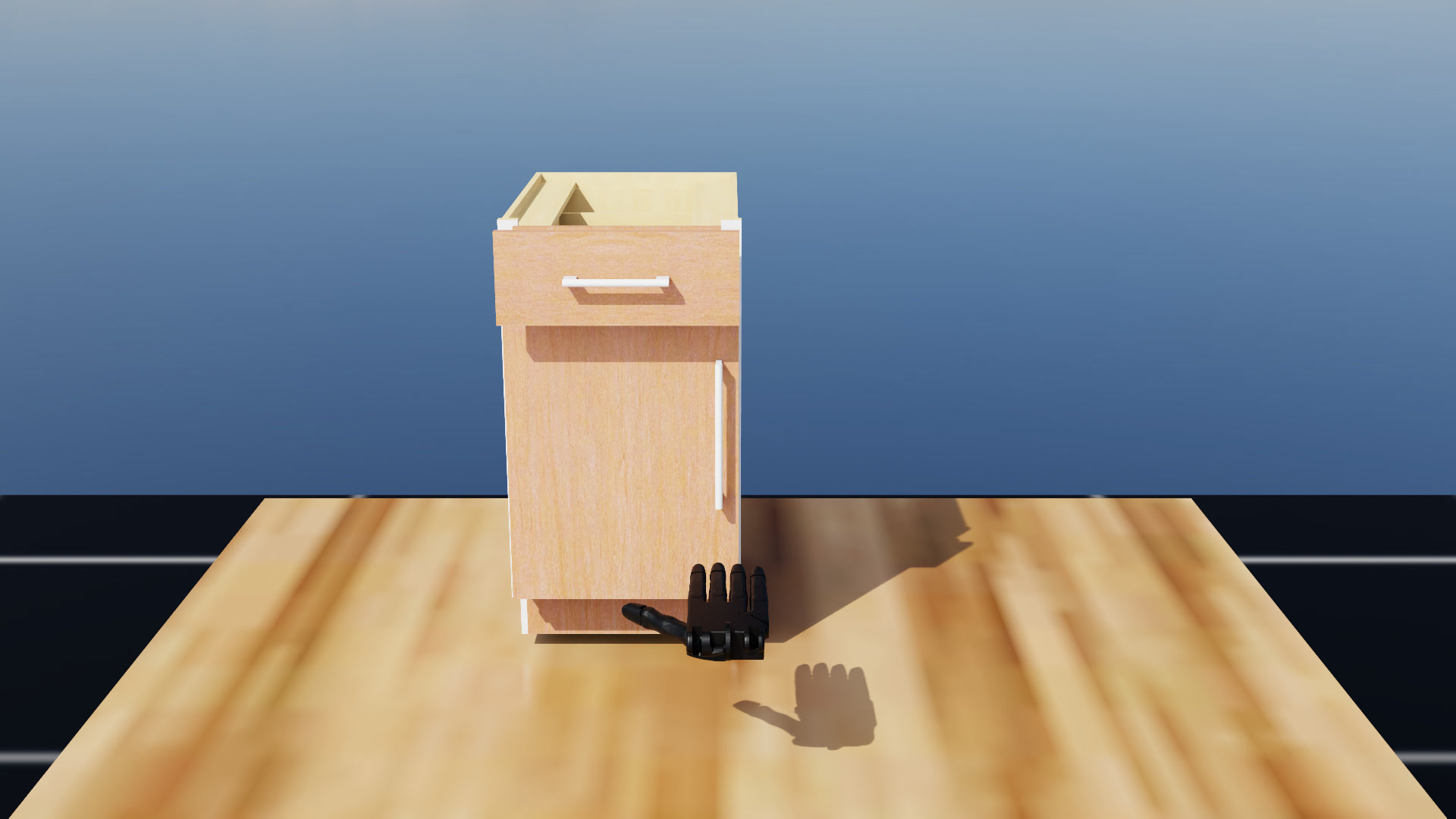} \\

    \addlinespace

    & \texttt{GraspBucket}
    & Reach for an articulated bucket and raise its handle from the resting pose.
    & The handle hinge rotates at least 40\% of its reachable range from its near-limit reset pose.
    & \includegraphics[width=0.95\linewidth]{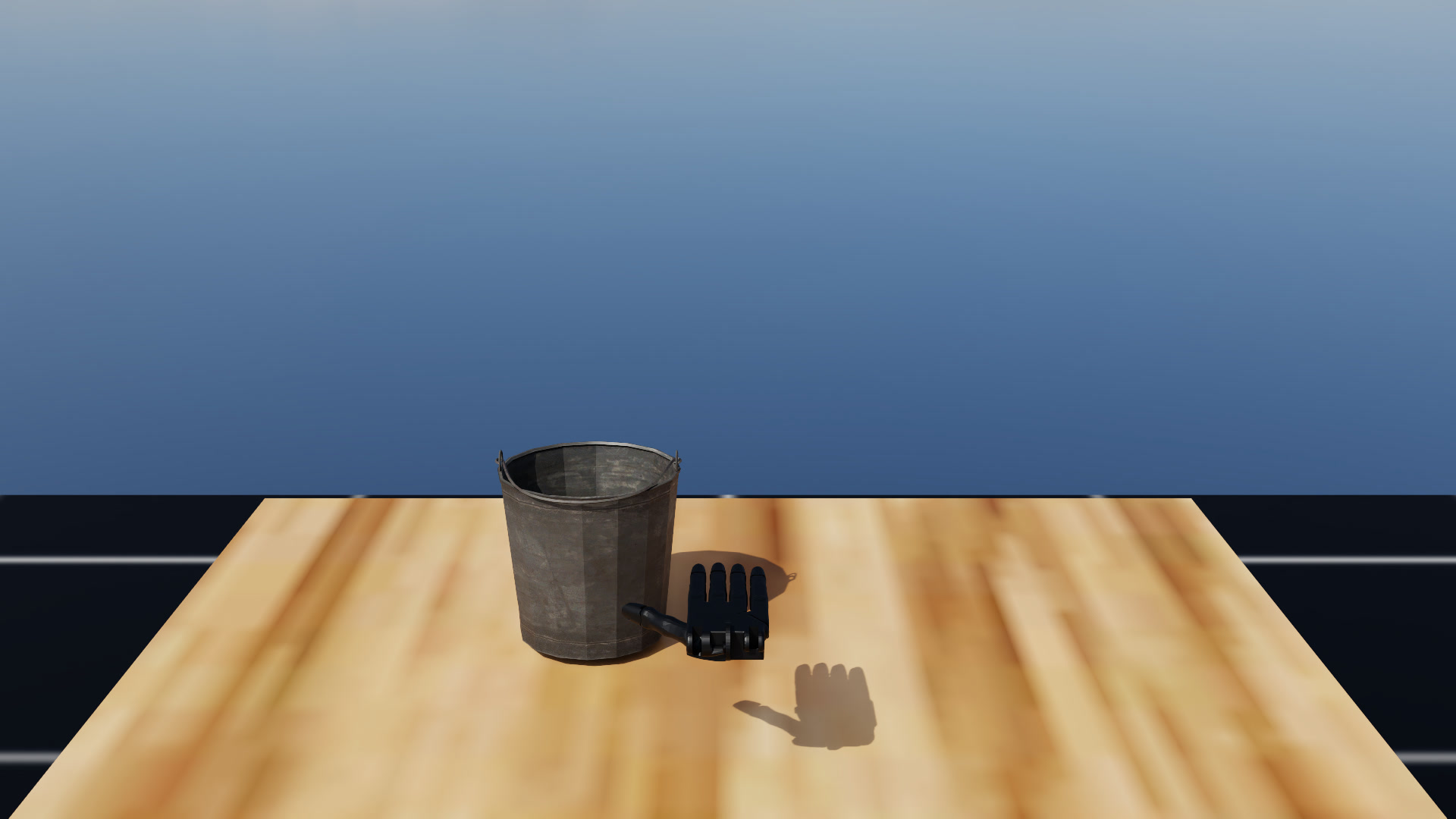} \\

    \addlinespace

    & \texttt{GraspPot}
    & Manipulate the articulated pot lid until its lid joint reaches the intended state.
    & The lid hinge reaches at least 80\% of its full opening range.
    & \includegraphics[width=0.95\linewidth]{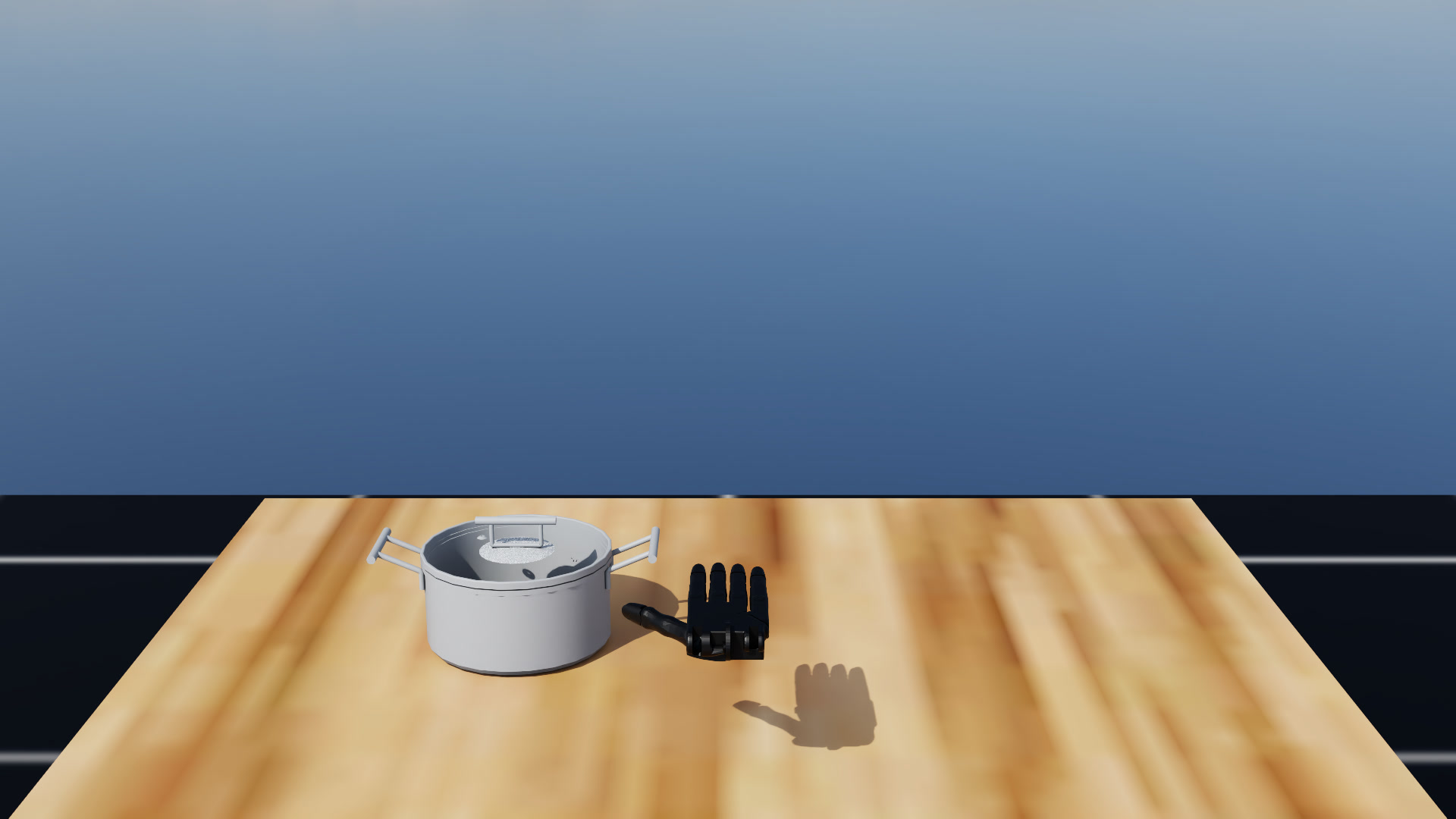} \\

    \addlinespace

    & \texttt{OpenLaptop}
    & Hold the laptop base and open the laptop lid around its hinge.
    & The lid swings at least 70\% of the way from its slightly-open reset angle near $-15^\circ$ toward the fully open limit near $-110^\circ$.
    & \includegraphics[width=0.95\linewidth]{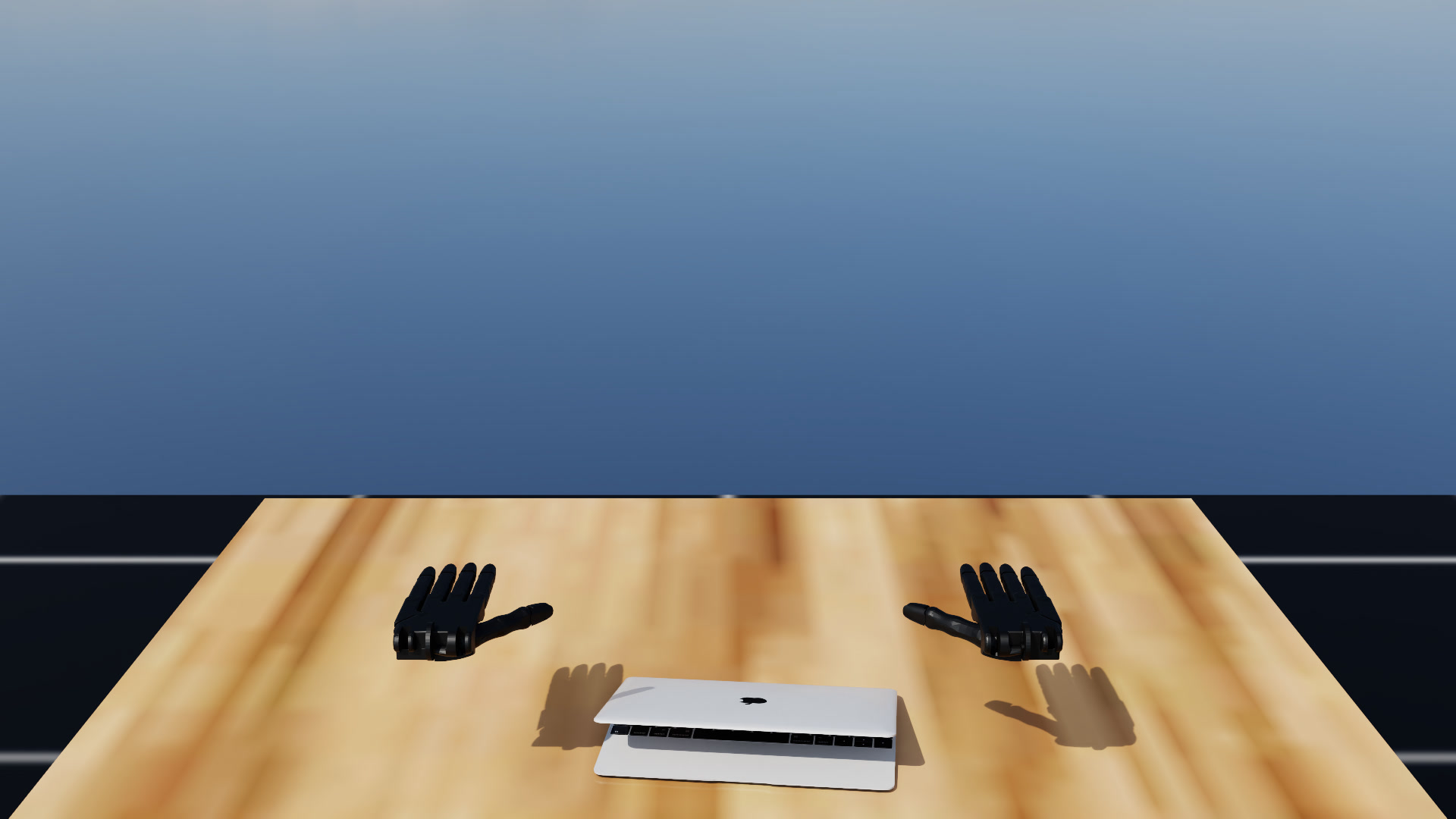} \\

    \addlinespace

    & \texttt{SqueezeScissors}
    & Hold and squeeze a pair of scissors so the hinge moves toward the closed state.
    & Either blade hinge closes at least 50\% of the way from its reset position.
    & \includegraphics[width=0.95\linewidth]{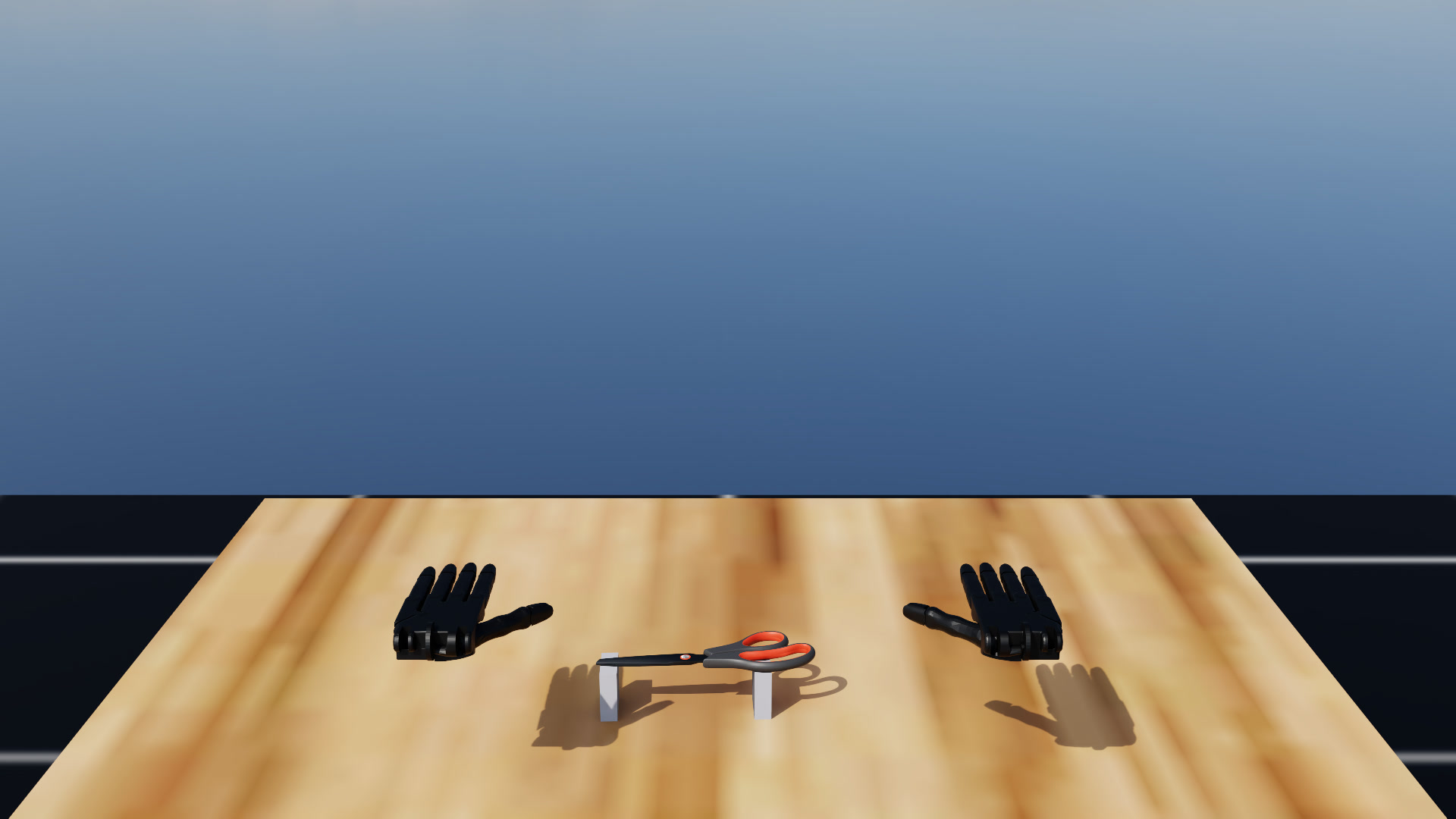} \\

    \addlinespace

    & \shortstack[l]{\texttt{SlideUtility}\\\texttt{Knife}}
    & Hold a utility knife body and slide its blade out from the handle.
    & The blade slides out at least 40\% of its travel from the retracted reset pose, and the whole knife is lifted at least 0.2 m off its support.
    & \includegraphics[width=0.95\linewidth]{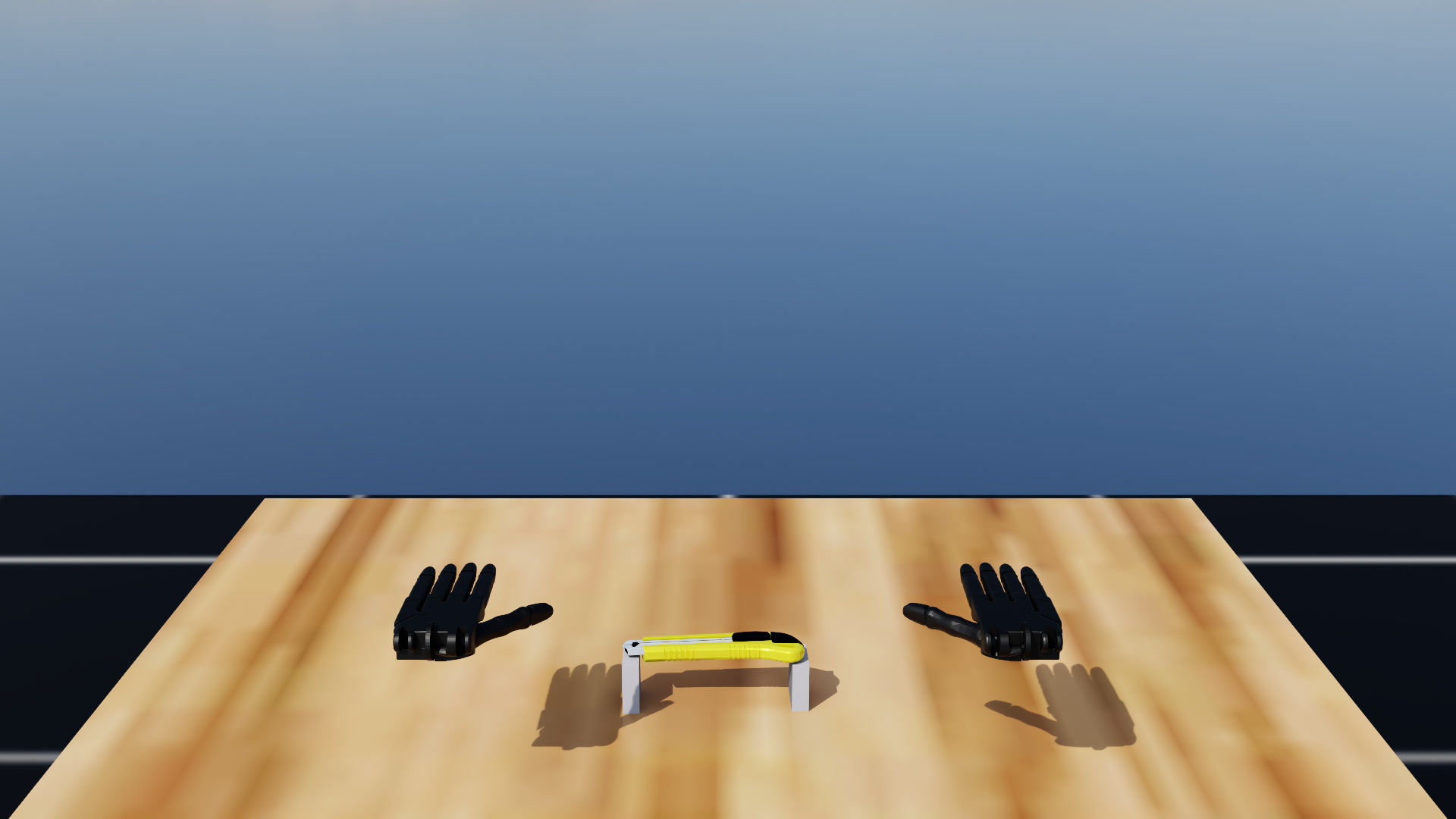} \\

    \addlinespace

    & \shortstack[l]{\texttt{LiftBasket}\\\texttt{Handle}}
    & Hold a shopping basket body and raise one of its articulated handles.
    & Either handle hinge swings at least 50\% of the way from its folded reset position.
    & \includegraphics[width=0.95\linewidth]{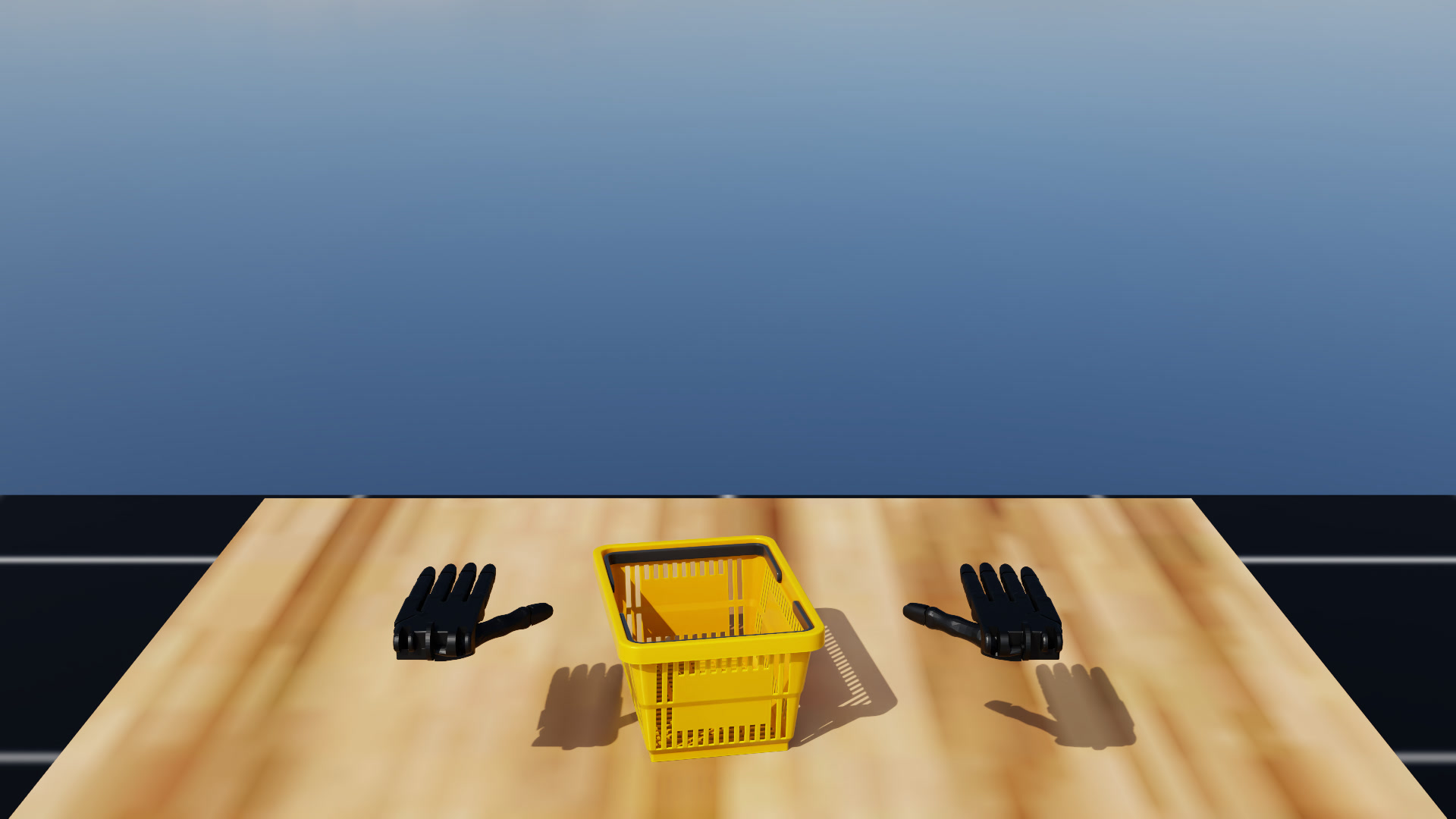} \\

    \addlinespace

    & \texttt{OpenStapler}
    & Hold a stapler body and open the upper arm around its hinge.
    & The shell hinge opens at least 20\% of the way from its closed reset position.
    & \includegraphics[width=0.95\linewidth]{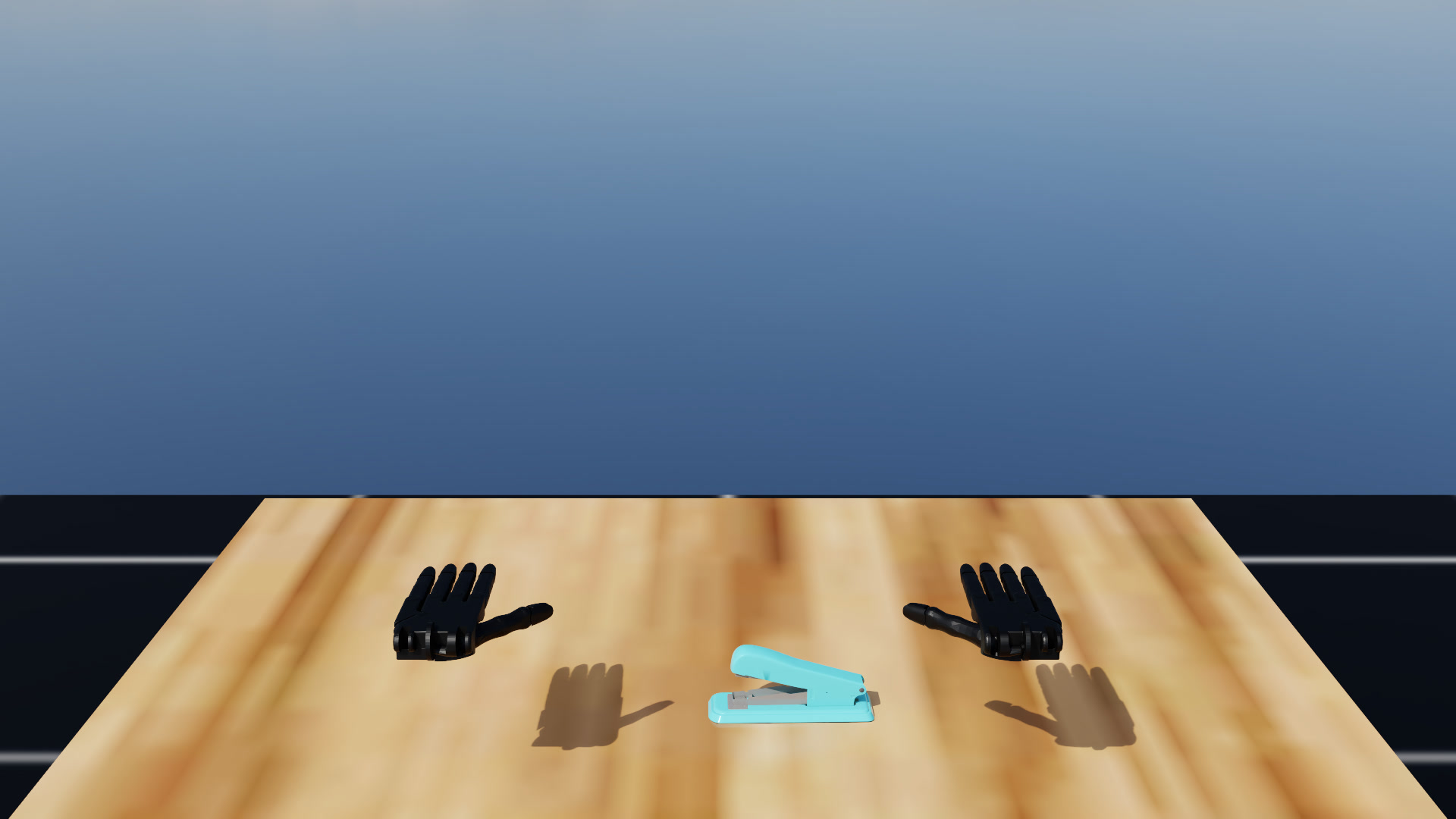} \\

    \addlinespace
    & \texttt{OpenFlatFolder}
    & Hold a flat folder and open its top flap around the hinge.
    & The flap hinge opens at least 60\% of the way from its near-closed reset position.
    & \includegraphics[width=0.95\linewidth]{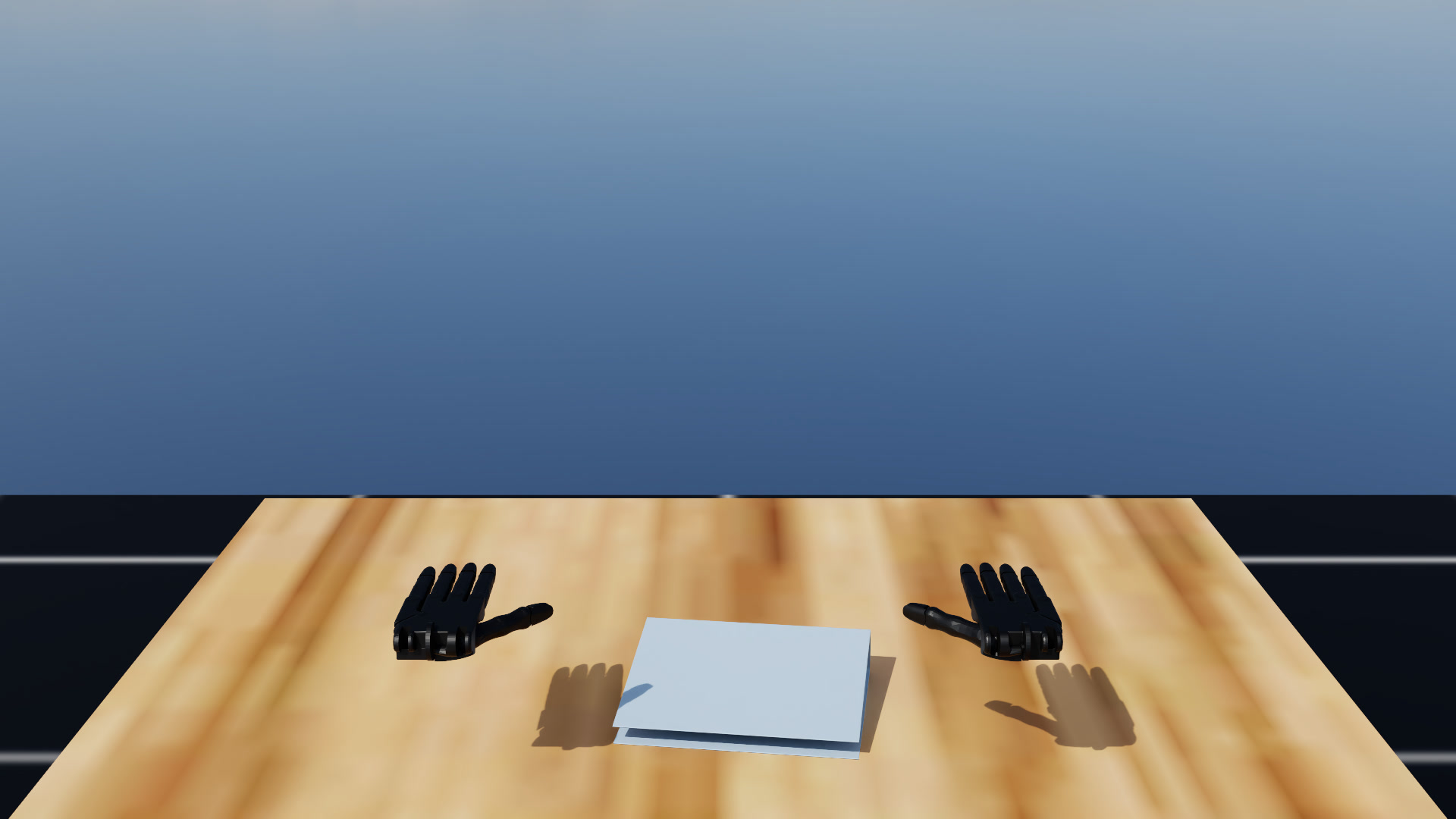} \\

    \addlinespace

    & \shortstack[l]{\texttt{Open}\\\texttt{Phone}}
    & Lift and unfold a foldable phone until its hinges open to the target state.
    & Both screen hinges independently open at least 50\% of the way from their folded reset position, and the phone is lifted at least 0.1 m above its spawn height.
    & \includegraphics[width=0.95\linewidth]{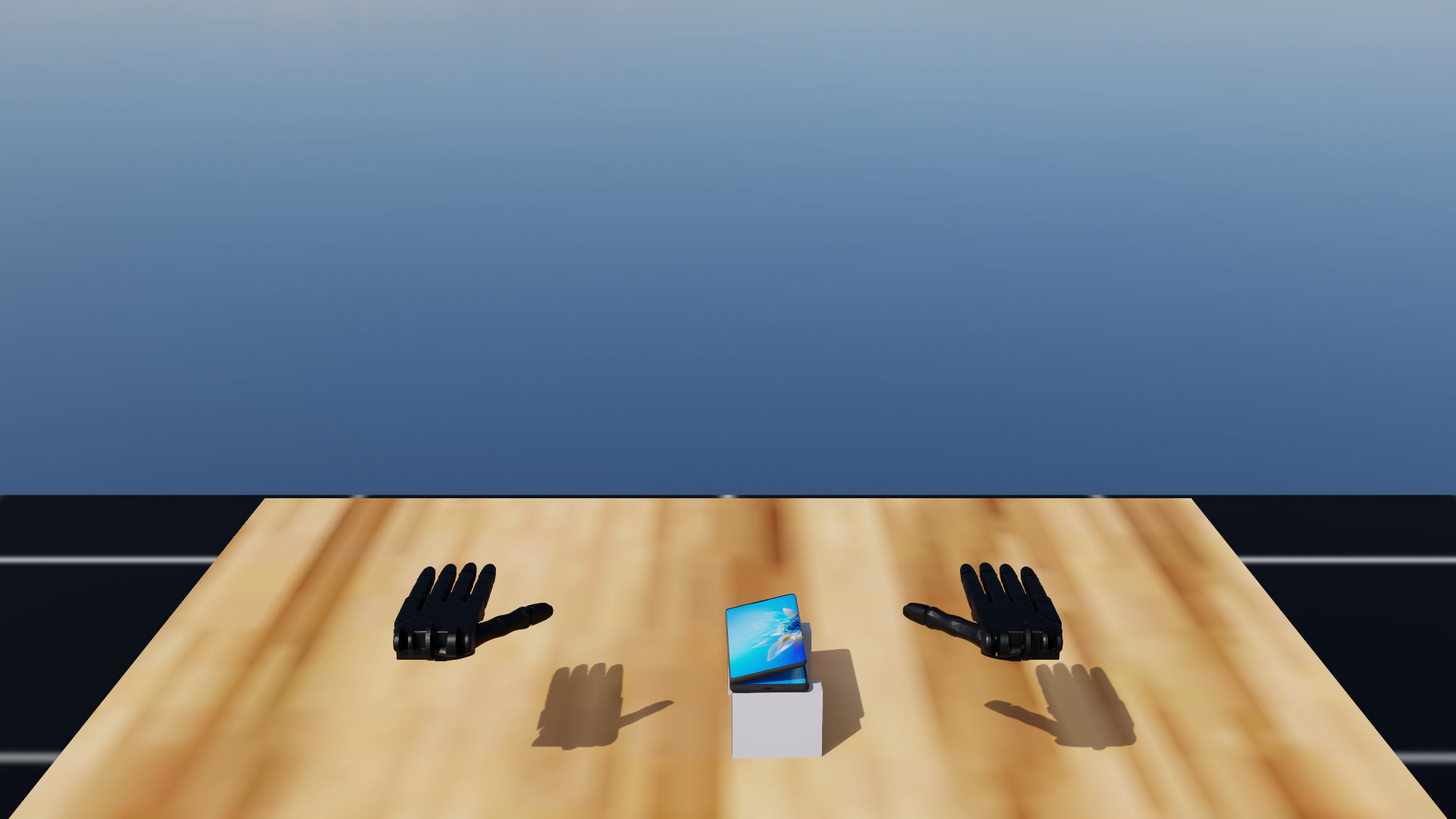} \\

    \addlinespace

    & \shortstack[l]{\texttt{OpenDouble}\\\texttt{Door}}
    & Open both target cabinet doors in a coordinated manner.
    & Both doors reach at least 80\% of their full swing range at the same time, and the gap between when each door first stays past that point is no more than 1.0s.
    & \includegraphics[width=0.95\linewidth]{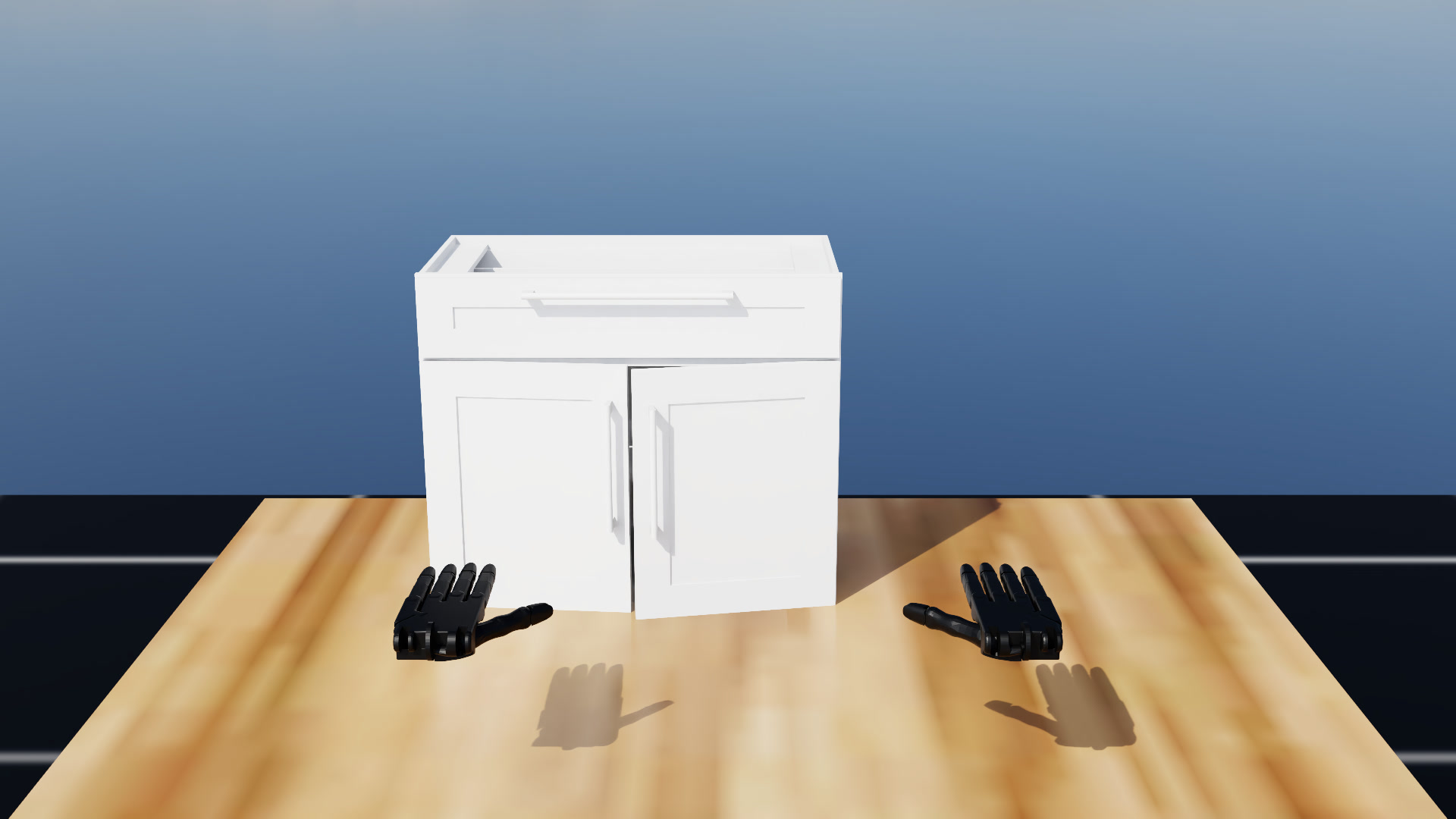} \\

    \addlinespace

    & \texttt{UnscrewCap}
    & Hold the tube and rotate the cap until it reaches the unscrewed state.
    & The cap is rotated at least 98\% of the way through its reachable unscrewing range from its screwed-in start.
    & \includegraphics[width=0.95\linewidth]{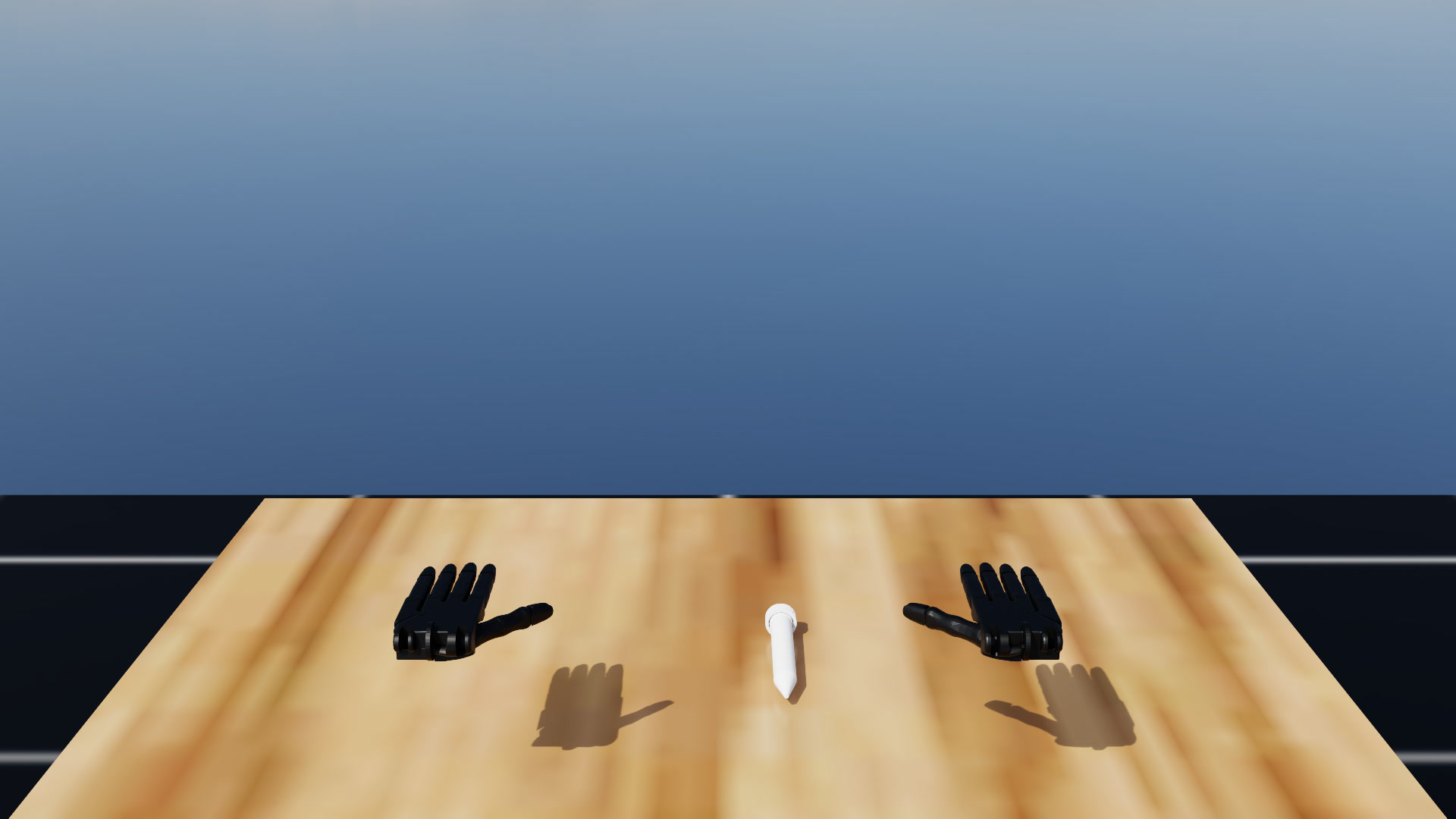} \\

    \addlinespace

    & \texttt{OpenGlasses}
    & Hold a pair of glasses and unfold both temple arms.
    & Both temple hinges independently open at least 80\% of the way from their slightly-open reset position.
    & \includegraphics[width=0.95\linewidth]{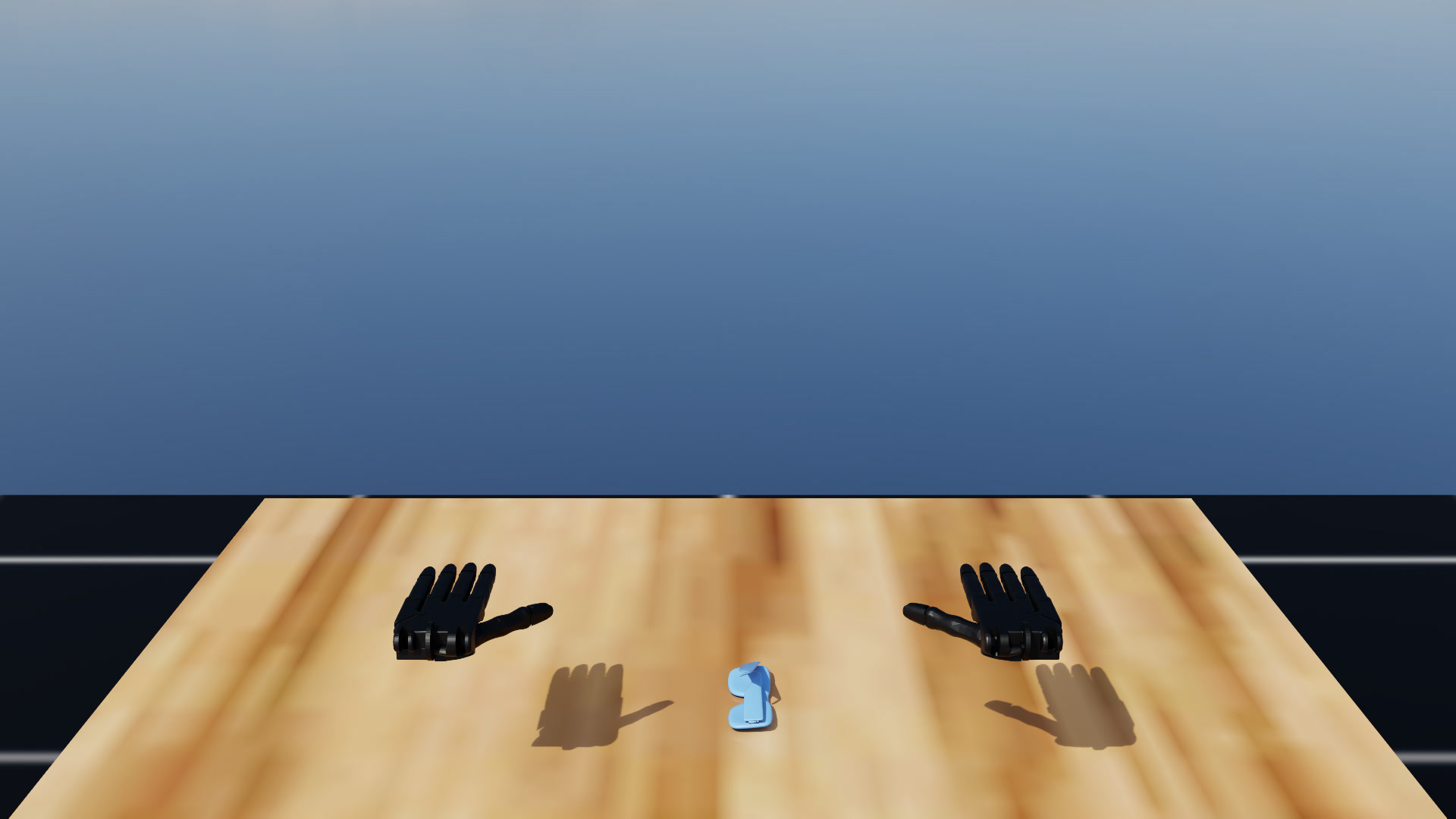} \\

    \addlinespace[4pt]
    \midrule
    \addlinespace[2pt]

    \addlinespace
    \categorycell{10}{Functional}
    & \texttt{GraspBleach}
    & Grasp a bleach bottle, lift it, and rotate it into a pouring posture while avoiding the nozzle.
    & The bottle is lifted at least 0.3 m, tilted at least 100$^\circ$ from vertical, and its nozzle brought within 0.02 m of the goal, all while not touching a small region around the nozzle.
    & \includegraphics[width=0.95\linewidth]{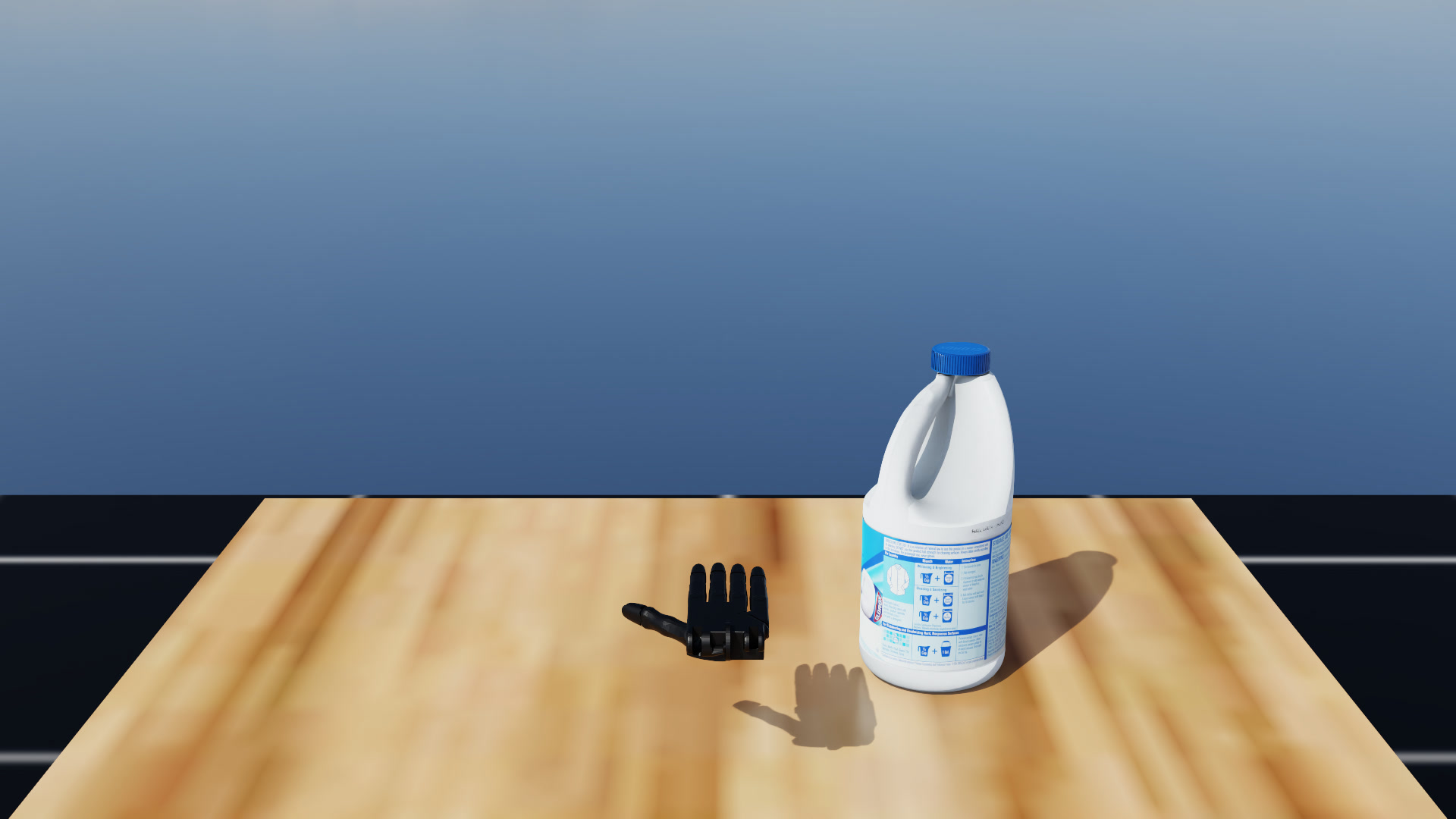} \\

    \addlinespace

    & \texttt{GraspPan}
    & Manipulate a frying pan into a controlled flat pose while avoiding contact with the cooking surface.
    & The pan face is brought within 10$^\circ$ of flat and its center within 0.025 m of the burner target, all while not touching the cooking surface.
    & \includegraphics[width=0.95\linewidth]{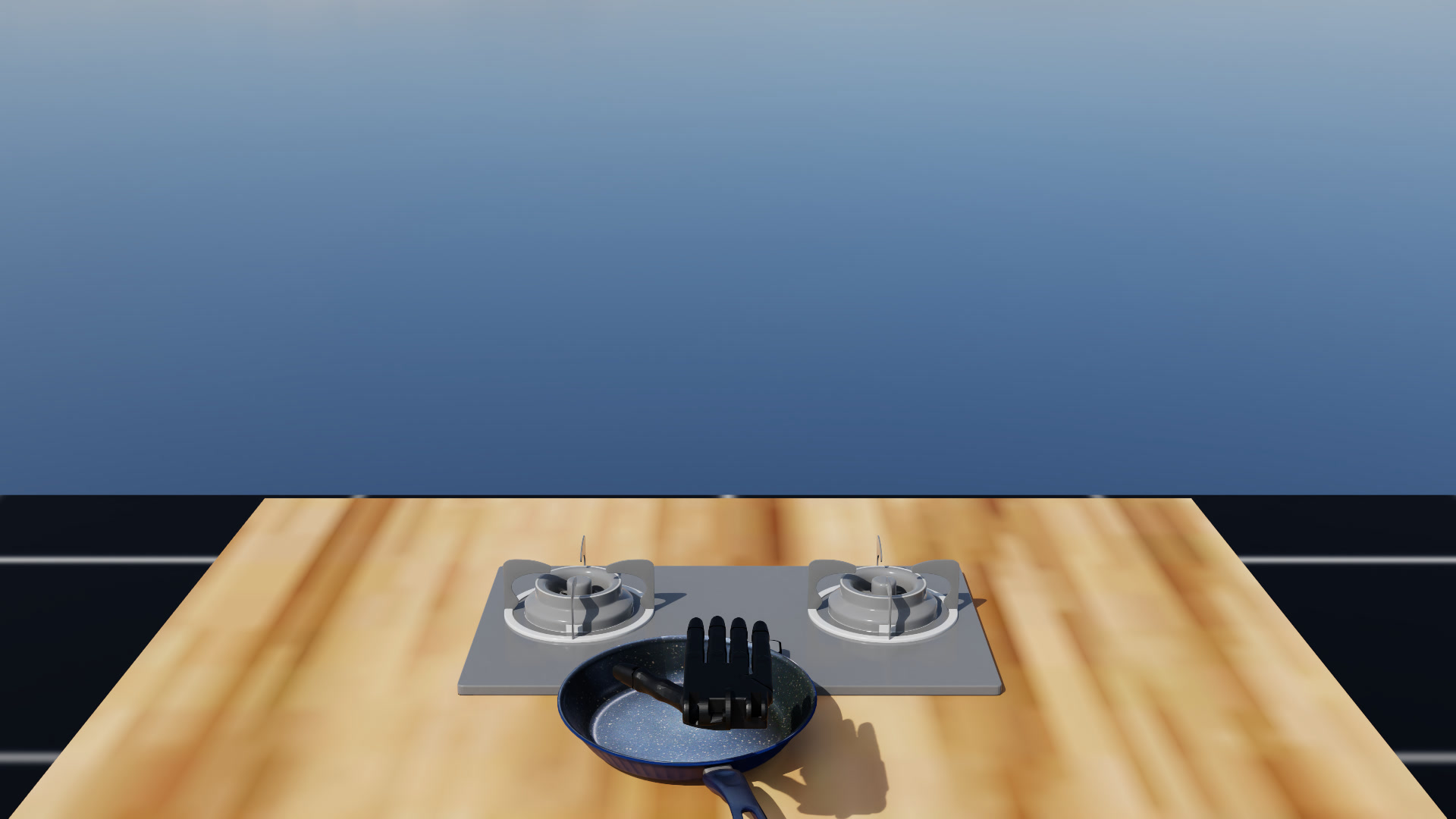} \\

    \addlinespace

    & \texttt{GraspKettle}
    & Grasp a kettle, lift it from the table, and rotate it into a pouring posture while avoiding the spout.
    & The kettle is lifted at least 0.15 m, tilted at least 45$^\circ$, and its spout brought within 0.05 m of the goal, all while not touching the spout.
    & \includegraphics[width=0.95\linewidth]{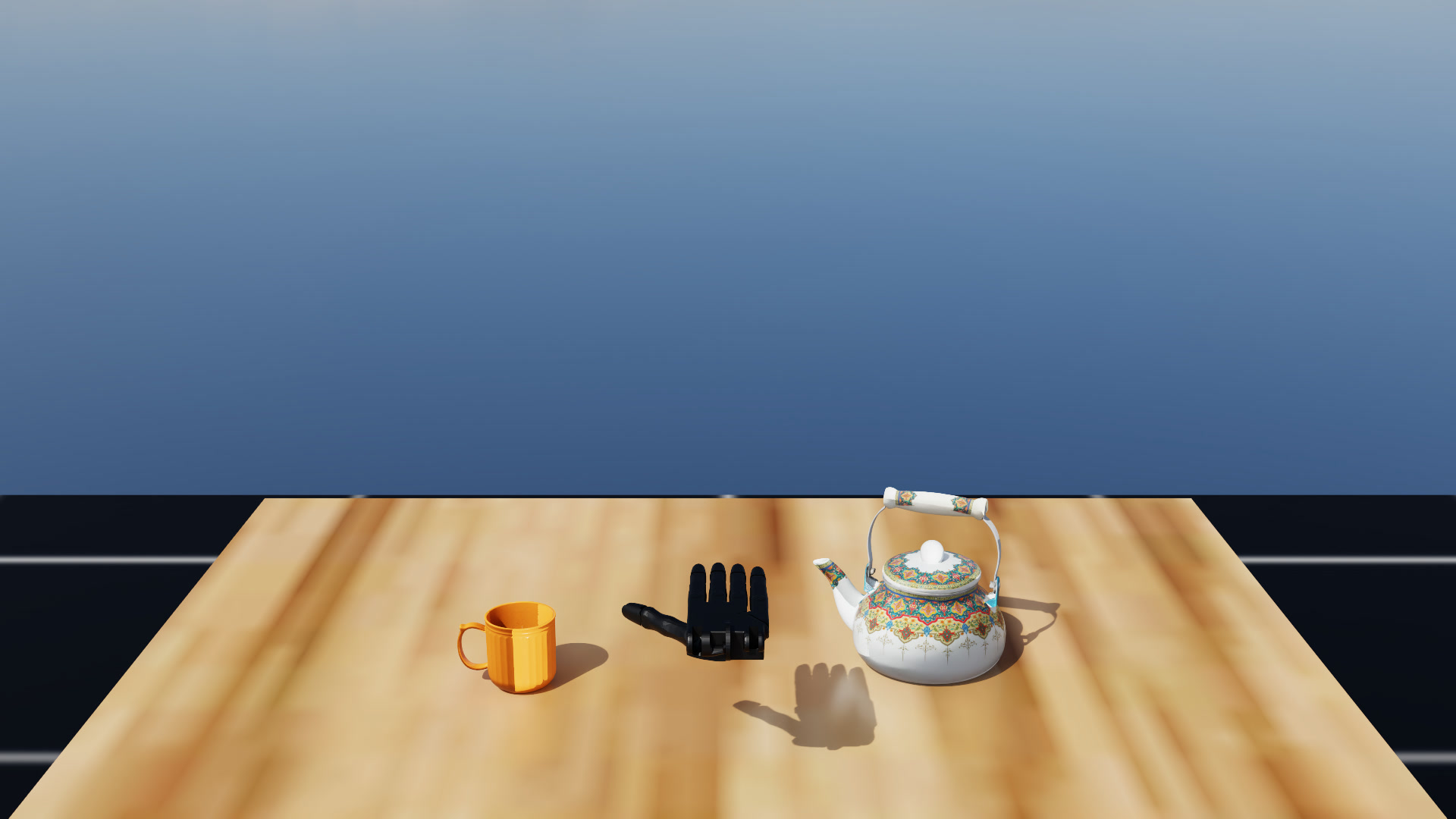} \\

    \addlinespace

    & \texttt{GraspCup}
    & Grasp a cup, lift it from the table, and rotate it into a pouring posture while avoiding the rim.
    & The cup is lifted at least 0.3 m, tilted at least 100$^\circ$, and its rim brought within 0.10 m of the goal, all while not touching the rim region.
    & \includegraphics[width=0.95\linewidth]{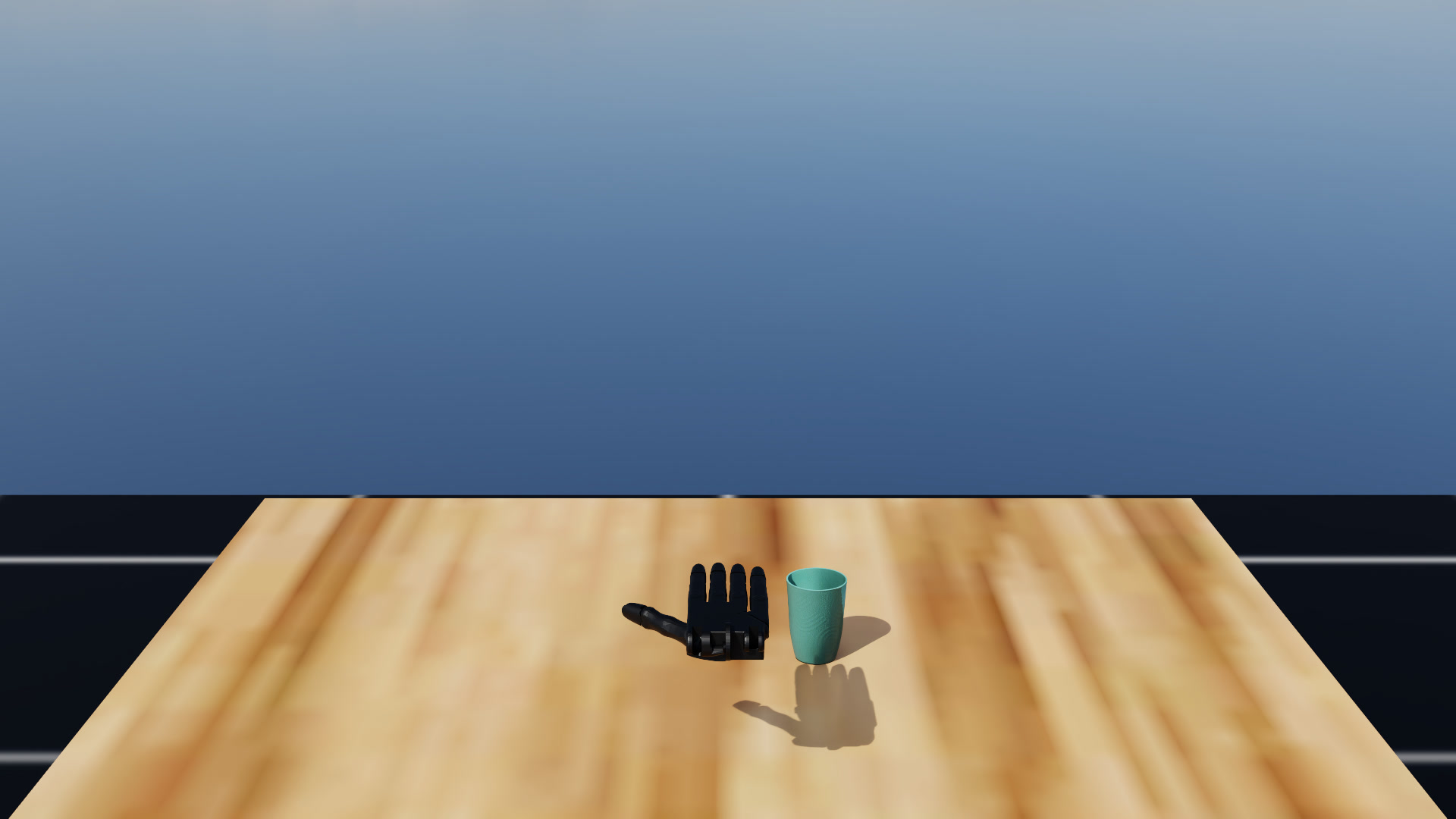} \\

    \addlinespace

    & \shortstack[l]{\texttt{RetrieveCup}}
    & Remove a cup from a rack and place it upright at the commanded goal.
    & The cup comes within 0.04 m of the goal on the table and stays within 15$^\circ$ of upright.
    & \includegraphics[width=0.95\linewidth]{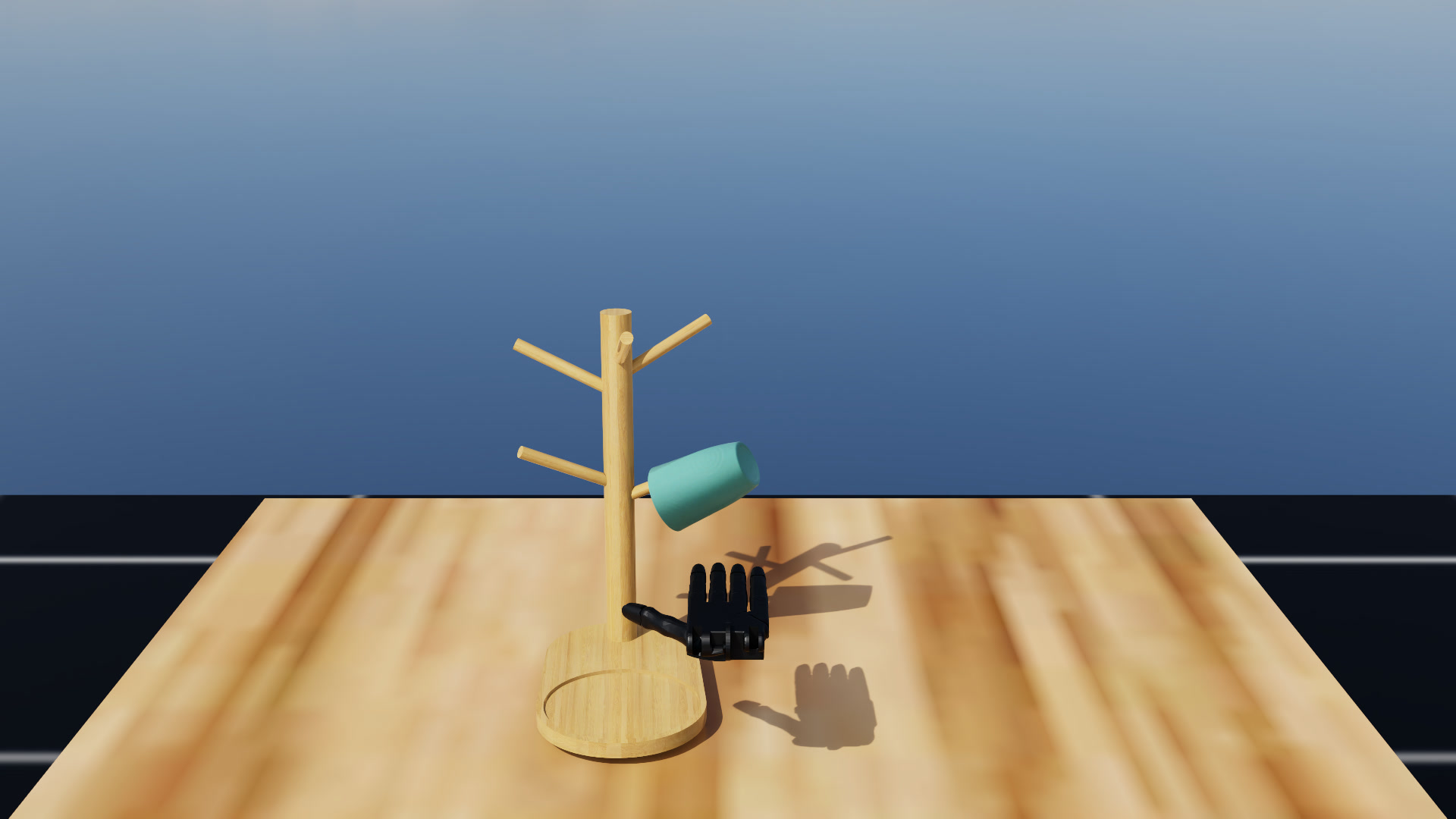} \\

    \addlinespace

    & \texttt{LiftBucket}
    & Grasp an articulated bucket and lift the bucket body away from the tabletop.
    & The bucket is lifted at least 0.12 m above its spawn height.
    & \includegraphics[width=0.95\linewidth]{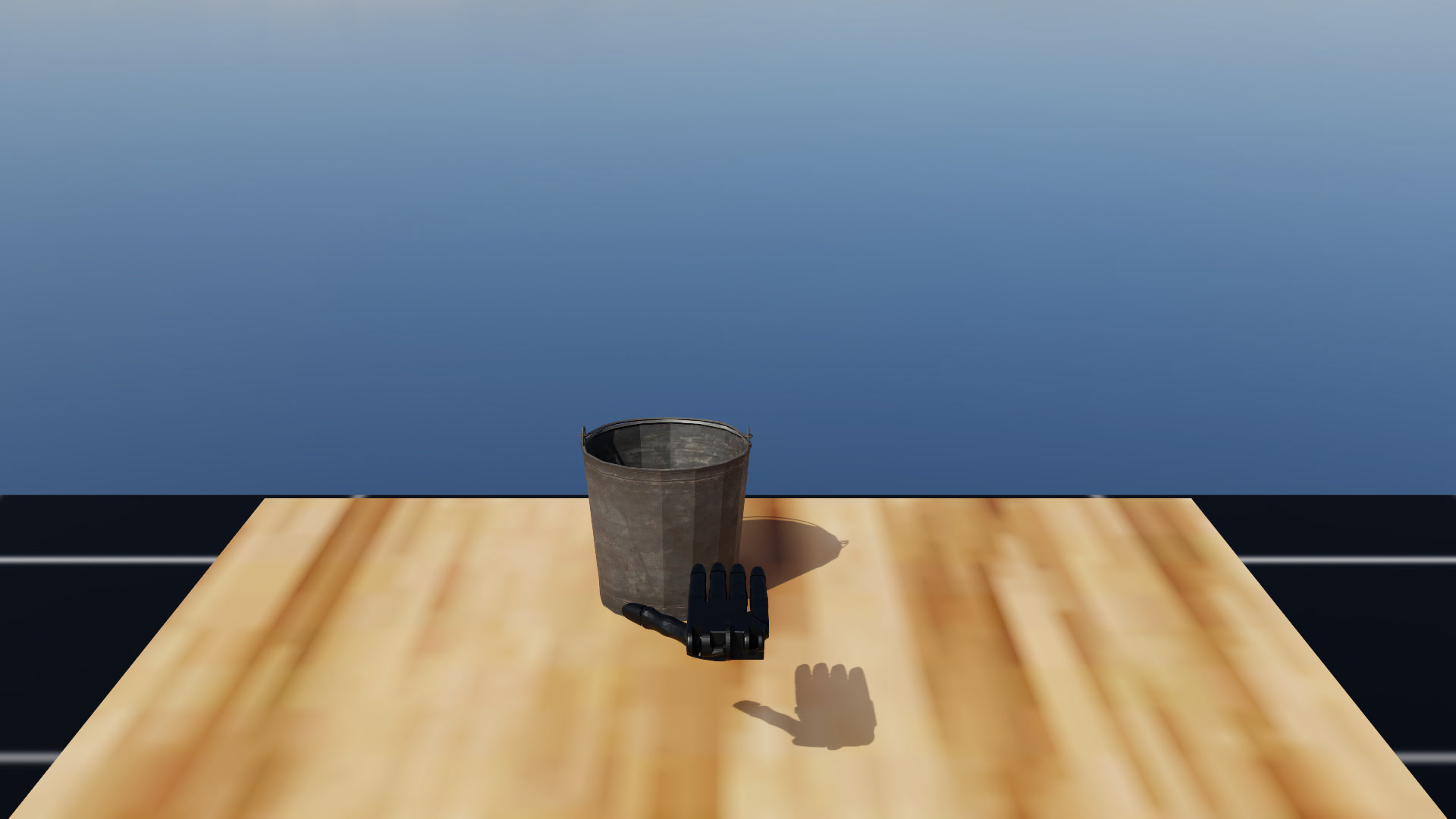} \\

    \addlinespace
    & \shortstack[l]{\texttt{Functional}\\\texttt{PourCan}}
    & Grasp a soup can, lift it from the table, and rotate it into a pouring posture.
    & The can is lifted at least 0.2 m, tilted at least 100$^\circ$, and its lip brought within 0.10 m of the goal, all while not touching the lip.
    & \includegraphics[width=0.95\linewidth]{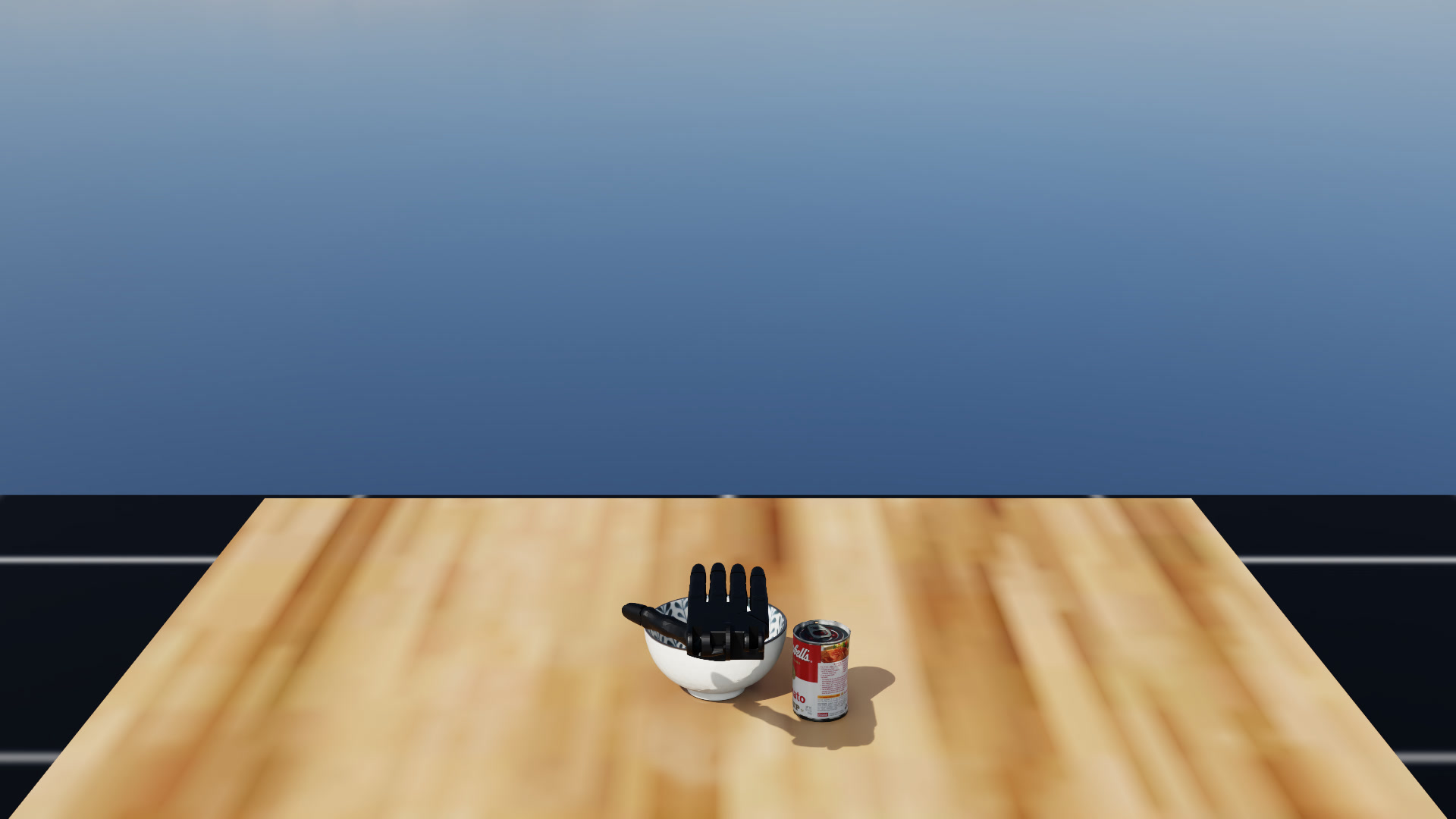} \\

    \addlinespace

    & \shortstack[l]{\texttt{Functional}\\\texttt{PourMug}}
    & Grasp a mug, lift it from the table, and rotate it into a pouring posture.
    & The mug is lifted at least 0.2 m, tilted at least 100$^\circ$, and its rim brought within 0.10 m of the goal, all while not touching the rim.
    & \includegraphics[width=0.95\linewidth]{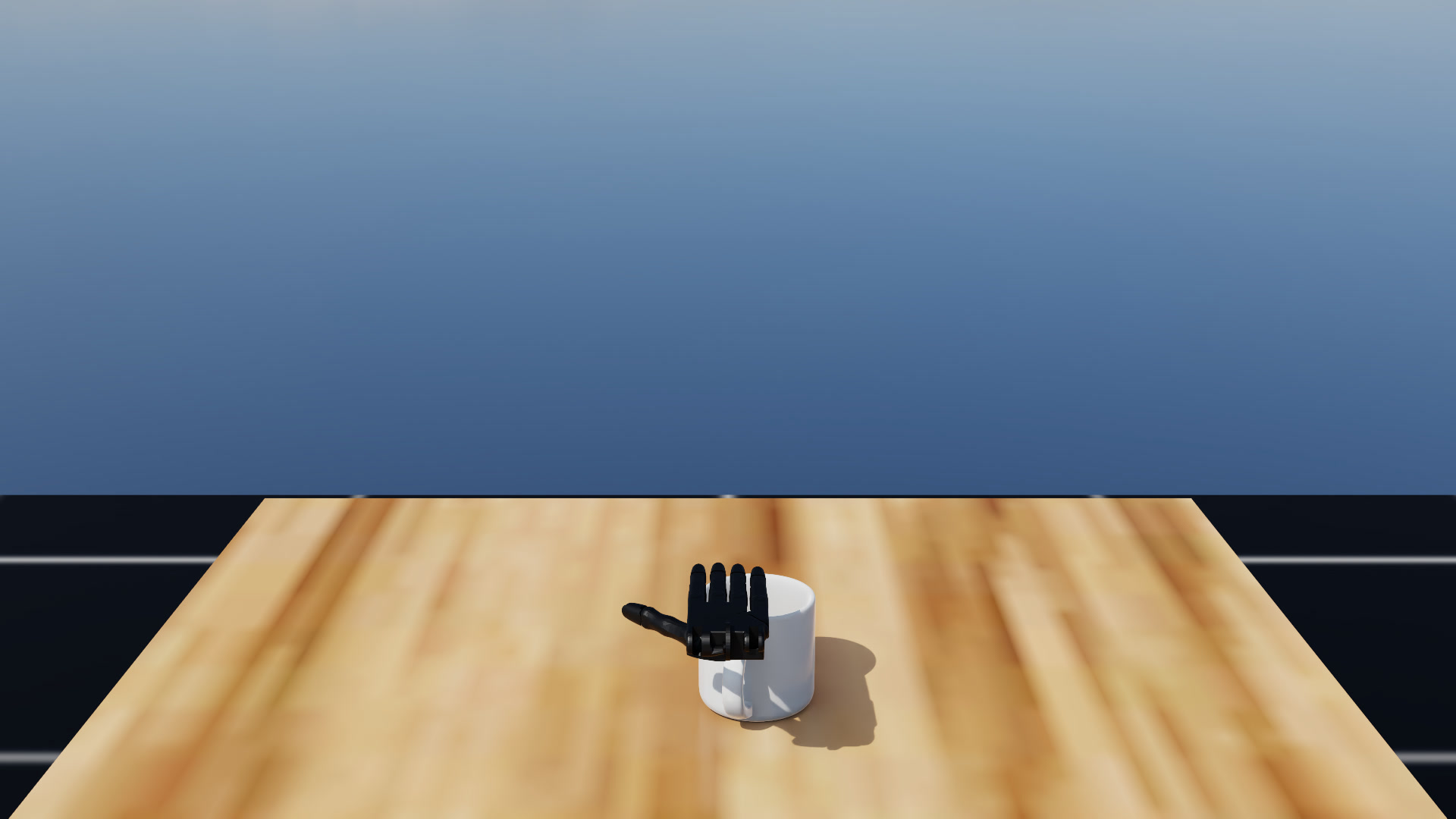} \\

    \addlinespace

    & \shortstack[l]{\texttt{Functional}\\\texttt{HammerStrike}}
    & Use a hammer to strike or press a nail into a board while avoiding unsafe fingertip contact.
    & The nail is pressed in at least 0.06 m, and the hand does not touch the nail head directly.
    & \includegraphics[width=0.95\linewidth]{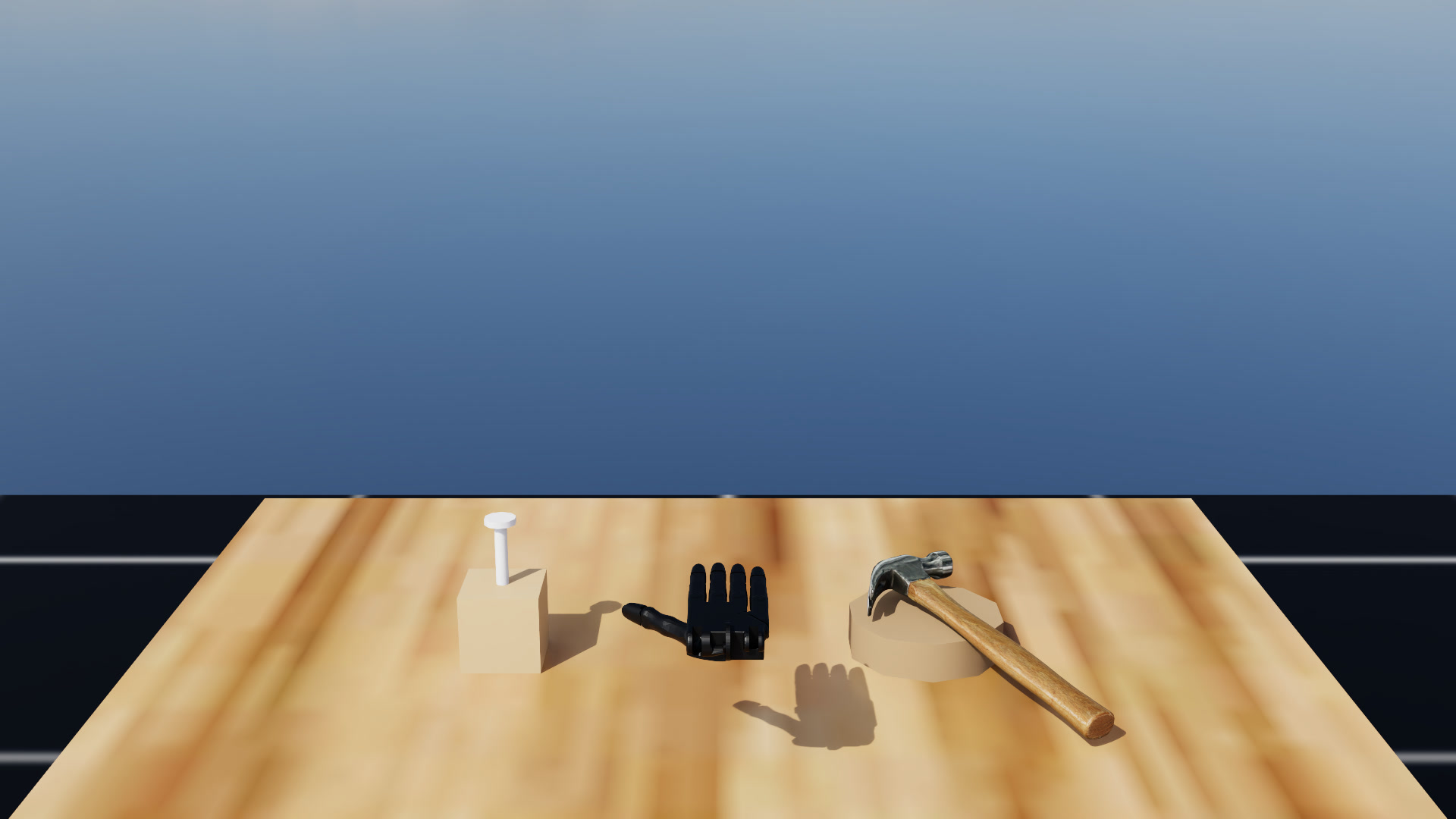} \\

    \addlinespace

    & \shortstack[l]{\texttt{Functional}\\\texttt{DrillApply}}
    & Pick up a drill and bring its bit onto a marked target point on the work surface.
    & The bit tip is brought within 0.025 m of the target while the drill is held near-vertical, all while not touching the chuck or bit.
    & \includegraphics[width=0.95\linewidth]{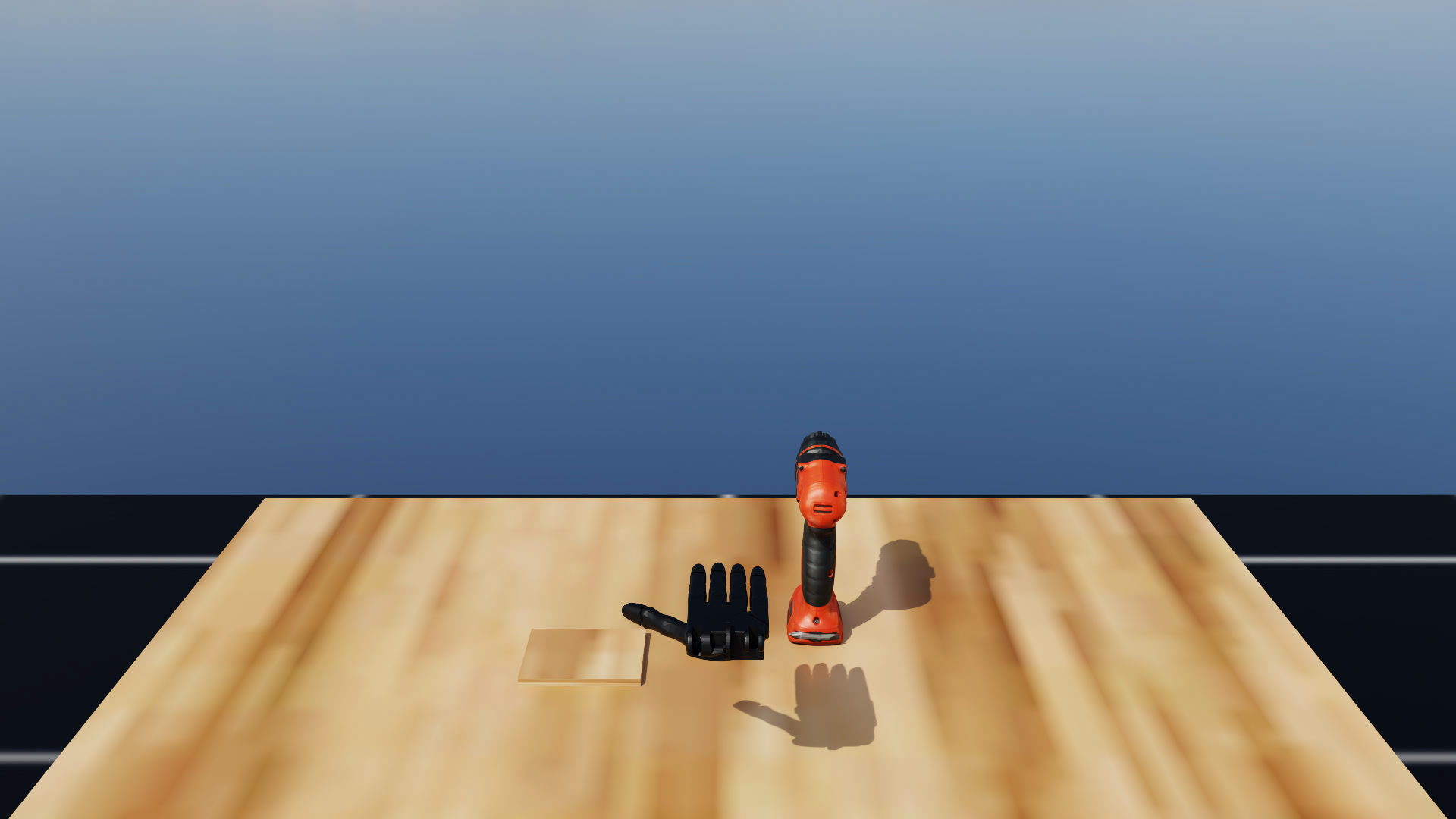} \\

    \addlinespace

    & \texttt{PourWineGlass}
    & Lift the wine glass and tilt it into a pouring orientation.
    & The glass is lifted at least 0.20 m above its spawn height and tilted at least 100$^\circ$ from vertical, all while not touching the opening of the glass.
    & \includegraphics[width=0.95\linewidth]{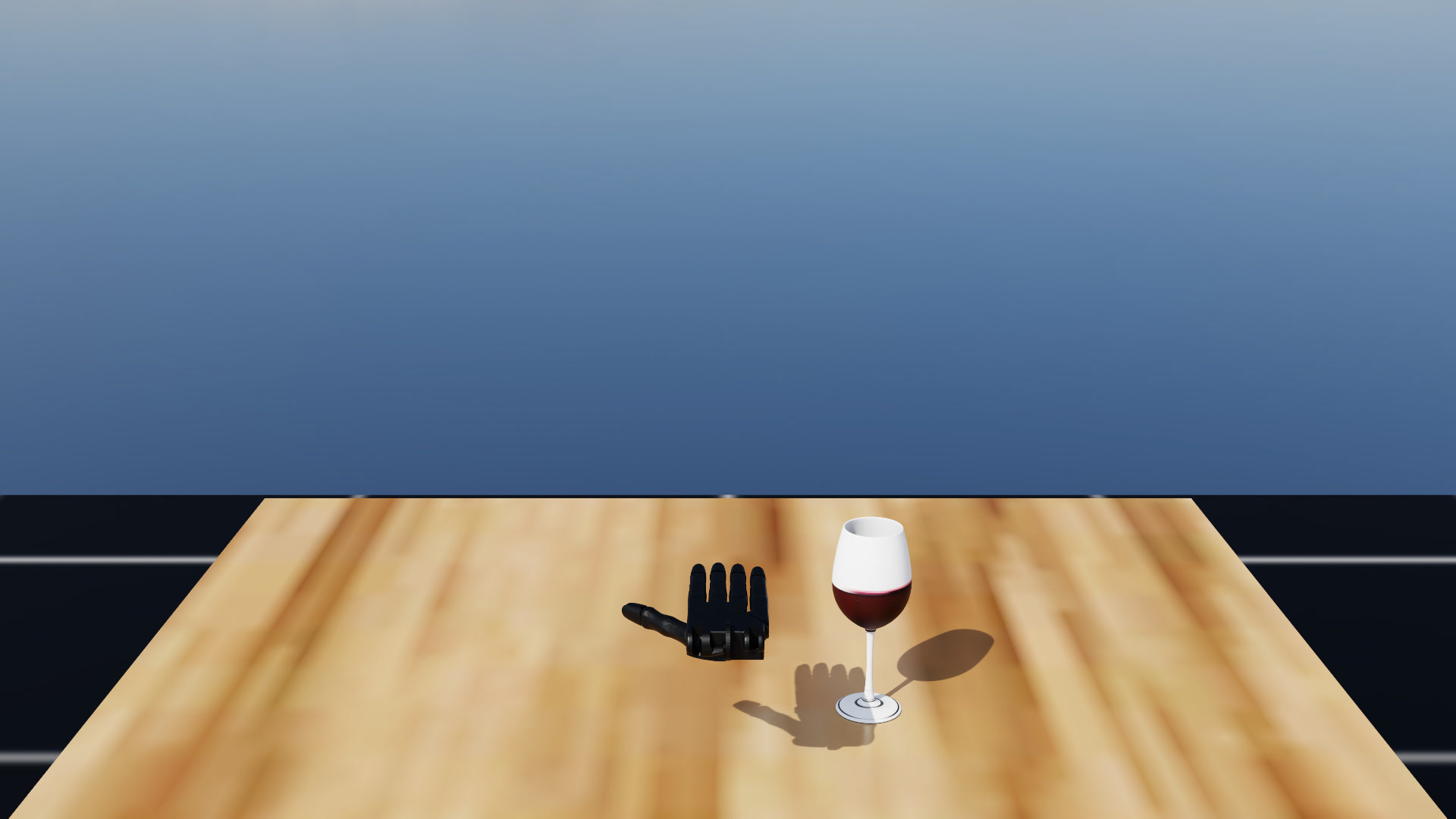} \\

    \addlinespace[4pt]
    \midrule
    \addlinespace[2pt]
    \newpage 
    \addlinespace
    \categorycell{8}{Contact-rich}
    & \texttt{NutThread}
    & Align a nut with a fixed bolt and thread it down onto the bolt.
    & The nut is centered within about 0.0025 m of the bolt axis and threaded down to within roughly 0.002 m of the target depth, about 1.5 turns down the bolt.
    & \includegraphics[width=0.95\linewidth]{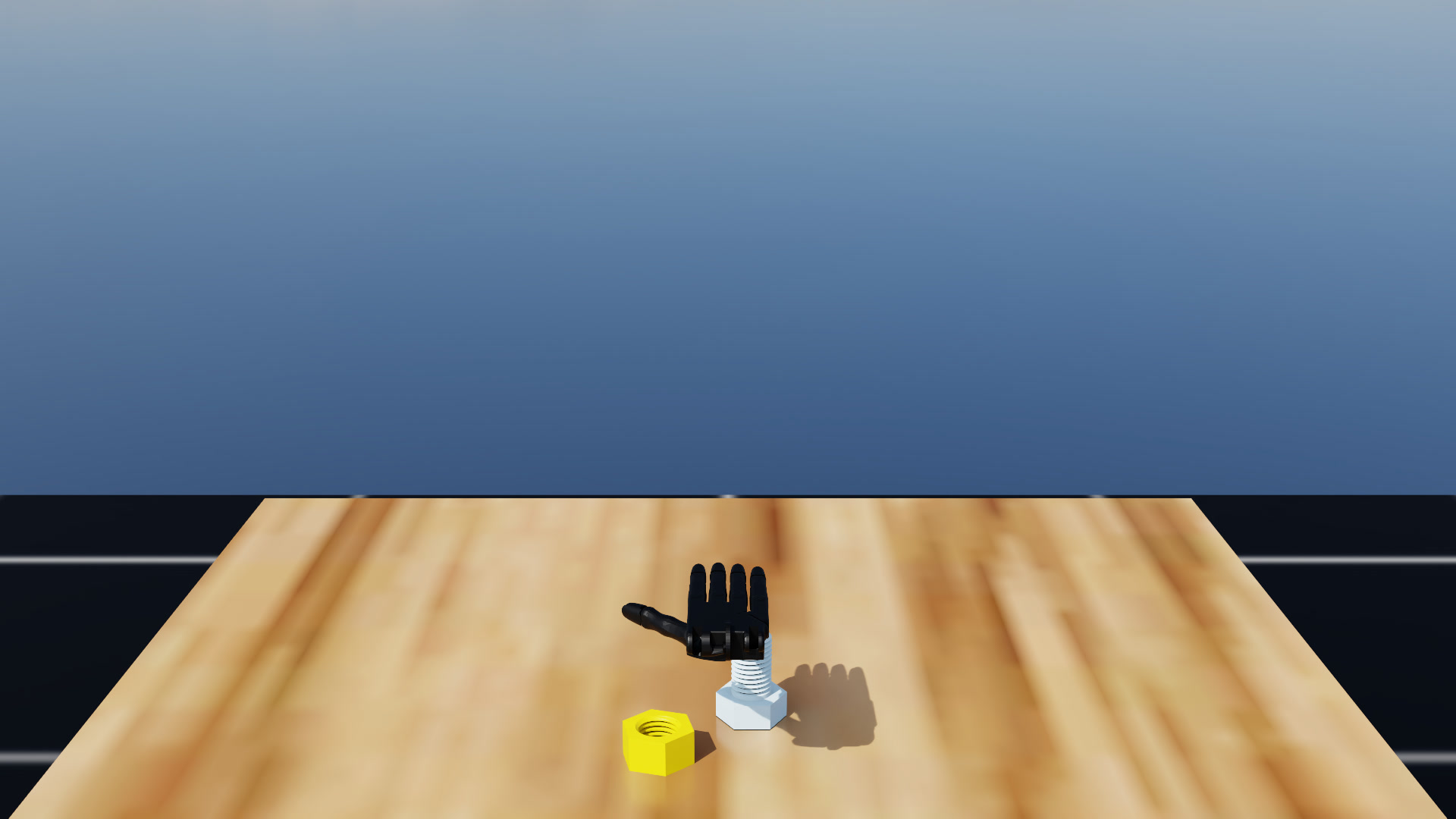} \\

    \addlinespace

    & \texttt{InsertPipette}
    & Pick up a pipette and guide its tip into the neck of the target glassware.
    & The tip is centered within 0.02 m of the glassware opening and pushed at least 0.01 m below the rim.
    & \includegraphics[width=0.95\linewidth]{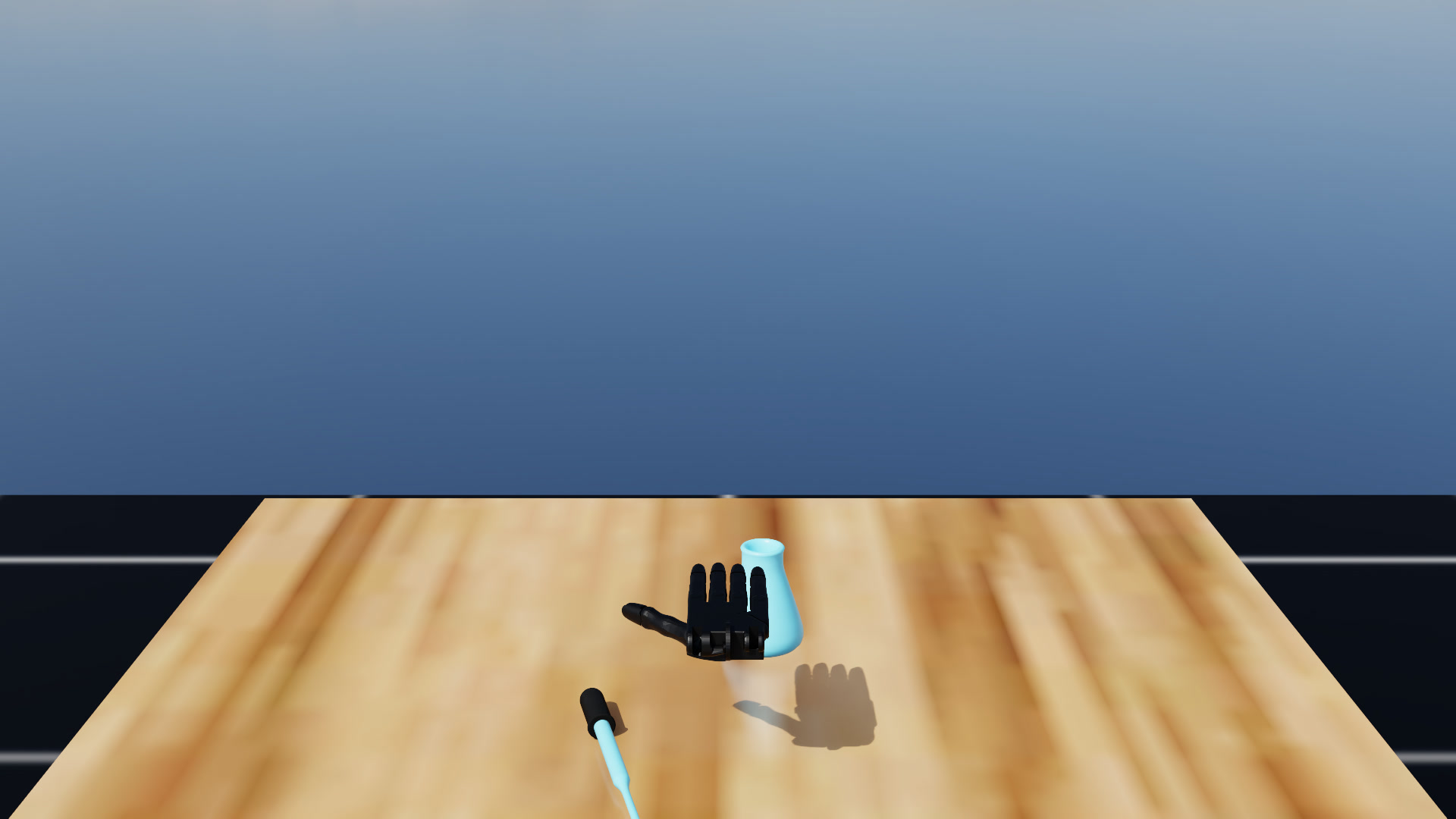} \\

    \addlinespace

    & \texttt{PlugCharger}
    & Align a charger plug with a fixed receptacle and insert the plug into the socket.
    & The plug tip is inserted into the receptacle with less than 0.0025 m of lateral and vertical misalignment.
    & \includegraphics[width=0.95\linewidth]{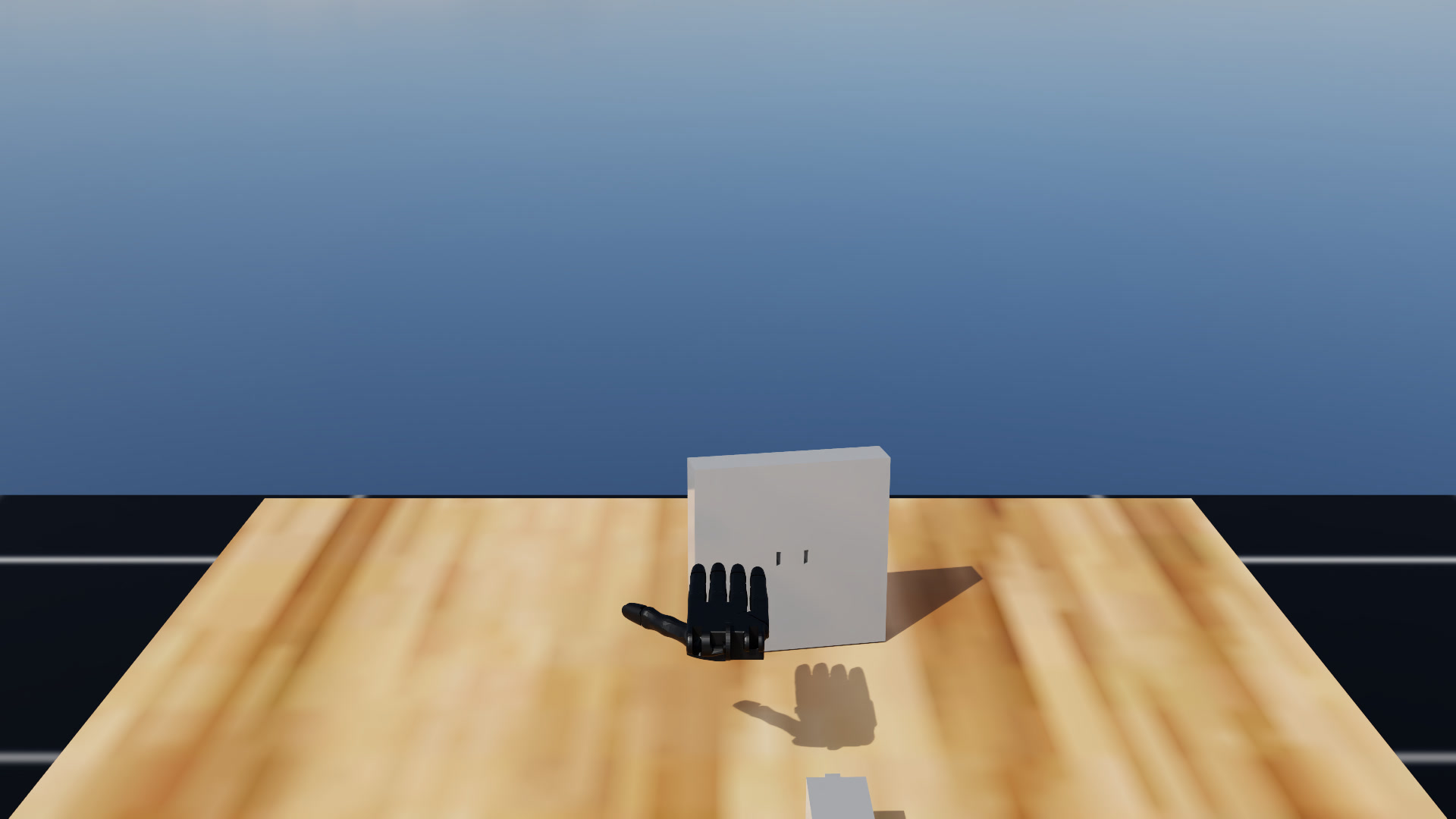} \\

    \addlinespace

    & \texttt{InsertPen}
    & Pick up a pen and guide either tip into the opening of a pen holder.
    & Either pen tip is centered within about 0.038 m of the holder opening and pushed at least 0.03 m below its rim.
    & \includegraphics[width=0.95\linewidth]{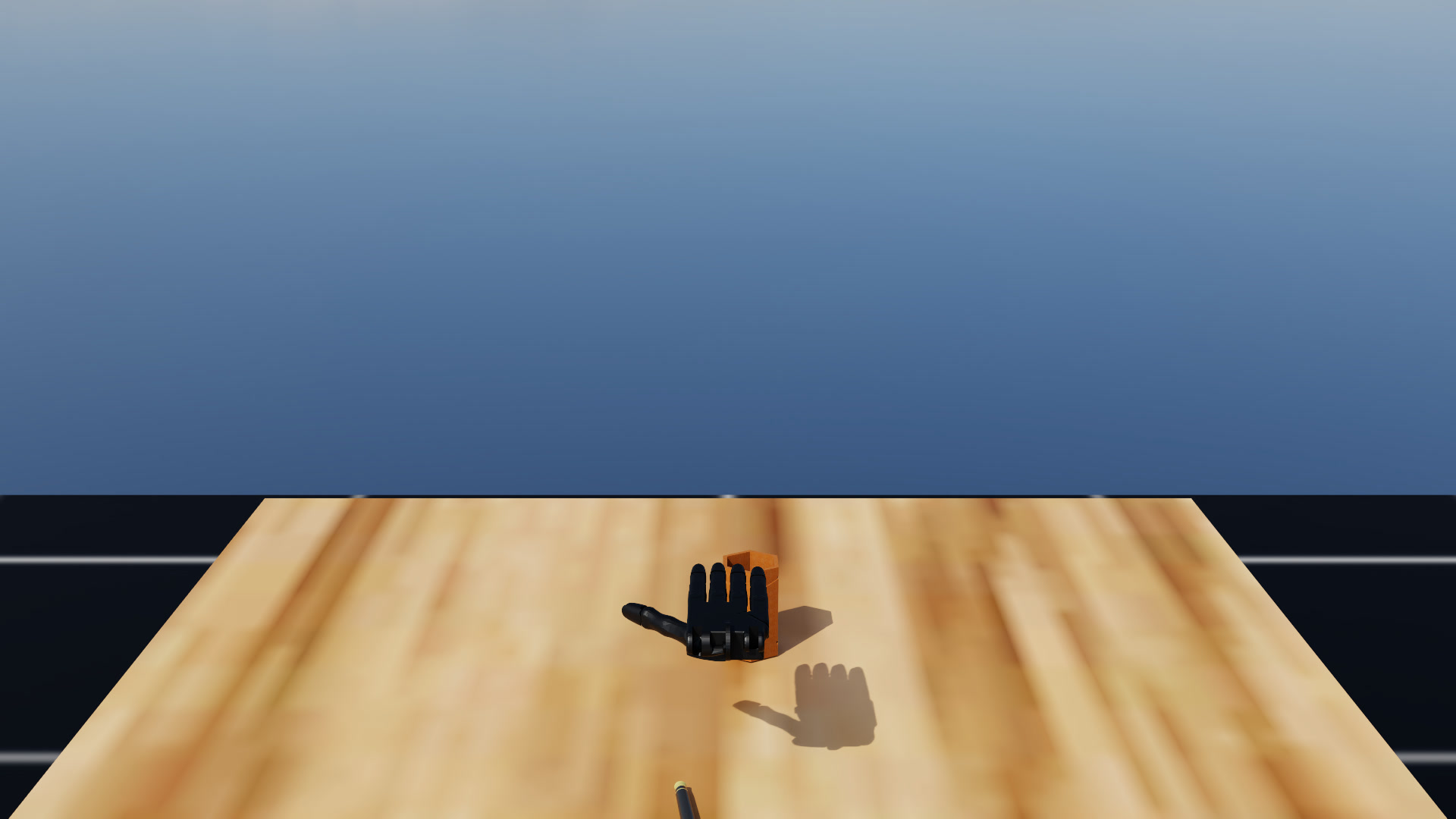} \\

    \addlinespace

    & \texttt{InsertGear}
    & Pick and place a gear onto the fixed gear base so that it seats into the mesh position.
    & The gear's shaft point is centered within 0.0025 m of the base's shaft and seated to within 0.003 m of full depth.
    & \includegraphics[width=0.95\linewidth]{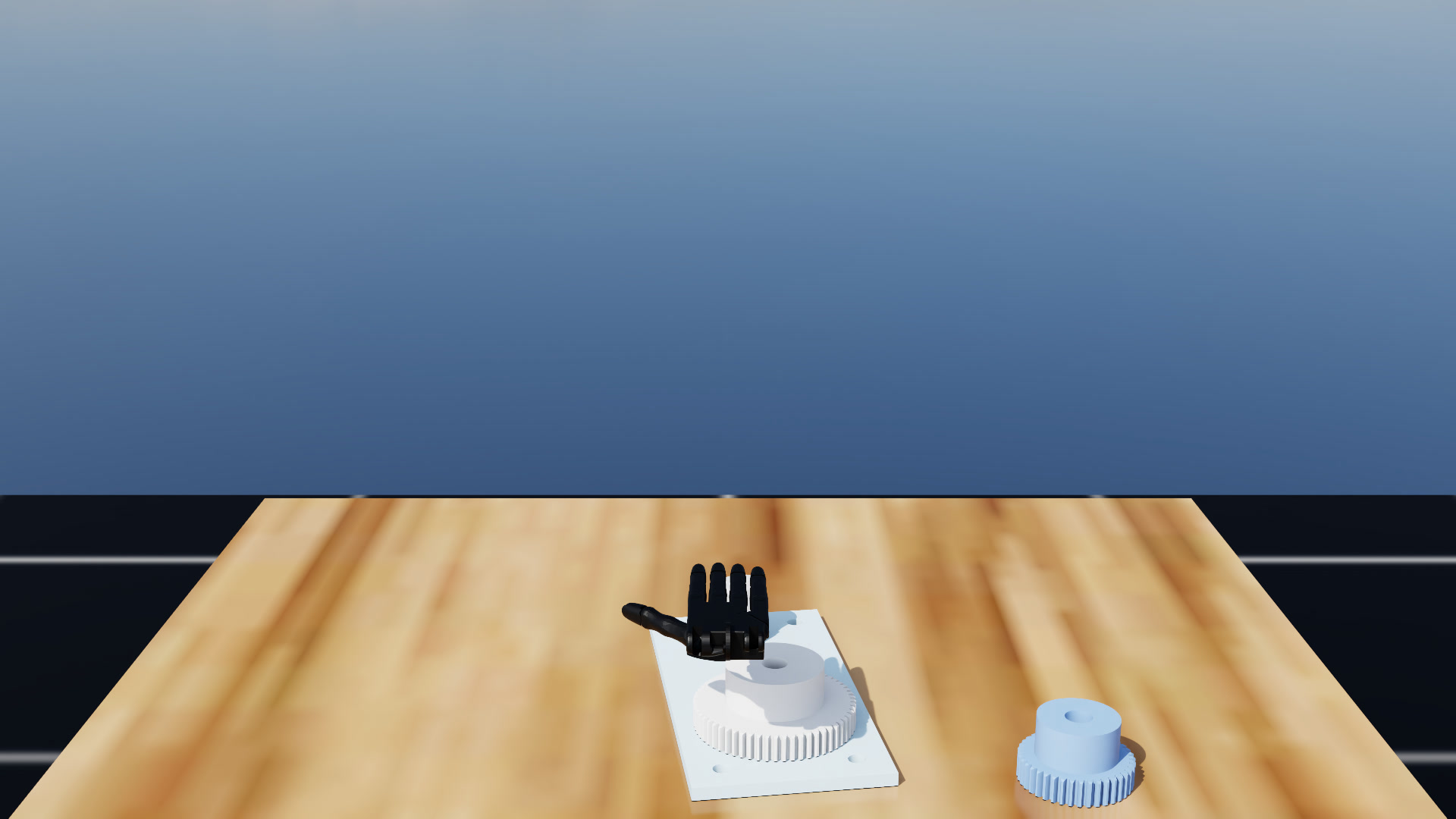} \\

    \addlinespace
    & \texttt{InsertPeg}
    & Align a side peg with a fixed tabletop hole and insert it into the opening.
    & The peg head is inserted to within 0.015 m along the axis and stays within the hole's 0.023 m bore radius laterally.
    & \includegraphics[width=0.95\linewidth]{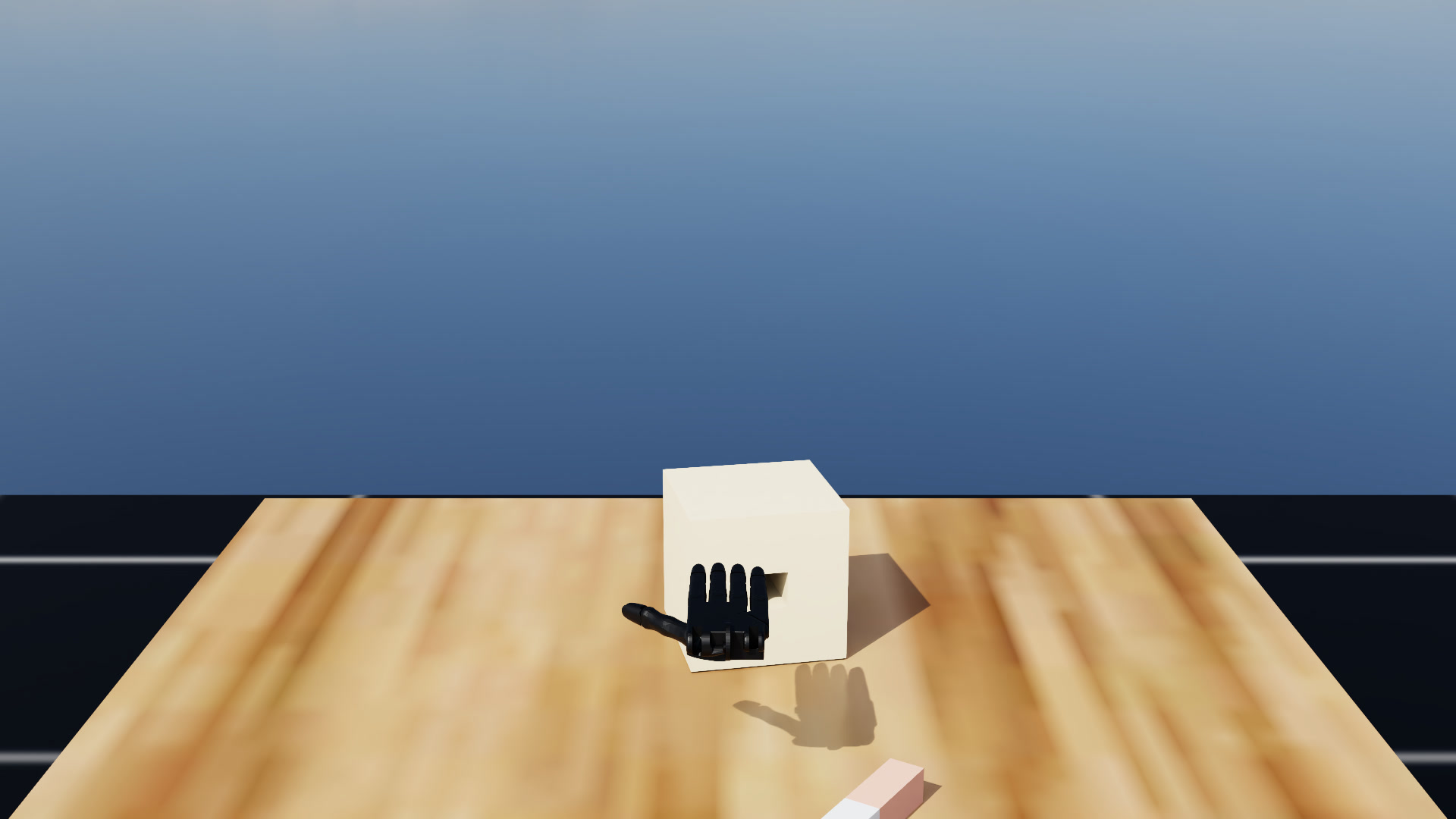} \\

    \addlinespace

    & \shortstack[l]{\texttt{PickFrom}\\\texttt{Clutter}}
    & Pick the green target object out of a cluttered corral containing distractor objects.
    & The target object is lifted at least 0.20 m above its default reset height; the 19 distractor objects are not checked.
    & \includegraphics[width=0.95\linewidth]{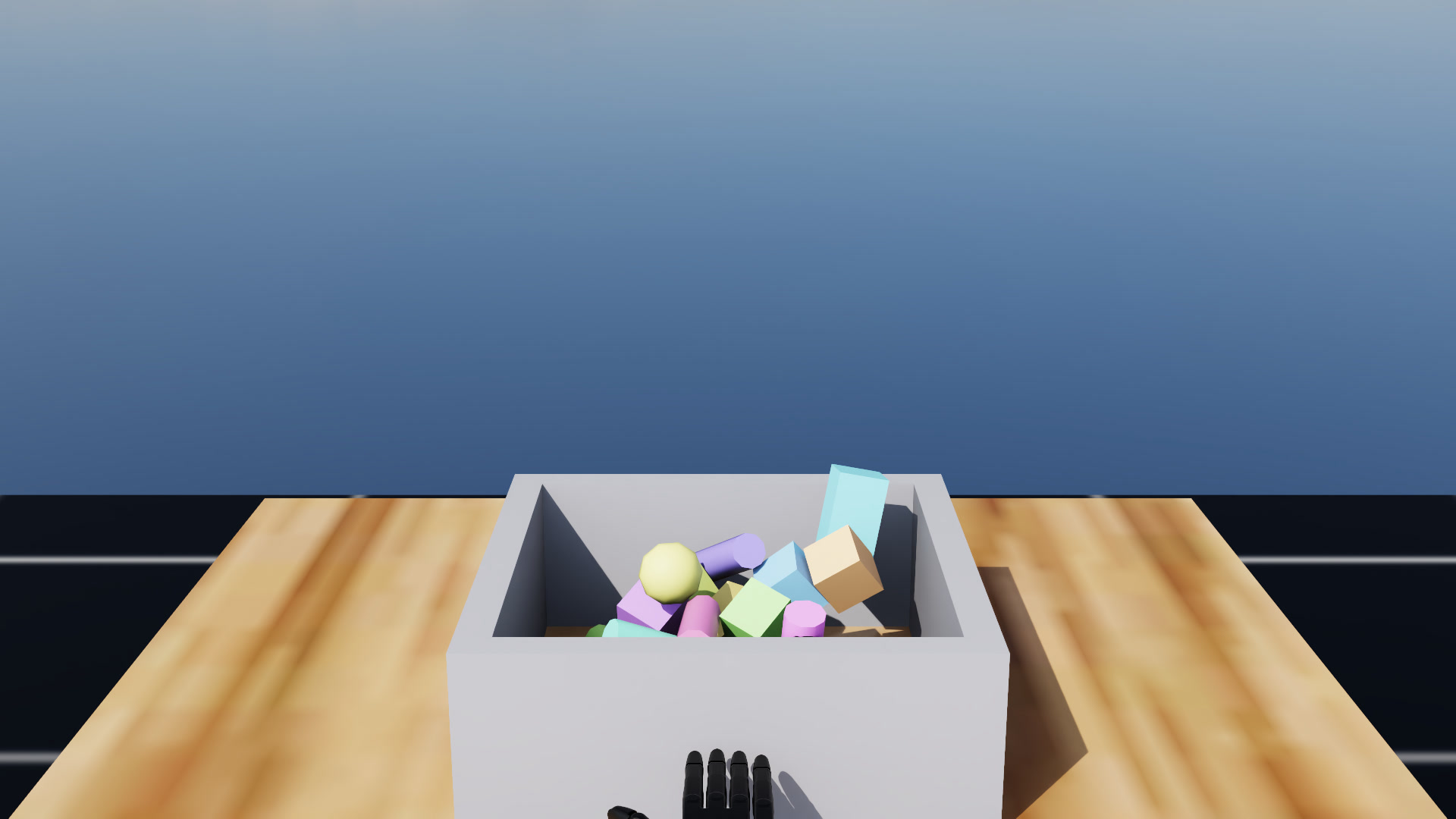} \\

    \addlinespace

    & \shortstack[l]{\texttt{PickThinObject}\\\texttt{FromContainer}}
    & Extract a thin object from a small container and move it to the commanded goal.
    & The object comes within 0.04 m of the extraction goal, which sits roughly 0.125 m above the object's start, clear of the container's 0.11 m walls and pulled forward out of the container.
    & \includegraphics[width=0.95\linewidth]{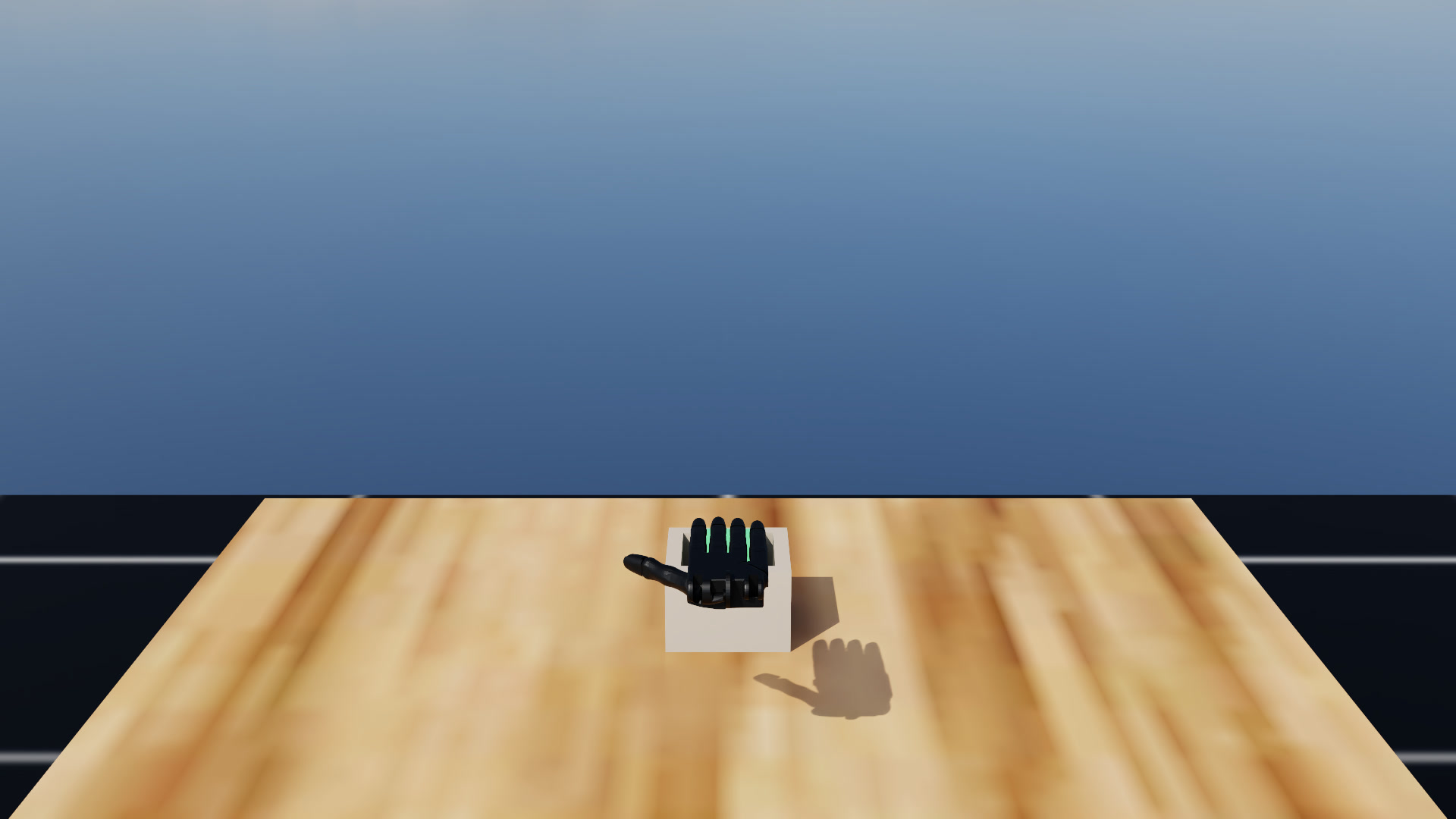} \\

    \addlinespace[4pt]
    \midrule
    \addlinespace[2pt]

    \addlinespace
    \categorycell{5}{Non-prehensile}
    & \shortstack[l]{\texttt{PushSphere}\\\texttt{UpSlope}}
    & Push a sphere up a sloped ramp and guide it to the commanded goal region.
    & The sphere comes within 0.08 m of a goal near the top of the 15$^\circ$ slope, sampled within about 0.30 m across the slope's width.
    & \includegraphics[width=0.95\linewidth]{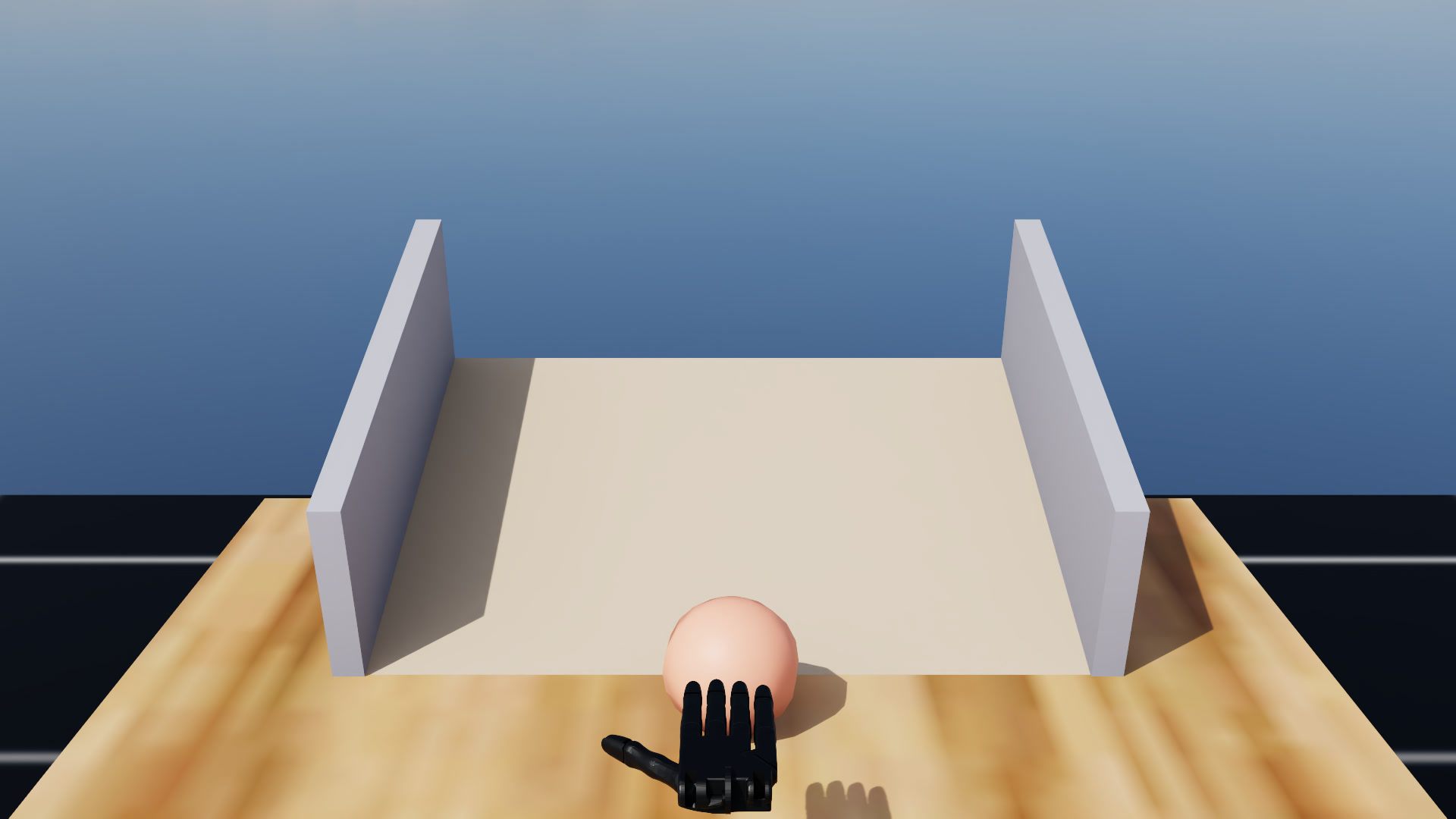} \\

    \addlinespace

    & \shortstack[l]{\texttt{PushSmallSphere}\\\texttt{ObstacleSlope}}
    & Push a small sphere uphill through randomly spawn obstacles until it crosses the target line.
    & The sphere crosses the target line, reaching or passing both the target's forward position and height.
    & \includegraphics[width=0.95\linewidth]{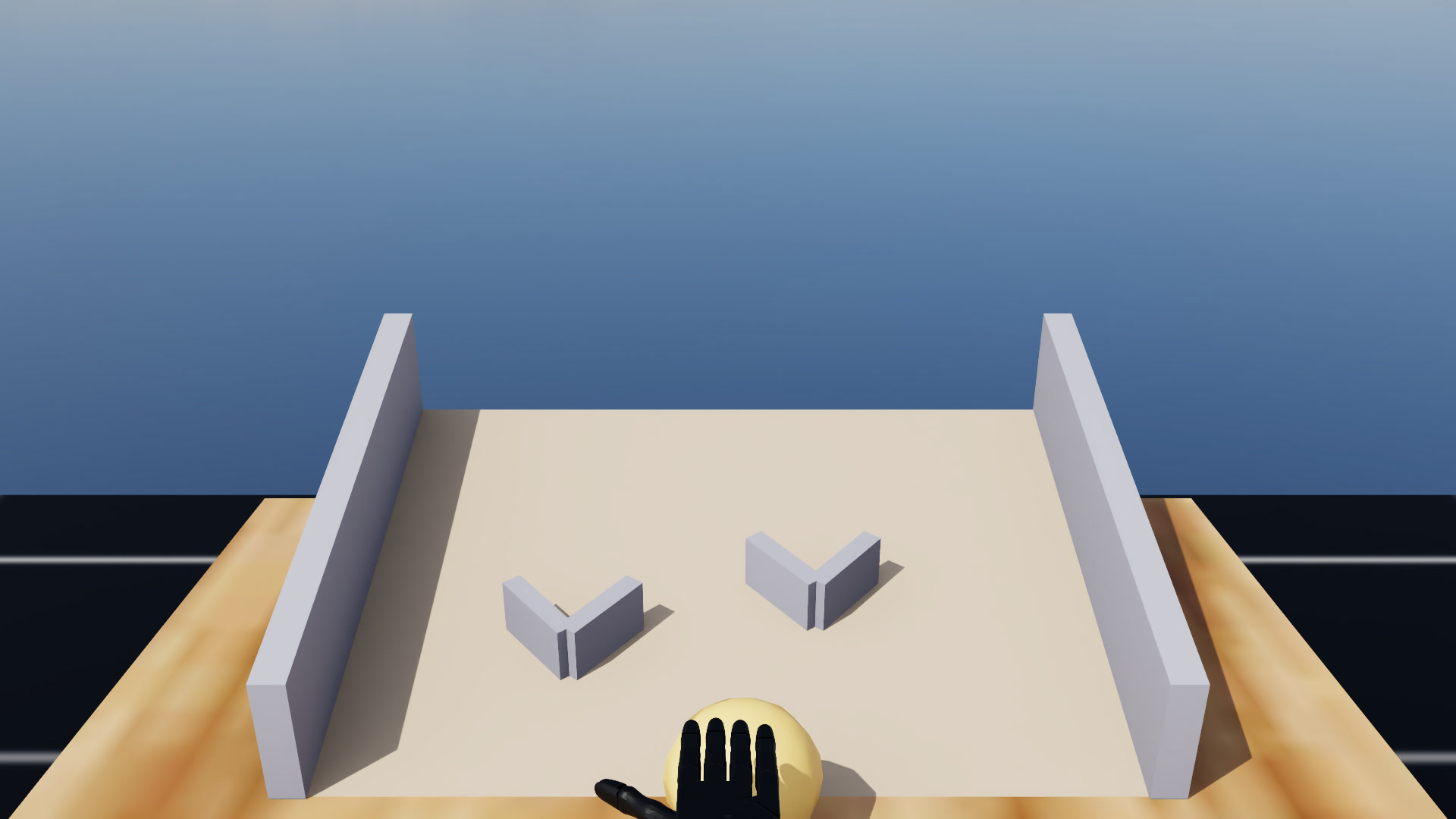} \\

    \addlinespace

    & \texttt{PushT}
    & Push a T-shaped object across the plane until it aligns with the target pose.
    & The pushed T-shape overlaps at least 90\% of the goal T's footprint.
    & \includegraphics[width=0.95\linewidth]{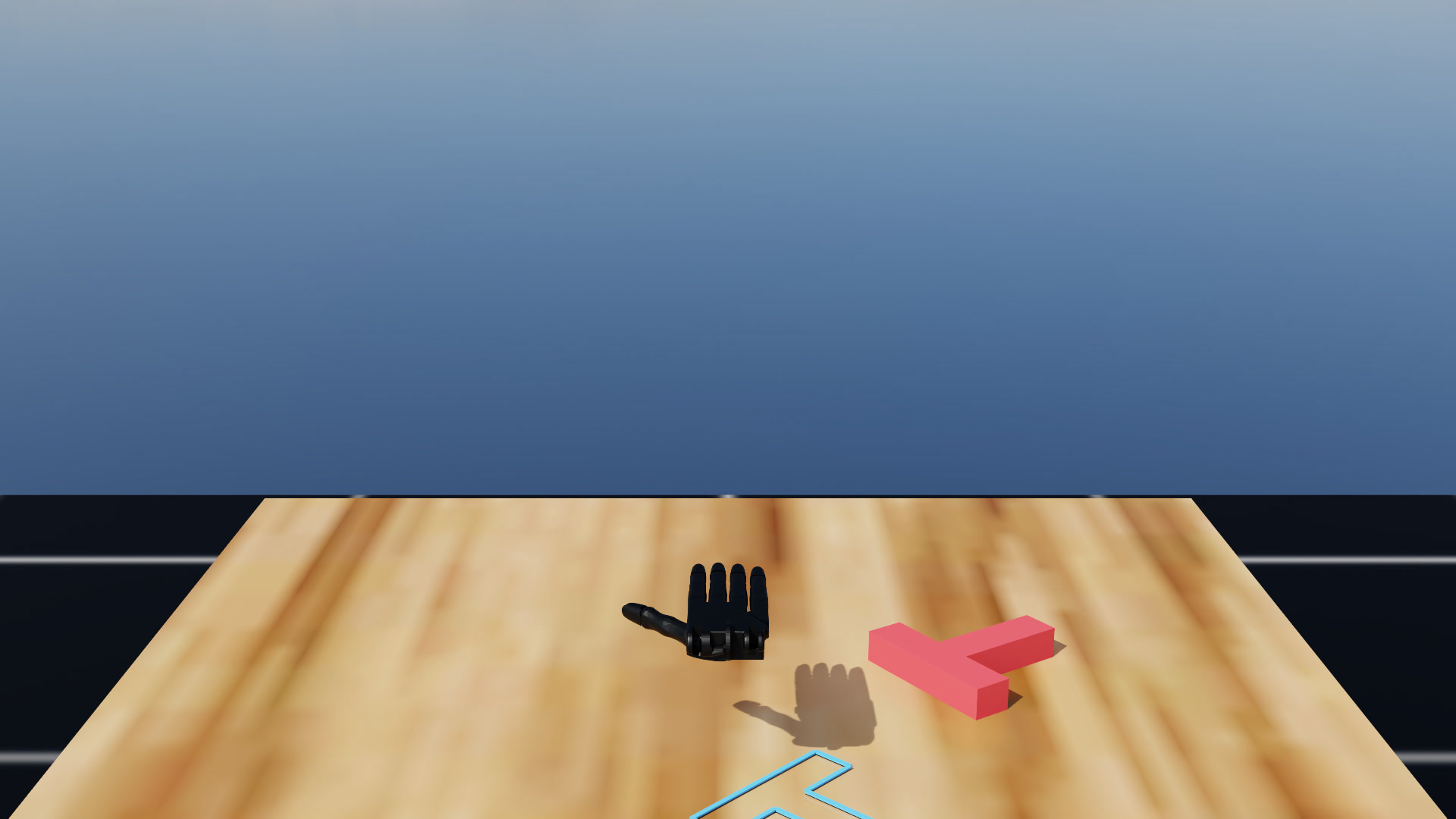} \\

    \addlinespace

    & \shortstack[l]{\texttt{PivotLargeCuboid}\\\texttt{AgainstWall}}
    & Use the hand and wall support to pivot a large thin cuboid into an upright pose.
    & The cuboid is lifted at least 0.20 m above its reset height and tilted to within 20$^\circ$ of vertical.
    & \includegraphics[width=0.95\linewidth]{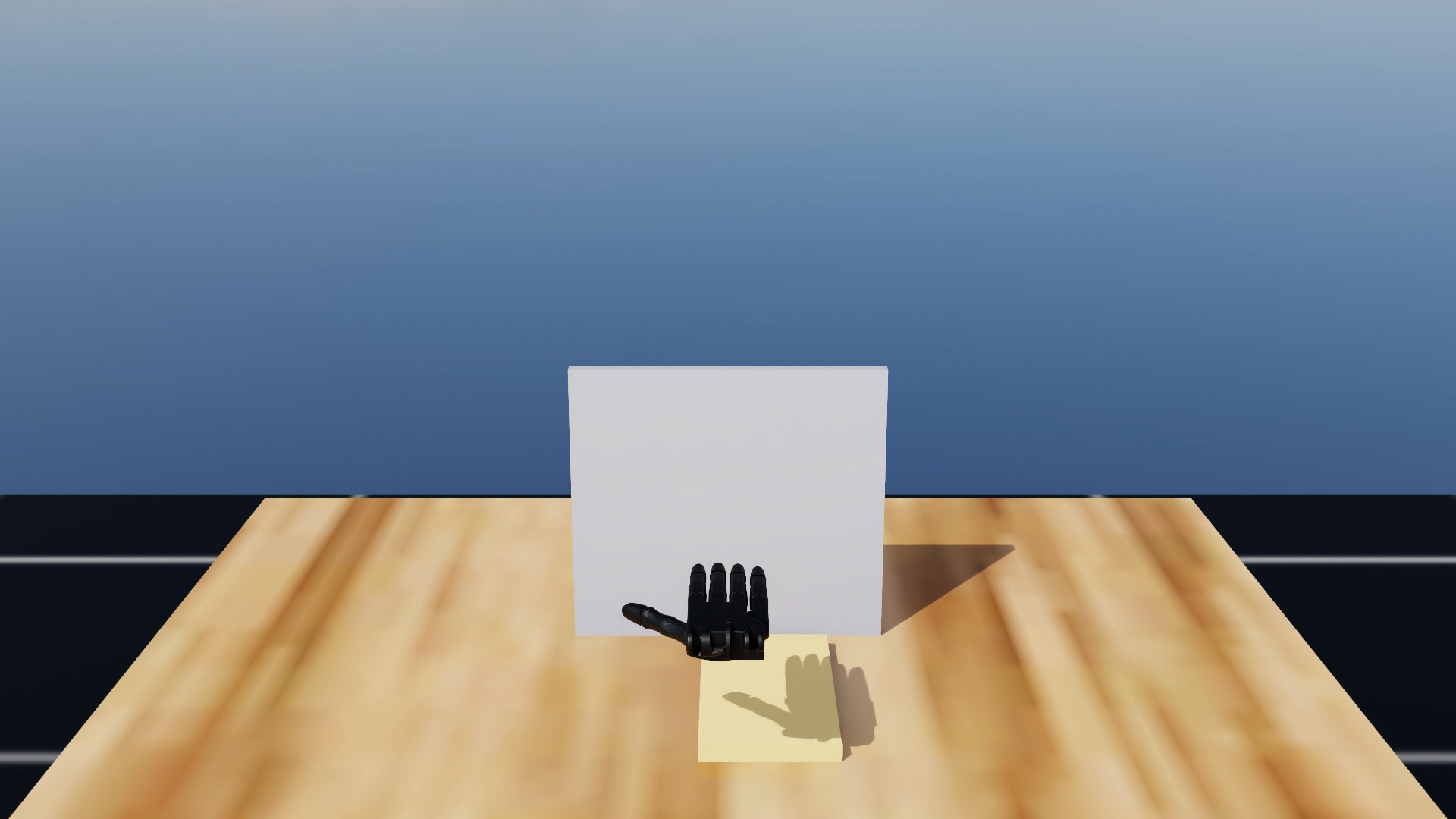} \\

    \addlinespace

    & \shortstack[l]{\texttt{TakeBook}\\\texttt{OffShelf}}
    & Extract a target book from a shelf and move it to the commanded pose.
    & The book's position is within 0.05 m and its orientation within about 26$^\circ$ of a goal that sits 0.18 m out and 0.05 m up from the shelf slot, tilted downward by 65$^\circ$.
    & \includegraphics[width=0.95\linewidth]{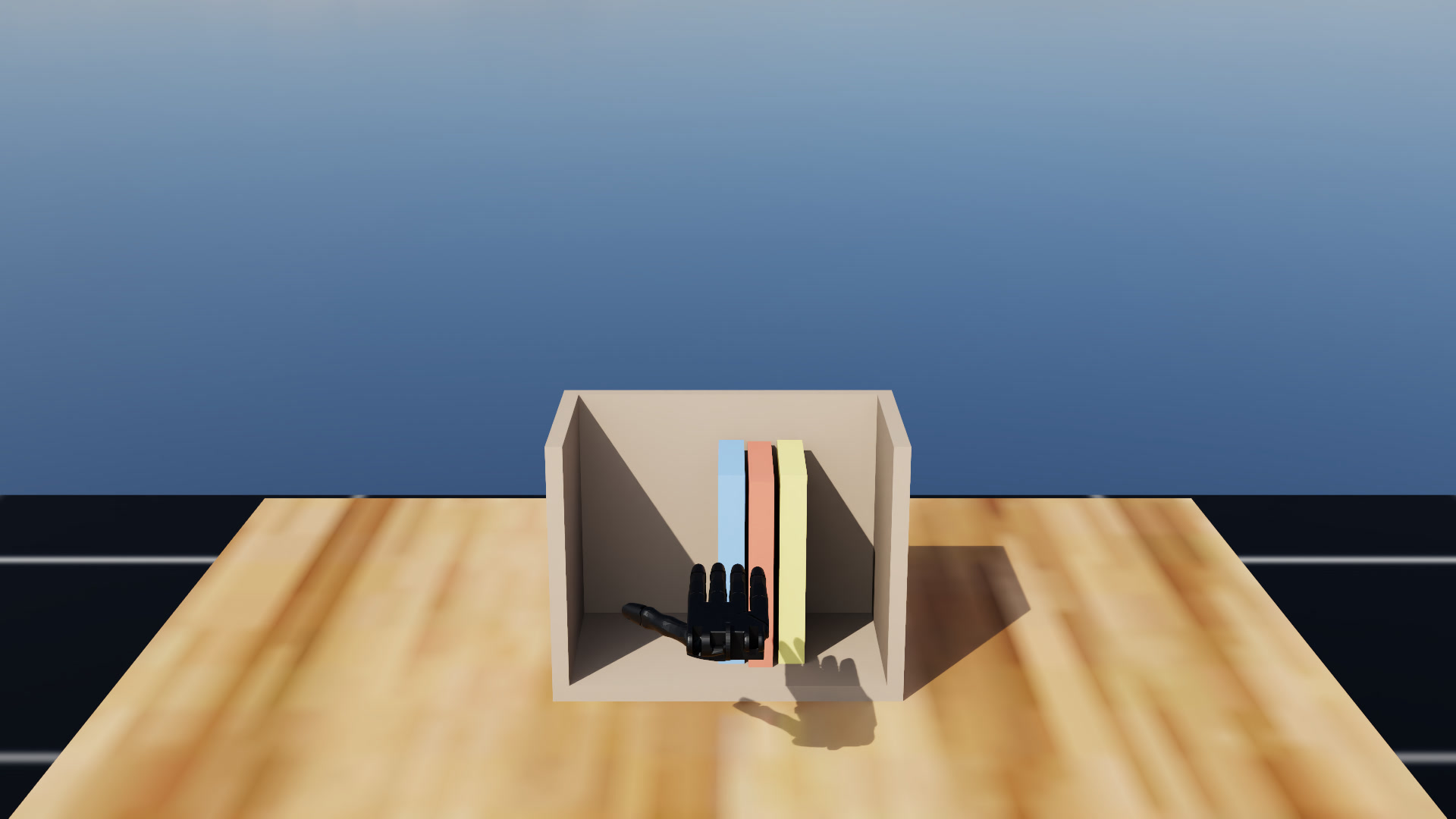} \\

    \addlinespace[4pt]
    \midrule
    \addlinespace[2pt]
    \newpage
    \addlinespace
    \categorycell{5}{Bimanual Coordination}
    & \shortstack[l]{\texttt{BimanualLift}\\\texttt{Tray}}
    & Use both hands to lift a tray while keeping the carried surface level.
    & The tray is lifted at least 0.20 m above reset height and kept within 10$^\circ$ of level.
    & \includegraphics[width=0.95\linewidth]{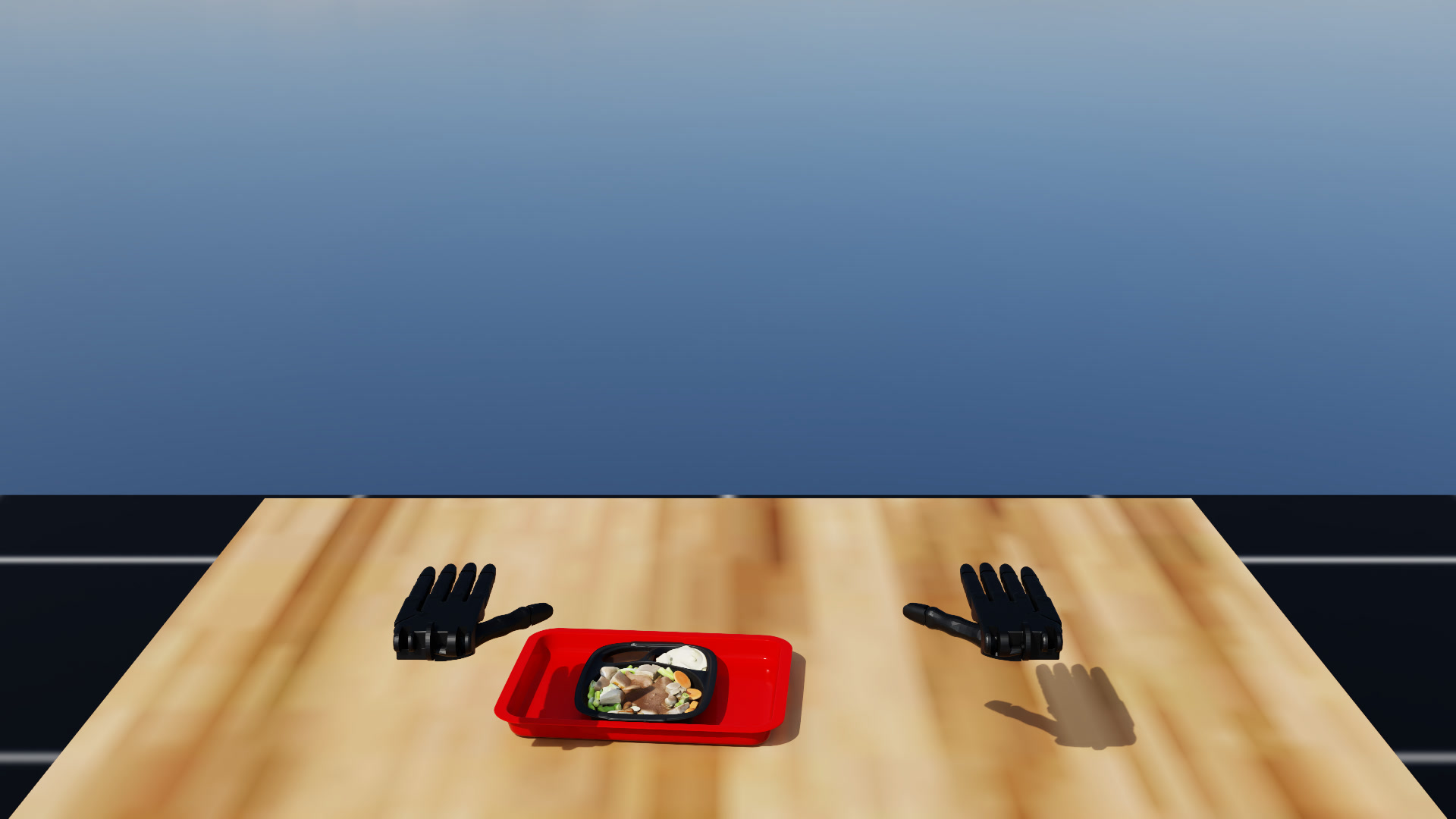} \\

    \addlinespace

    & \shortstack[l]{\texttt{BimanualLift}\\\texttt{Basket}}
    & Use both hands to grasp and lift a basket while maintaining a level carrying pose.
    & The basket is lifted at least 0.20 m above reset height and kept within 10$^\circ$ of level.
    & \includegraphics[width=0.95\linewidth]{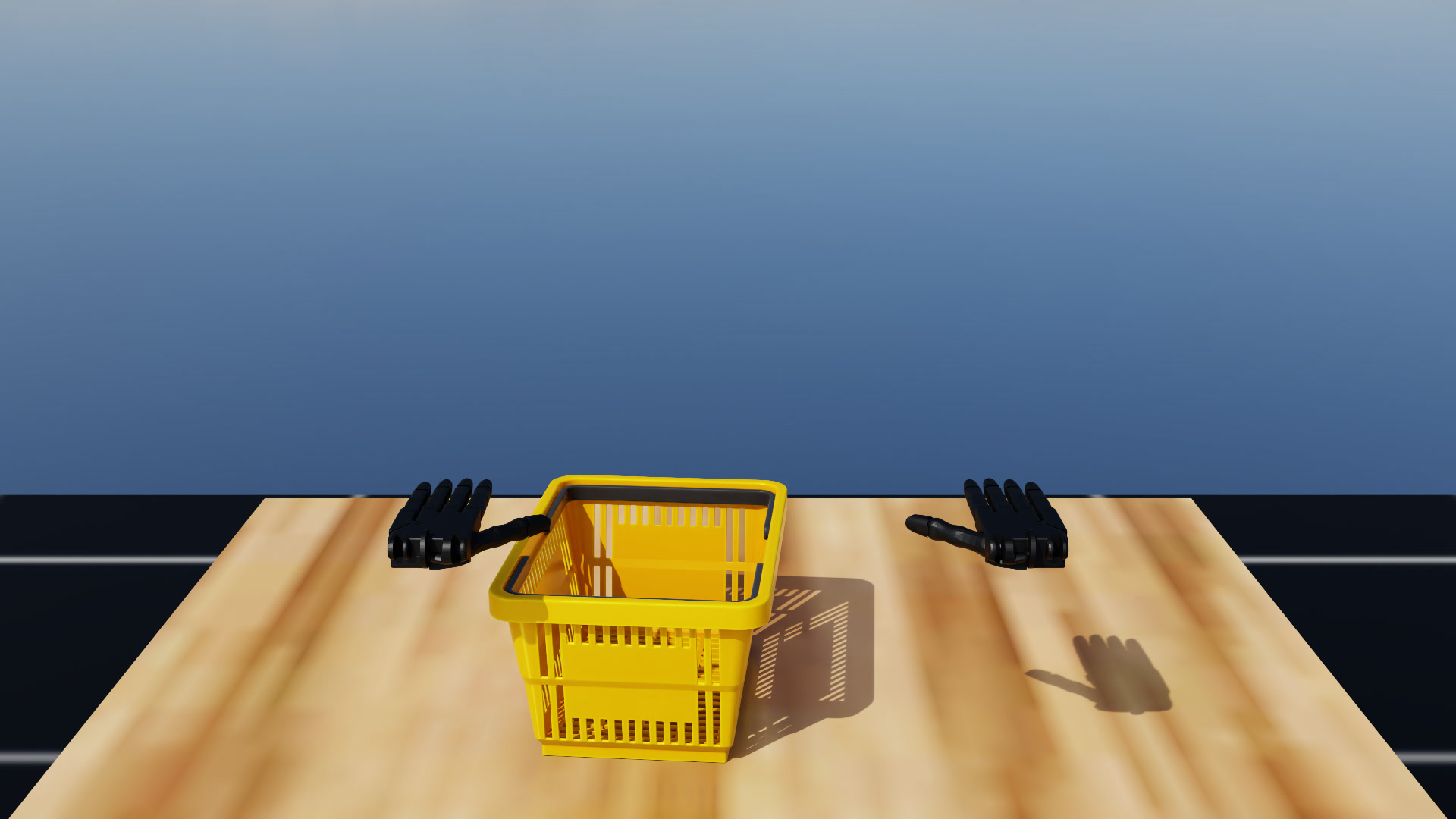} \\

    \addlinespace

    & \shortstack[l]{\texttt{BimanualLift}\\\texttt{Carton}}
    & Use both hands to grasp a carton and lift it clear of the tabletop.
    & The carton is lifted at least 0.20 m above its spawn height.
    & \includegraphics[width=0.95\linewidth]{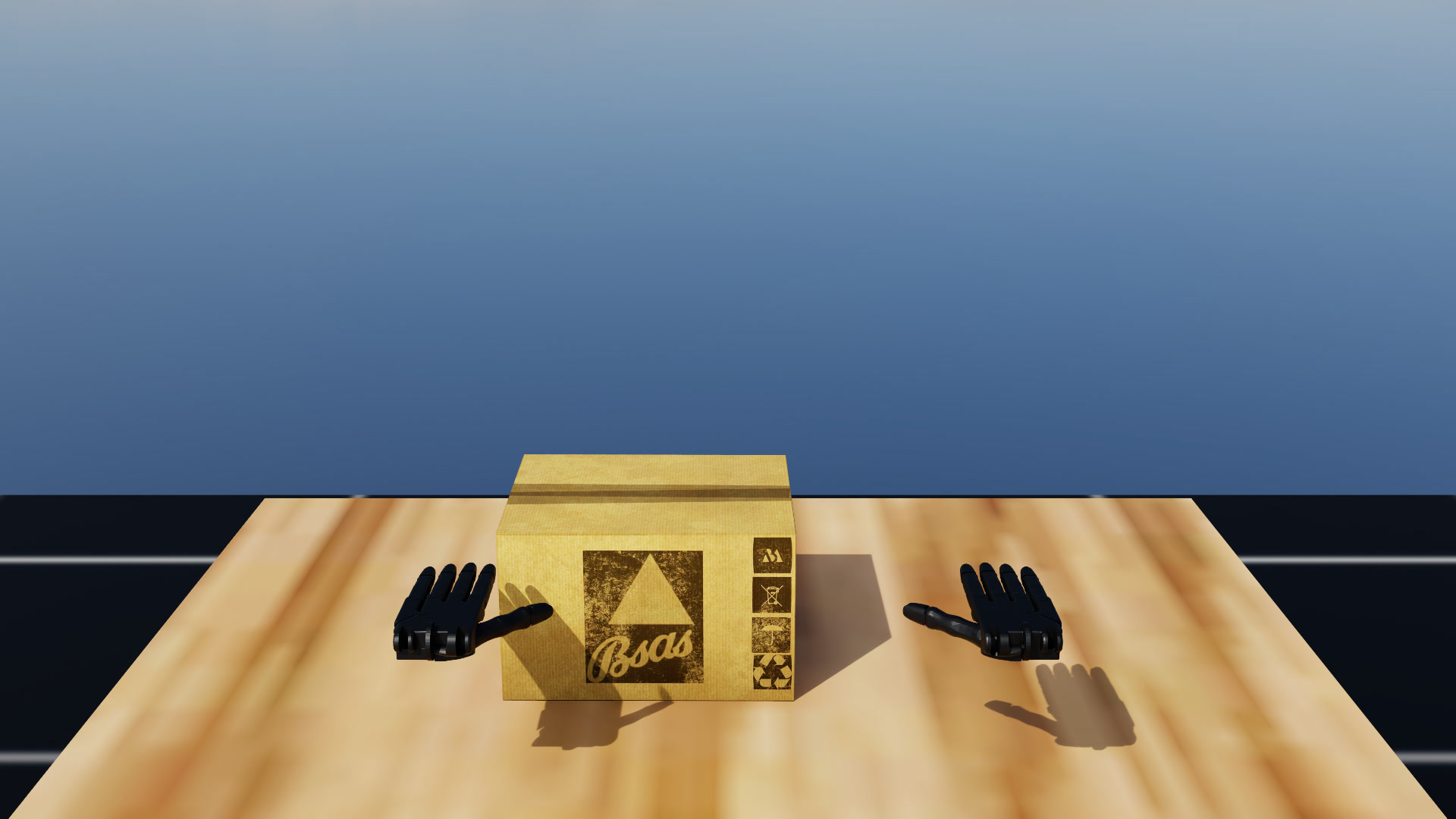} \\

    \addlinespace

    & \shortstack[l]{\texttt{BimanualLift}\\\texttt{DutchOven}}
    & Use both hands to grasp a Dutch oven and lift the cookware from the table.
    & The Dutch oven is lifted at least 0.30 m above its spawn height.
    & \includegraphics[width=0.95\linewidth]{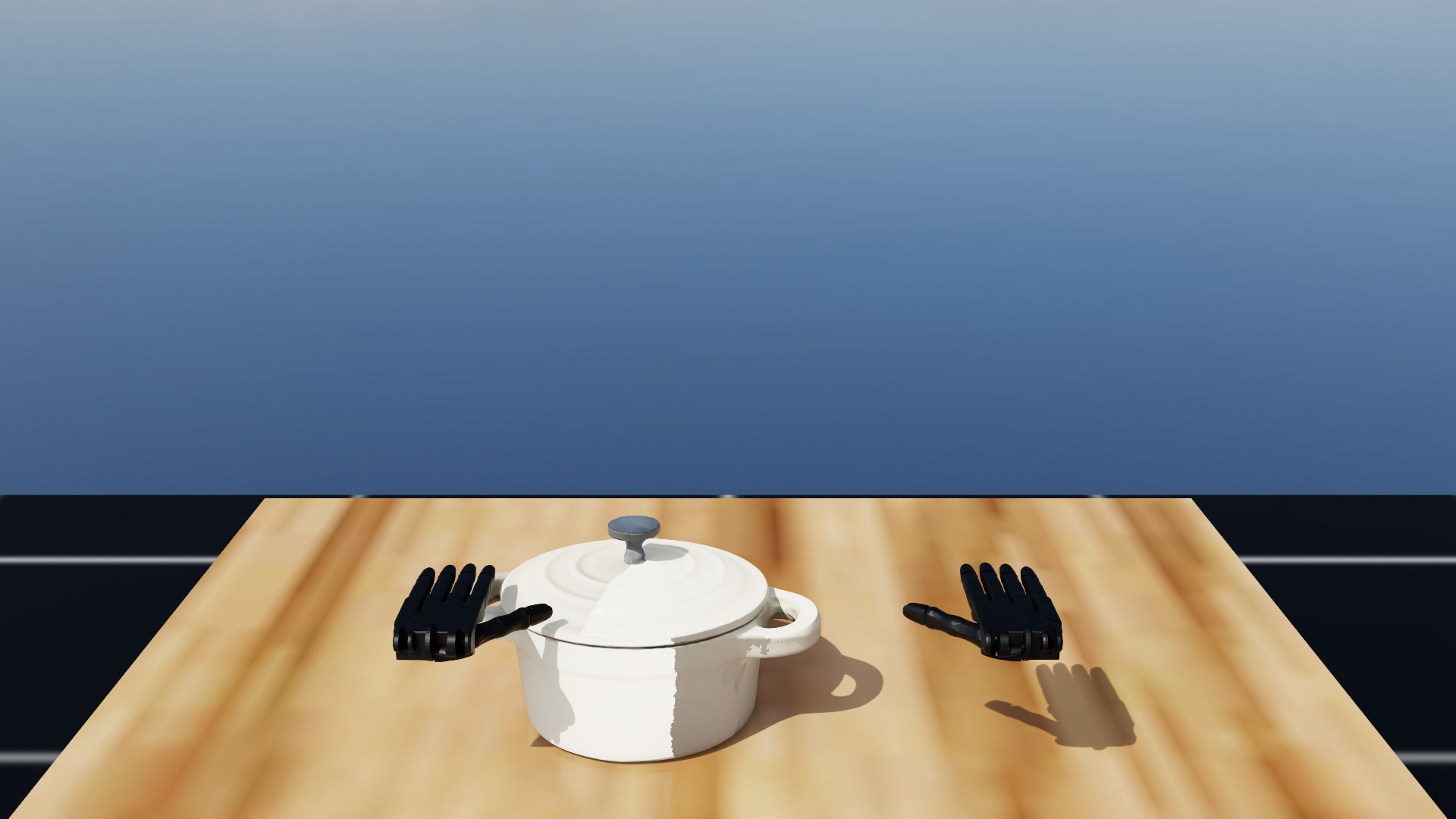} \\

    \addlinespace

    & \shortstack[l]{\texttt{Bimanual}\\\texttt{Handover}}
    & Transfer a cube so that the left hand holds it while the right hand releases it and the object remains stable.
    & The left hand grips firmly the cube with a contact force of at least 1.6 N on at least one link, while the right hand having no active contact with the cube.
    & \includegraphics[width=0.95\linewidth]{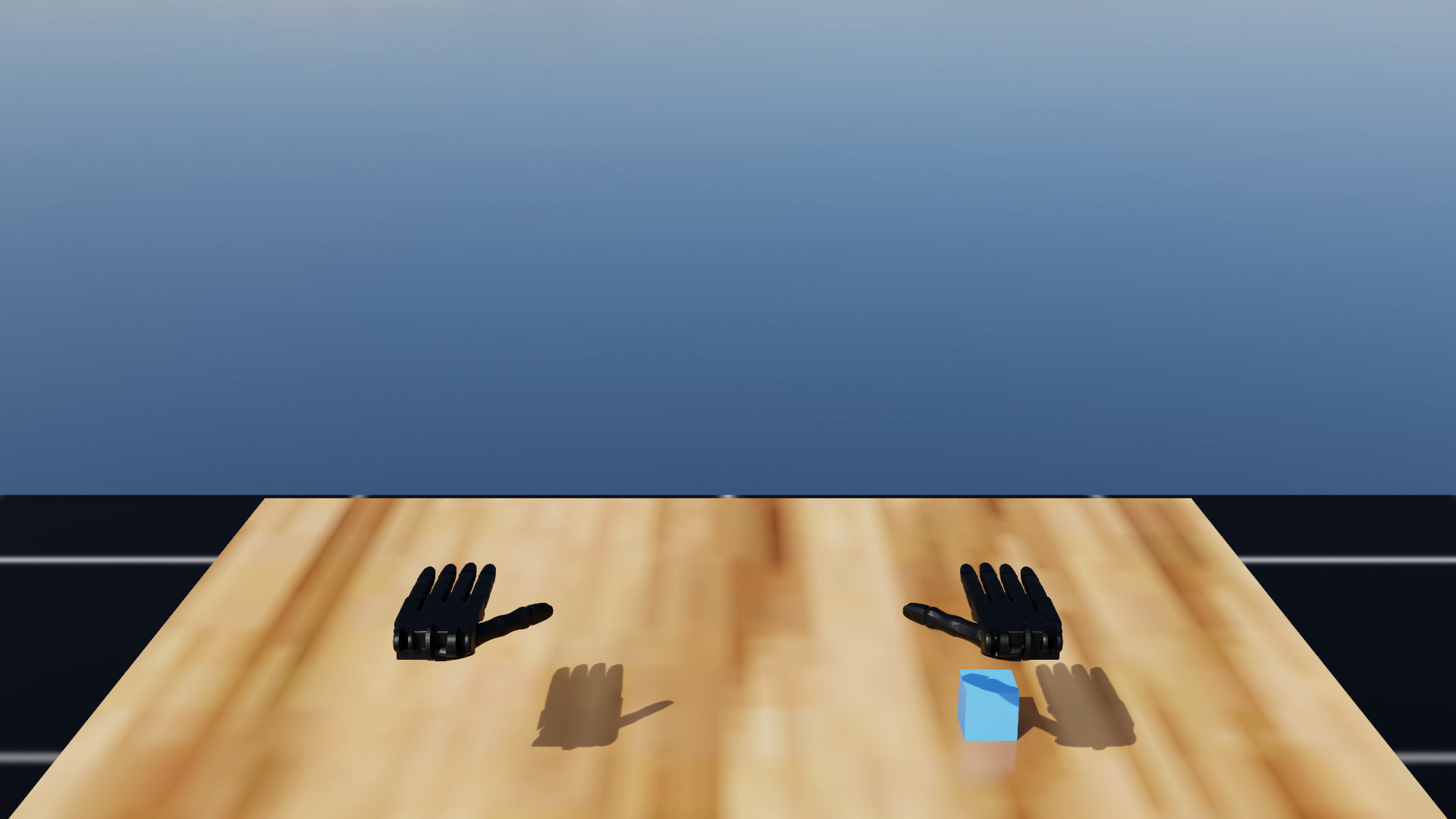} \\

    \addlinespace[4pt]
    \midrule
    \addlinespace[2pt]

    \addlinespace
    \categorycell{5}{Long-horizon}
    & \shortstack[l]{\texttt{CookFood}}
    & Pour ingredients into a pot, keep the food contained, move the pot to the stove, and turn the stove on.
    & First, in any order: the bottle and the can are each lifted at least 0.10 m, tilted at least 100$^\circ$, and their spout brought within 0.10 m of the pot. Then, all at once: the stove knob is turned at least 45$^\circ$, the pot is centered within 0.08 m of the burner and stays within 15$^\circ$ of flat, and all active ingredients remain inside the pot.
    & \includegraphics[width=0.95\linewidth]{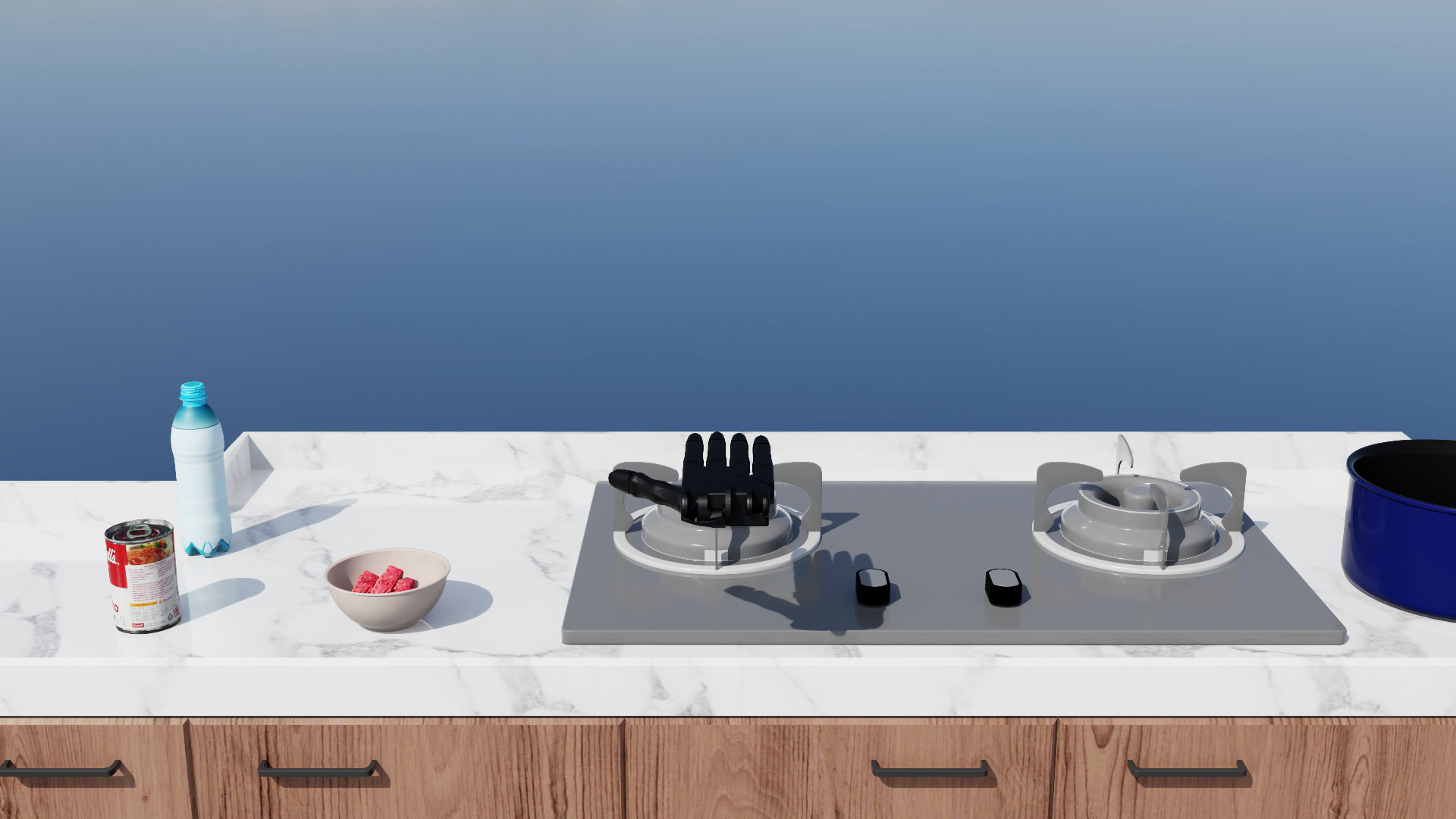} \\

    \addlinespace

    & \shortstack[l]{\texttt{MakeCoffee}}
    & Pour milk into a mug, place the mug under the coffee machine, and activate the machine switch.
    & In order: pour the milk into the mug by tilting the bottle at least 70$^\circ$ and bringing its spout close to the mug, place the mug under the coffee machine within about 0.09 m horizontally and 0.07 m vertically of its target spot, then rotate the switch lever at least 35$^\circ$ from rest.
    & \includegraphics[width=0.95\linewidth]{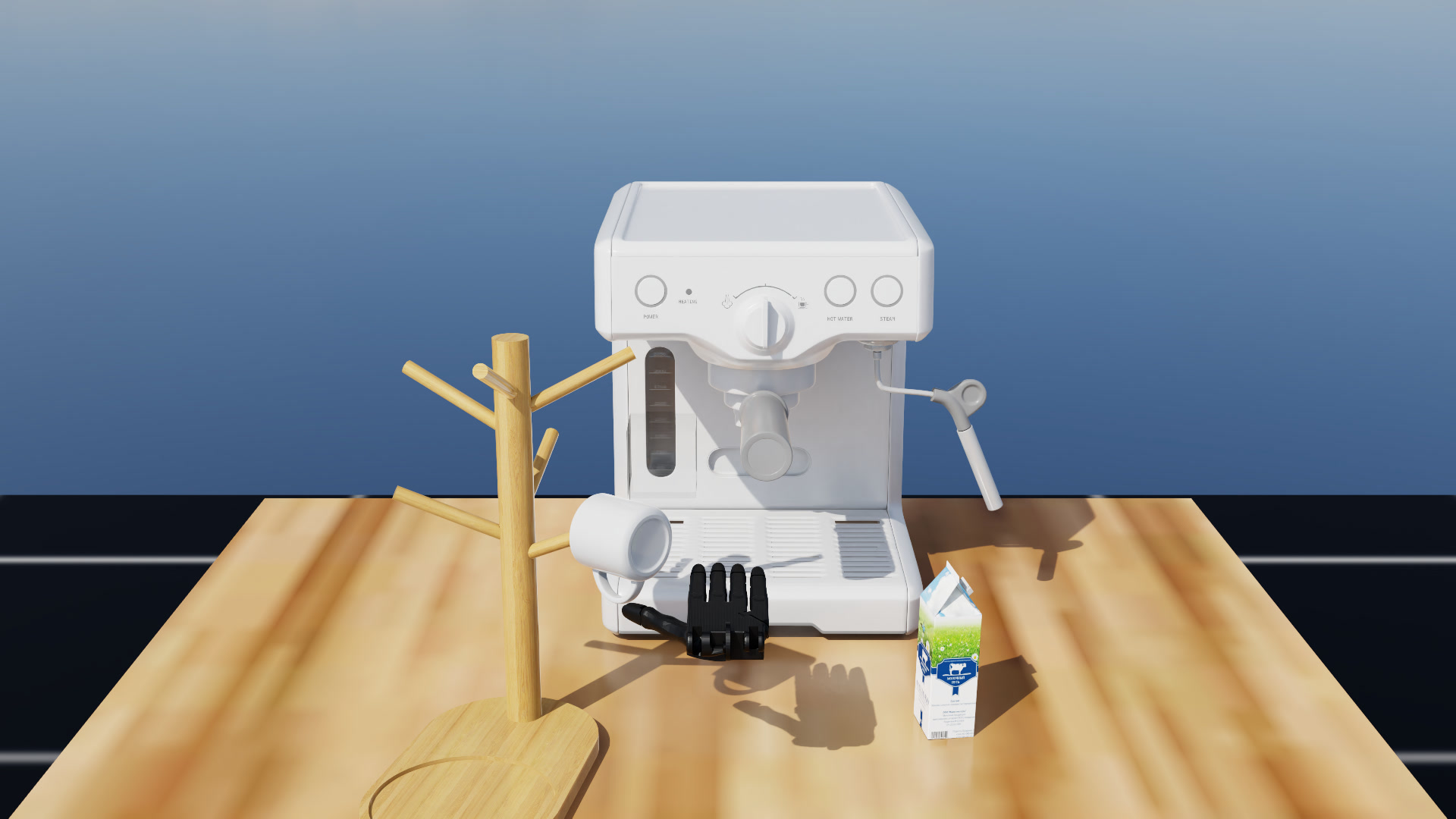} \\

    \addlinespace

    & \shortstack[l]{\texttt{MicrowaveFood}}
    & Open the fridge, retrieve food, place it into the microwave, close the appliance, and turn the dial.
    & In order: open the fridge at least a quarter of the way, close it back down, open the microwave door at least halfway, place the food within 0.15 m of its goal, close the microwave door, then turn either dial at least 18$^\circ$ from rest.
    & \includegraphics[width=0.95\linewidth]{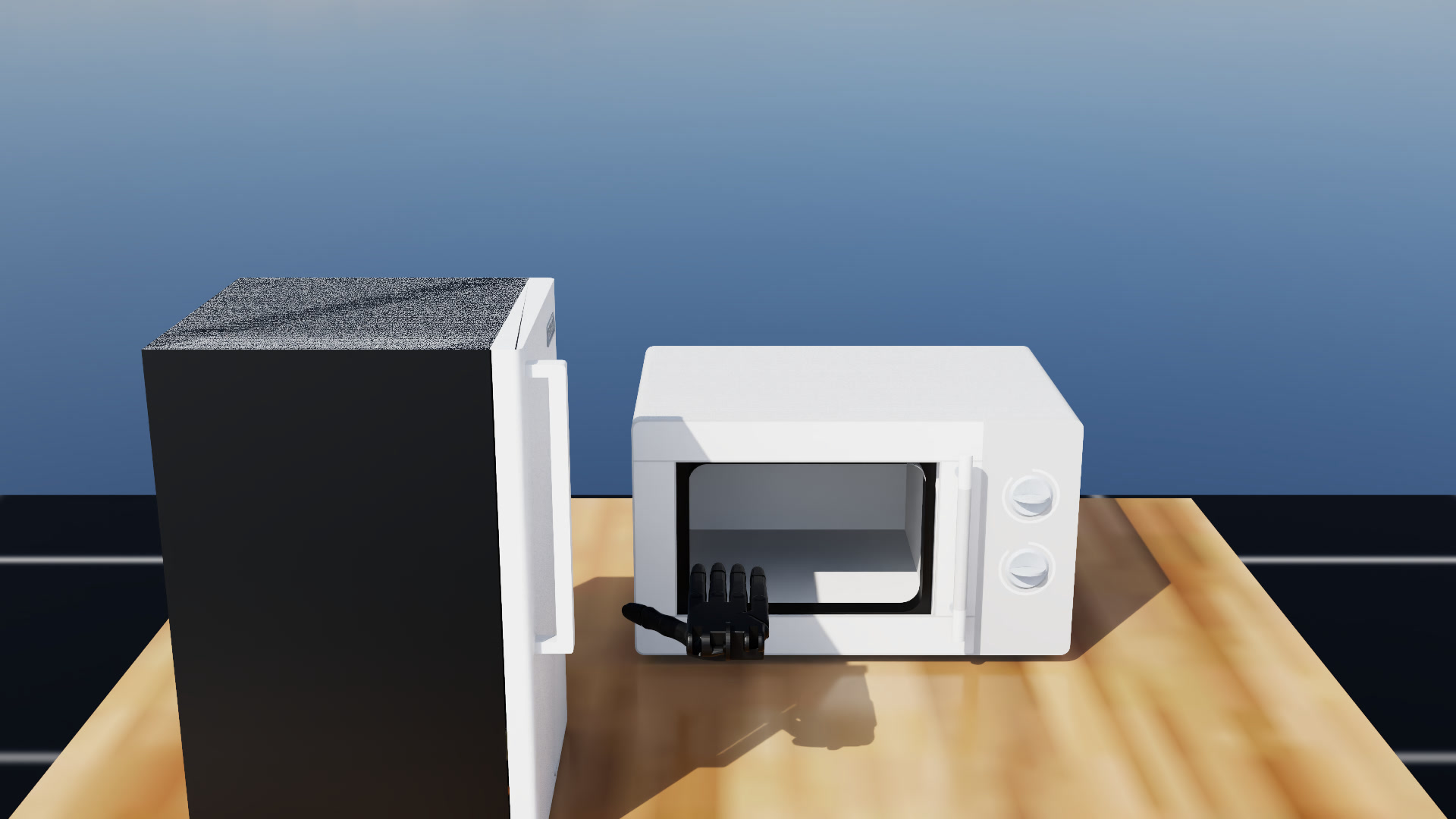} \\

    \addlinespace

    & \shortstack[l]{\text{OvenBake}}
    & Season salmon with condiments, load it into the oven, close the oven door, and start the oven.
    & First, in any order, the salmon is seasoned with both salt and pepper by lifting, tilting, and pouring each jar near the salmon. Then, in order: the salmon is placed inside the oven cavity, the door is closed while the salmon stays inside, and the oven knob is turned about 40$^\circ$ while the salmon still stays inside.
    & \includegraphics[width=0.95\linewidth]{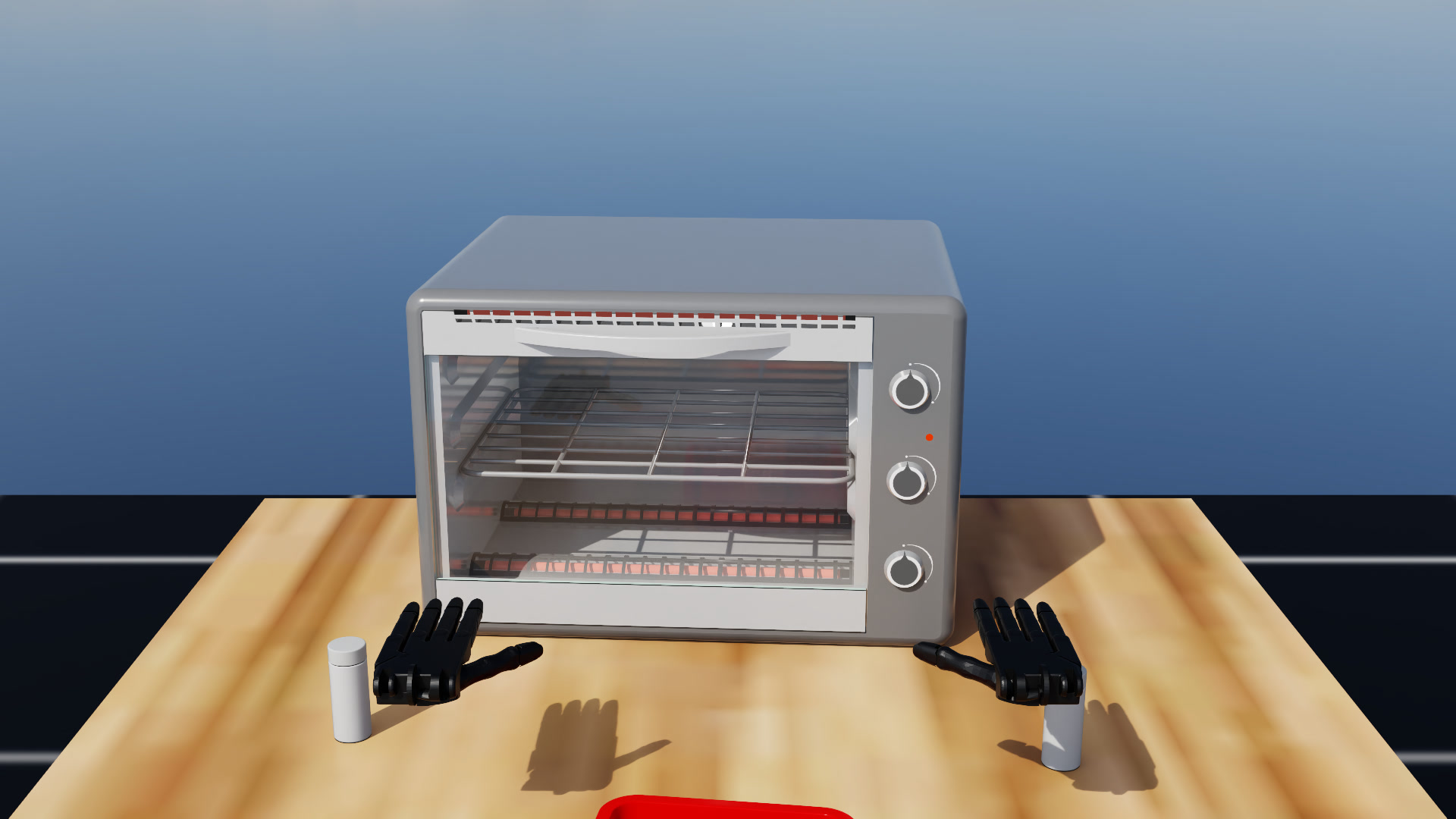} \\

    \addlinespace

    & \shortstack[l]{\texttt{CleanTable}}
    & Sort active objects into the drawer or trash can, then close the corresponding receptacles.
    & In any order: the drawer is fully closed with every active drawer-assigned object contained inside it, and the trash lid is closed and latched with every active trash-assigned object contained inside it.
    & \includegraphics[width=0.95\linewidth]{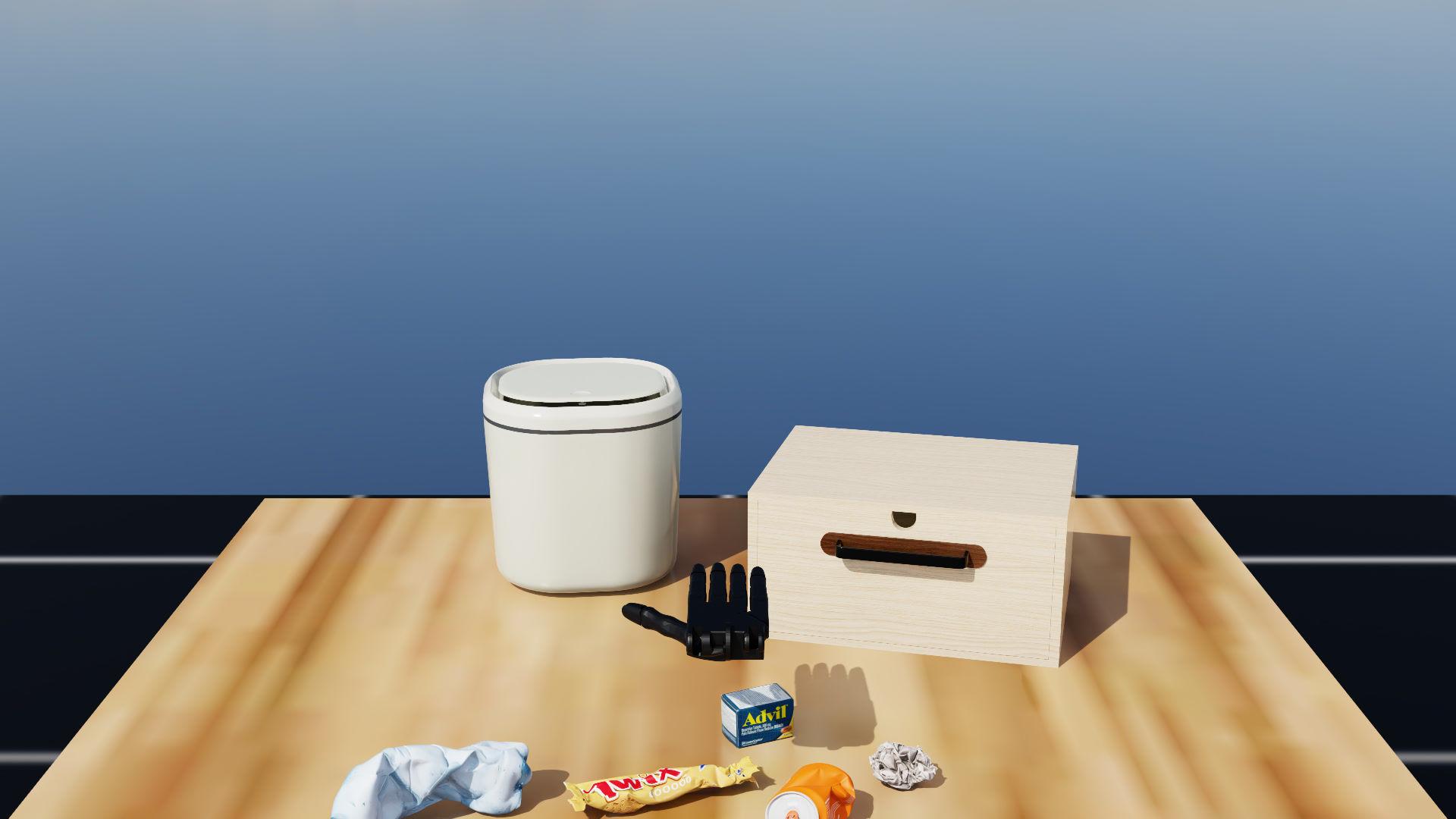} \\

\end{longtable}
\endgroup

\subsection*{Multi-goal composite tasks (\texttt{multigraspenvs})}
\label{sec:multigoal-tasks}

The \texttt{multigraspenvs} benchmark composes 39 environment variants by pairing one of nine articulated-object tasks drawn from the primitive and articulation categories with one of four rigid-object addon tasks. The primary set is \{OpenMicrowave, OpenDoor, OpenDrawer, LiftLid, GraspPot, PushButton, RotateKnob, TurnOnSwitch, GraspBucket\}, and the addon set is \{PourCan, PickUpStick, PourMug, RelocateSphere\}. Every primary task is paired with every addon, giving $9 \times 4 = 36$ variants, and PourMug additionally serves as a primary task paired with the remaining three addons \{PourCan, PickUpStick, RelocateSphere\}, giving 3 more variants, for 39 total. Each variant places both the primary task's asset and the addon object in the same scene, and the agent must interact with both.

An environment is marked successful when both of the following hold at the same time:
\begin{itemize}
    \item The addon object is lifted at least $h_{\text{addon}}$ above its reset height, regardless of which addon task it originally comes from.
    \item The primary task's articulated joint also reaches its own single-task success threshold $\tau_{\text{primary}}$. For when the primary object is the mug, the same lift-then-pour procedure apply. 
\end{itemize}

Both conditions must hold simultaneously for the task to count as a success. $h_{\text{addon}}$ is a single fixed lift threshold shared across all 39 variants, while $\tau_{\text{primary}}$ is task-specific and matches each primary task's own standalone threshold.

\section{Imitation Learning Baseline \& Implementation Details}

\noindent\textbf{$\pi_{0.5}$.}
We fine-tune the official 3.3B-parameter $\pi_{0.5}$ checkpoint.
At each control step the observation comprises (i) two 256$\times$256 RGB images from a fixed third-person camera and a wrist-mounted camera (three cameras -- third-person, left wrist, and right wrist -- for the bimanual variant) and (ii) the per-task natural-language instruction; the policy is conditioned on vision and language only, as the proprioceptive-state input is disabled.
The Gemma action expert predicts a 10-step chunk of absolute 28-/56-dimensional joint targets via flow matching.
We fully fine-tune all model parameters on 32 H20 GPUs for 2K steps, using AdamW ($\beta=(0.9, 0.95)$, weight decay $10^{-10}$, gradient-norm clipping at $1.0$), a global batch size of $512$, and a cosine learning-rate schedule (100-step warmup, peak $10^{-4}$ decaying to $10^{-6}$).

\noindent\textbf{OpenVLA.}
We fine-tune \texttt{openvla-7b}~\cite{kim24openvla} with the OFT recipe~\cite{kim2025fine}, replacing the discrete 7-token action head with a continuous $L_1$-regression head that emits an 8-step chunk of absolute joint targets.
The observation comprises the same RGB camera set as $\pi_{0.5}$ (two views single-hand, three views bimanual), the current 28-/56-dimensional joint position encoded by a learned proprioceptive projector, and the per-task language instruction.
Fine-tuning attaches LoRA (rank 32) adapters to every linear layer of the base VLA, while the regression head and the proprioceptive projector are trained from scratch; the underlying base weights remain frozen.
We train on 8 H20 GPUs for 3K steps with AdamW at a learning rate of $10^{-4}$ (decayed by $10\times$ via a multi-step schedule), a per-GPU batch size of $8$, and random-crop image augmentation.

\noindent \textbf{Diffusion Policy.}
Following the state-based variant of diffusion policy~\cite{chi2025diffusion}, we train a separate policy per task on the normalized proprioceptive state, with no visual input. 
The denoiser is the standard 1D conditional U-Net with FiLM conditioning, channel widths $(256, 512, 1024)$ and two residual blocks per stage. Observations are passed through a $256$-d encoder and projected to a $128$-d conditioning vector. We use an observation horizon of $T_o{=}2$ and predict action chunks of length $T_a{=}16$ at stride $1$, re-planning from fresh observations after each chunk. The denoiser is trained with the DDPM objective over $100$ timesteps under a squared-cosine $\beta$ schedule, and $20$ denoising steps are taken at inference. We optimize with AdamW ($\text{lr}{=}1{\times}10^{-4}$, weight decay $1{\times}10^{-4}$), batch size $256$, gradient-norm clipping at $1.0$, and an EMA of decay $0.995$ on the policy weights. Each policy is trained for up to $300$ epochs with early stopping on the held-out validation loss.
Following the state-based variant of Diffusion Policy~\cite{chi2025diffusion}, we train a per-task U-Net denoiser on normalized proprioceptive state with no visual input. 
It uses an observation horizon of $T_o{=}2$ and predicts action chunks of length $T_a{=}16$, re-planning after each chunk. 

\noindent \textbf{3D Diffusion Policy.}
Following DP3~\cite{Ze2024DP3}, we train a separate policy per task that conditions on a point cloud plus proprioception and has no RGB or language input.
Each observation is a workspace-cropped point cloud back-projected from a single front-view depth camera (two views are fused for bimanual tasks), expressed in world frame and farthest-point-downsampled to $N{=}512$ points; the proprioceptive input is the current 28-/56-dimensional joint position.
A lightweight PointNet encoder (per-point MLP with LayerNorm, max-pooled, projected to $64$-d) summarizes the cloud, and the proprio history is flattened and concatenated to form the global conditioning vector.
The denoiser is the same 1D conditional U-Net with FiLM conditioning as in the state-based variant, but widened to channel widths $(512, 1024, 2048)$ with two residual blocks per stage (kernel size $5$, GroupNorm with $8$ groups).
It uses an observation horizon of $T_o{=}2$ and predicts action chunks of length $T_a{=}16$, executing the first $8$ steps before re-planning.
The denoiser is trained with the DDIM objective over $100$ timesteps under a squared-cosine $\beta$ schedule, and $10$ denoising steps are taken at inference.
Per-task statistics over the training split normalize point clouds, proprioception, and actions to $[-1,1]$.
We optimize with AdamW ($\text{lr}{=}1{\times}10^{-4}$, weight decay $1{\times}10^{-6}$, $\beta{=}(0.95, 0.999)$), batch size $128$, gradient-norm clipping at $1.0$, and an EMA of max-decay $0.9999$ on the policy weights.
Each policy is trained for $100$ epochs.

\section{Observation Modes}
\label{sec:appendix_observation_groups}

DexVerse organizes observations into nine groups: \texttt{policy}, \texttt{proprio}, \texttt{contact}, \texttt{state}, \texttt{privileged}, \texttt{goal}, \texttt{rgb}, \texttt{depth}, and \texttt{pointcloud}. Each group holds observation terms with a single semantic role, so a consumer (policy, asymmetric critic, or perception backbone) can subscribe to exactly the groups it needs.

\begin{itemize}
    \item \textbf{\texttt{policy}:}
    Contains the previous action.

    \item \textbf{\texttt{proprio}:}
    Contains the joint positions of every joint of the robots.

    \item \textbf{\texttt{contact}:}
    Contains a per-fingertip 3D contact-force vector for each configured contact sensor, expressed in the robot base frame. This group is \texttt{None} when contact sensors are disabled for a task.

    \item \textbf{\texttt{state}:}
    Contains observable task state such as object poses, articulated-object poses, joint positions. Also contains derived geometry such as functional-point positions and functional-axis directions, and, for the long-horizon tasks, stage progress. The exact terms are task-specific; this group is \texttt{None} for tasks that do not populate it.

    \item \textbf{\texttt{privileged}:}
    Contains simulation-only quantities that are withheld from the policy, such as robot joint velocities, full body state (position, orientation, linear velocity, and angular velocity) for the robot's hand links, and, where applicable, object linear and angular velocities.

    \item \textbf{\texttt{goal}:}
    Contains the commanded task target, such as a target object pose or position that the object should reach for task success. This group is \texttt{None} for tasks with no separate goal, such as PickCube.

    \item \textbf{\texttt{rgb}:}
    Contains an RGB image from each configured camera, kept as separate per-camera tensors.

    \item \textbf{\texttt{depth}:}
    Contains a distance-to-image-plane depth image from each configured camera, kept as separate per-camera tensors.

    \item \textbf{\texttt{pointcloud}:}
    Contains a point cloud from a single camera, or merged from multiple camera views when multiview is enabled.
\end{itemize}

Task configurations select the observation groups and terms required by each environment, giving downstream policies a unified interface across state-based, image-based, point-cloud-based, and privileged observations.

\subsection*{Observation-mode presets}
\label{sec:appendix_observation_presets}

While the user is free to select the observation groups to use, DexVerse also defines some observation-mode presets, each of which enables a fixed set of groups and disables every other managed group. A preset is applied to an environment's \texttt{ObservationsCfg} through a config field. The \texttt{privileged} group is never enabled by any preset since it contains information that is hardly observable in the real world and should only be intentionally enabled by the user. 

\begin{table}[h]
    \centering
    \begin{tabular}{lccc}
    \toprule
    Preset & Groups enabled & History length & Multiview \\
    \midrule
    \texttt{state} & policy, proprio, contact, state, goal & 0 & no \\
    \texttt{rgb} & policy, proprio, goal, rgb & 3 & no \\
    \texttt{rgb\_depth} & policy, proprio, goal, rgb, depth & 3 & no \\
    \texttt{pointcloud} & policy, proprio, goal, pointcloud & 3 & no \\
    \texttt{3view\_rgb} & policy, proprio, goal, rgb & 3 & yes \\
    \texttt{3view\_rgb\_depth} & policy, proprio, goal, rgb, depth & 3 & yes \\
    \texttt{3view\_pointcloud} & policy, proprio, goal, pointcloud & 3 & yes \\
    \bottomrule
    \end{tabular}
    \caption{Observation-mode presets. Each preset enables only the listed groups and sets history length to the given length}
    \label{tab:observation_presets}
\end{table}

\end{document}